\documentclass[10pt,twocolumn,letterpaper]{article}

\usepackage[pagenumbers]{cvpr} 

\usepackage[accsupp]{axessibility}
\usepackage{graphicx}
\usepackage{amsmath}
\usepackage{amssymb}
\usepackage{booktabs}
\usepackage{multirow}
\usepackage{array}
\usepackage{appendix}
\usepackage{tabularx}

\newcommand \blfootnote[1]{
    \begingroup
        \renewcommand
        \thefootnote{}\footnote{#1}
        \addtocounter{footnote}{-1}
        \vspace{-1ex}
    \endgroup
}

\usepackage[utf8]{inputenc} 
\usepackage[T1]{fontenc}    
\usepackage{url}            
\usepackage{amsfonts}       
\usepackage{nicefrac}       
\usepackage{microtype}      
\usepackage{xcolor}         
\usepackage{float}
\usepackage{mathrsfs}

\definecolor{cvprblue}{rgb}{0.21,0.49,0.74}
\usepackage[pagebackref,breaklinks,colorlinks,allcolors=cvprblue]{hyperref}

\title{Training for Identity, Inference for Controllability:\\A Unified Approach to Tuning-Free Face Personalization}
\author{
Lianyu Pang$^{1}$\quad 
Ji Zhou$^{1}$\quad 
Qiping Wang$^{2}$\quad
Baoquan Zhao$^{1}$\\
Zhenguo Yang$^{3}$\quad 
Qing Li$^{4}$\quad 
Xudong Mao$^{1,\dagger}$\\
{$^1$Sun Yat-sen University\quad$^2$East China Normal University}\\
{$^3$Guangdong University of Technology\quad$^4$The Hong Kong Polytechnic University}\\
}

\begin{document}

\twocolumn[{%
\vspace{-1em}
\maketitle
\renewcommand\twocolumn[1][]{#1}%
\vspace{-0.1in}
\begin{center}
    \centering
    \vspace{-17pt}
    \includegraphics[width=1\textwidth]{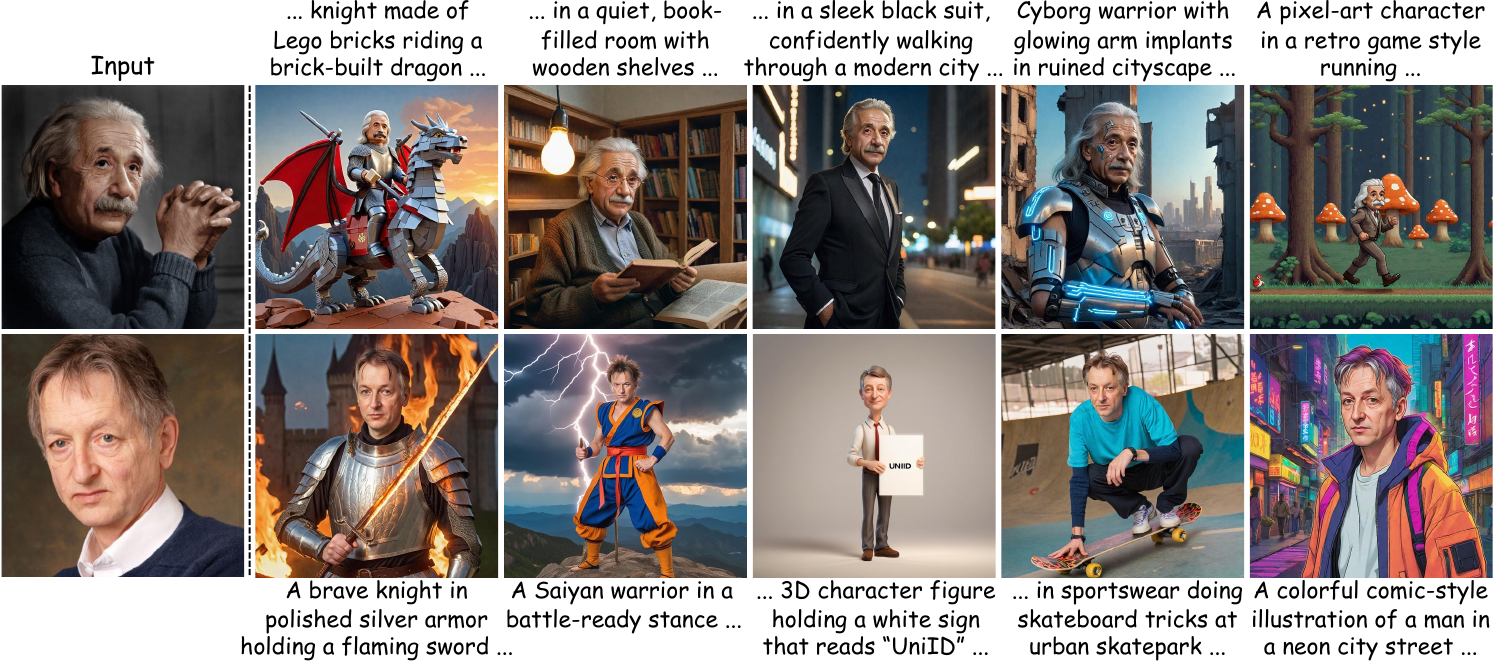}
    \captionof{figure}{
    UniID enables high-quality face personalization with flexible text controllability and consistent identity preservation.
    }
    \label{fig:teaser}
\end{center}

}]

\blfootnote{\textsuperscript{$\dagger$} Corresponding author.}

\begin{abstract}

Tuning-free face personalization methods have developed along two distinct paradigms: text embedding approaches that map facial features into the text embedding space, and adapter-based methods that inject features through auxiliary cross-attention layers. While both paradigms have shown promise, existing methods struggle to simultaneously achieve high identity fidelity and flexible text controllability. We introduce UniID, a unified tuning-free framework that synergistically integrates both paradigms. Our key insight is that when merging these approaches, they should mutually reinforce only identity-relevant information while preserving the original diffusion prior for non-identity attributes. We realize this through a principled training-inference strategy: during training, we employ an identity-focused learning scheme that guides both branches to capture identity features exclusively; at inference, we introduce a normalized rescaling mechanism that recovers the text controllability of the base diffusion model while enabling complementary identity signals to enhance each other. This principled design enables UniID to achieve high-fidelity face personalization with flexible text controllability. Extensive experiments against six state-of-the-art methods demonstrate that UniID achieves superior performance in both identity preservation and text controllability. Code will be available at \url{https://github.com/lyuPang/UniID}.
\end{abstract}
 
\section{Introduction}

Text-to-image personalization has emerged as a critical capability in generative modeling, enabling users to generate customized content by conditioning diffusion models on a few reference images of a target concept. This technology has found widespread applications in image editing~\cite{kawar2023imagic}, virtual try-on~\cite{Zhu_2024_CVPR}, and image animation~\cite{SadTalker}. Among various personalization tasks, face personalization~\cite{xiao2023fastcomposer,basis} presents unique challenges and opportunities: faces are not only among the most frequently personalized subjects but also demand exceptionally high fidelity due to human sensitivity to facial details.

Early personalization methods~\cite{textual-inversion,dreambooth,customdiffusion,pang2024cross} achieve impressive results through test-time fine-tuning of diffusion models. However, fine-tuning-based methods require substantial computational costs per identity. Recent tuning-free approaches~\cite{ipa,wei2023elite,pulid} address this limitation by extracting visual features via pre-trained encoders and injecting them directly into diffusion models, enabling instant personalization without per-identity optimization.

Current tuning-free face personalization typically follows two distinct paradigms. The first paradigm, which we term \textit{text embedding approach}~\cite{gal2023encoderbased,li2023photomaker,xiao2023fastcomposer,ostashev2024moa}, maps facial features extracted by image encoders into the text embedding space. The second paradigm, which we term \textit{adapter approach}~\cite{ipa,pulid,wang2024instantid}, injects features via auxiliary cross-attention layers parallel to text cross-attention layers. While both paradigms have demonstrated success, they exhibit limitations in simultaneously preserving identity fidelity and maintaining text controllability.

\begin{figure}[t]
 \centering
 \includegraphics[width=1.0\linewidth]{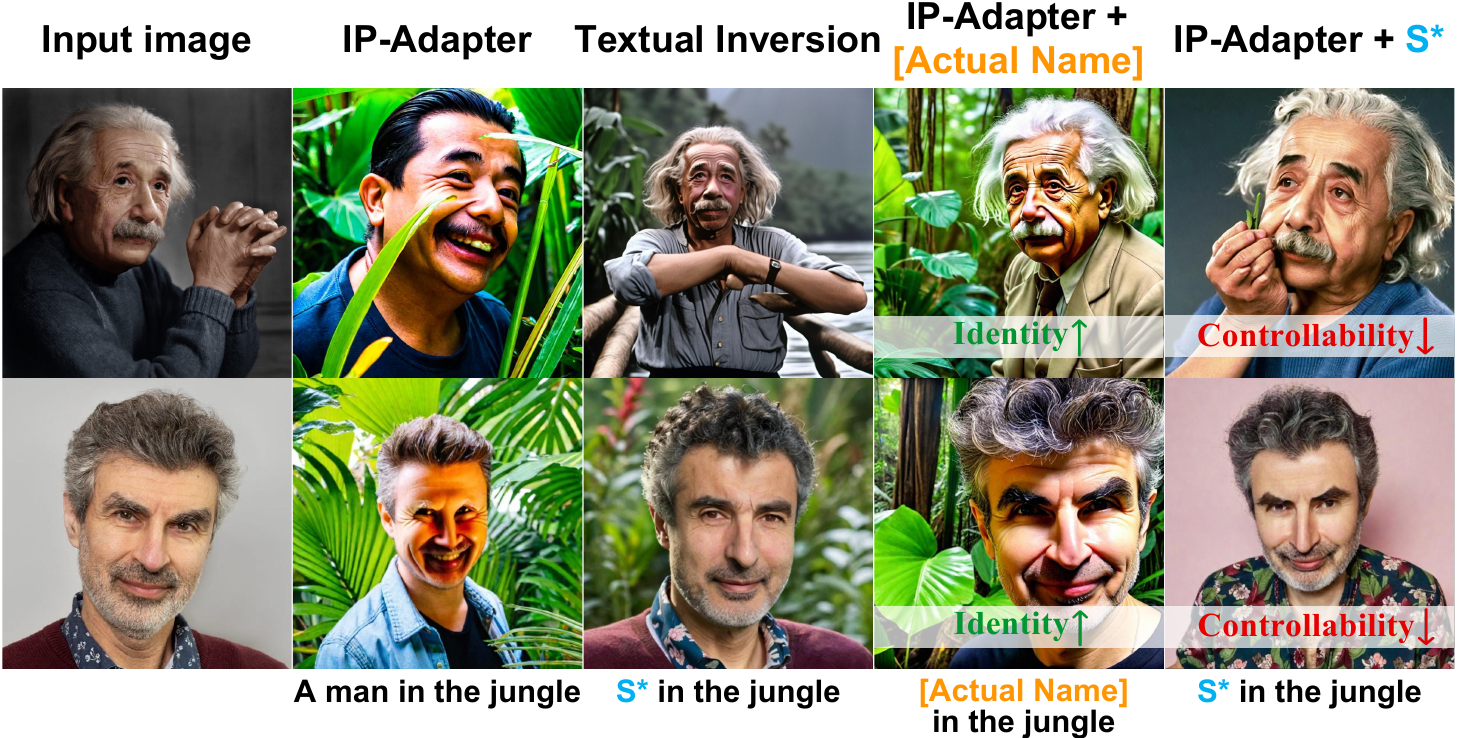}
    \caption{Pairing IP-Adapter with ground-truth identity names significantly enhances identity preservation. However, augmenting IP-Adapter with learned text embeddings substantially degrades text controllability.}
\label{fig:motivation}
\vspace{-8pt}
\end{figure}

Given that these two paradigms inject facial information at fundamentally different locations within the diffusion architecture, a natural question arises: \textit{Can we achieve superior personalization by synergistically combining both approaches, simultaneously leveraging text embeddings and cross-attention injection?} Our preliminary investigation (Figure~\ref{fig:motivation}) reveals several key insights. Pairing IP-Adapter with prompts containing the person's actual name substantially improves identity preservation, confirming that well-designed text embeddings enhance fidelity. However, naively combining learned text embeddings with IP-Adapter causes severe overfitting to the input image, resulting in significant degradation of text controllability. This suggests that effective synergy requires principled coordination of how identity information flows through both branches. Despite the potential benefits, a principled framework for effectively unifying these paradigms remains underexplored.

In this work, we introduce \textit{UniID}, a unified tuning-free framework that synergistically combines text embedding and adapter approaches while preserving both identity fidelity and text controllability. Our key insight is that when merging the two branches, they should mutually reinforce only identity information, while non-identity aspects such as scene composition are controlled by the original diffusion model's prior knowledge. Specifically, during training, we employ an identity-focused learning scheme that guides both the text embedding and adapter branches to capture exclusively identity-relevant features. At inference, we introduce a normalized rescaling strategy that recovers the text controllability of the original diffusion model in both branches while enabling their complementary identity signals to mutually reinforce each other. Through this strategic training-inference paradigm, UniID achieves superior identity fidelity while preserving the text controllability of the original model.

We validate our method through extensive qualitative and quantitative experiments against six state-of-the-art baselines. Through effective integration of the text embedding and adapter branches, UniID achieves superior performance in both identity preservation and text controllability compared to baseline methods.

 \section{Related Work}

\paragraph{Text-to-Image Personalization.}
Text-to-image personalization involves adapting pretrained generative models to synthesize novel images of specific subjects based on user-provided textual prompts. Early approaches primarily adopted optimization-based methods, including tuning new text embeddings~\cite{textual-inversion} or fine-tuning the parameters of diffusion models~\cite{dreambooth,customdiffusion}. Some subsequent studies focus on enhancing the identity preservation of the concept~\cite{voynov2023p,neti,zhou2023enhancing,hua2023dreamtuner,he2023data,jones2024customizing,jiang2024mc}, while others aim to improve text controllability~\cite{tewel2023keylocked,arar2024palp,avrahami2023breakascene,huang2024realcustom}. These approaches achieve high fidelity through extensive fine-tuning, but consequently suffering from significant computational overhead. To mitigate these challenges, a common approach to reduce optimization complexity is to limit the number of parameters for tuning~\cite{hu2022lora,LoRADiffusion,han2023svdiff,basis,Consistent_Characters}. Recent efforts have shifted toward tuning-free personalization methods~\cite{ipa,suti,jia2023taming,shi2023instantbooth,wei2023elite,RB_Modulation,Personalize_anything}, employing an image encoder to inject extracted features into the diffusion model.

\vspace{5pt}
\noindent\textbf{Encoder-based Personalization.}
Encoder-based personalization methods have emerged to address limitations associated with traditional optimization-based approaches, particularly their computational costs associated with per-subject optimization. This area of research has particularly concentrated on the personalization of human faces~\cite{chen2023photoverse,li2023photomaker,ruiz2023hyperdreambooth,valevski2023face0,wang2024instantid,kong2024omg,ostashev2024moa,kim2024instantfamily,wu2024infinite,cui2024idadapter,cheng2024resadapter,wang2024stableidentity,chen2023dreamidentity,StyleGAN_SD,T2I_Adapter,FlashFace}, owing to the broad applicability of facial synthesis tasks. Initial encoder-based techniques~\cite{gal2023encoderbased,q_former} utilize a two-stage framework: first, training an encoder to produce a coarse representation of the subject, and subsequently refining this representation through minimal additional tuning steps. More recent advances aim to entirely eliminate inference-time tuning by injecting encoder-extracted features directly into diffusion models through a purely feed-forward mechanism. Broadly, these methods can be categorized into two types. The first category maps extracted features directly into textual embeddings~\cite{gal2023encoderbased,xiao2023fastcomposer,ostashev2024moa}. For instance, PhotoMaker~\cite{li2023photomaker} enhances identity by encoding multiple identity images into a stacked textual embedding. The second category employs additional cross-attention layers to integrate the extracted features into diffusion models\cite{pulid,lcm,wei2023elite,wang2024instantid,nested}, exemplified by methods like IP-Adapter. InstantID~\cite{wang2024instantid} further extends the IP-Adapter framework by integrating ControlNet, achieving superior identity preservation. While InstantID~\cite{wang2024instantid} adopts a dual-branch architecture that combines both textual embeddings and cross-attention injection, it fails to effectively integrate information from these two branches, resulting in suboptimal performance in both identity preservation and text controllability. Additionally, some studies~\cite{pulid,lcm} leverage fast-sampling techniques to obtain clean images, enabling the use of identity-preserving losses. Despite their effectiveness, methods that introduce new cross-attention layers often face challenges in maintaining strong adherence to original textual prompts, thus requiring carefully designed loss functions or specialized training datasets to balance identity preservation and text controllability. In contrast, encoding subjects directly into textual embeddings achieves superior text controllability but commonly results in diminished identity preservation.
 \section{Preliminaries}

\paragraph{Diffusion Models.}
Diffusion models constitute a powerful class of generative models producing high-quality images via iterative denoising. These models typically comprise a forward diffusion phase, progressively perturbing original images with Gaussian noise based on a defined noise schedule, and a reverse denoising phase, where a denoising network is trained to reconstruct the original image from its noisy variant. Formally, the denoising objective seeks to minimize the discrepancy between actual noise and predicted noise, defined as:
\begin{align}
  \mathcal{L}_{\text{diffusion}}=\mathbb{E}_{x, c, \varepsilon, t}\left[\left\|\varepsilon-\varepsilon_{\theta}\left(x_{t}, t, c\right)\right\|_{2}^{2}\right],
  \label{eq:diffusion_loss}
\end{align}
where the denoising network $\varepsilon_{\theta}$ is tasked with recovering the original image $x_0$ from its noisy variant $x_t$, given a specific timestep $t$ and the conditioning vector $c$. 

Our model is built upon Stable Diffusion XL~\cite{sdxl}, a type of Latent Diffusion Models~\citep{ldm}. The model is trained on a lower-dimensional latent space produced by an autoencoder, with text conditioning from two separate text encoders.

\paragraph{IP-Adapter.}
IP-Adapter~\cite{ipa} introduces an effective, tuning-free method to integrate visual information into pre-trained text-to-image diffusion models. It operates by embedding additional image features into the model without altering its pre-trained parameters. This is achieved by adding new cross-attention layers alongside the original cross-attention layers. Image features, typically extracted via an image encoder, are projected into separate key-value representations for cross-attention operations. Formally, this augmentation can be defined as:
\begin{align}
Z_\text{new} = \text{Attention}(Q,K,V) + \lambda \cdot \text{Attention}(Q,K',V'),
\label{eq:ipa}
\end{align}
where $Q$, $K$, and $V$ denote the query, key, and values matrices of the original cross-attention, $K'$ and $V'$ denote trainable projections of visual features, and $\lambda$ controls the strength of the new cross-attention.
 \begin{figure*}[t]
    \centering
    \includegraphics[width=1\textwidth]{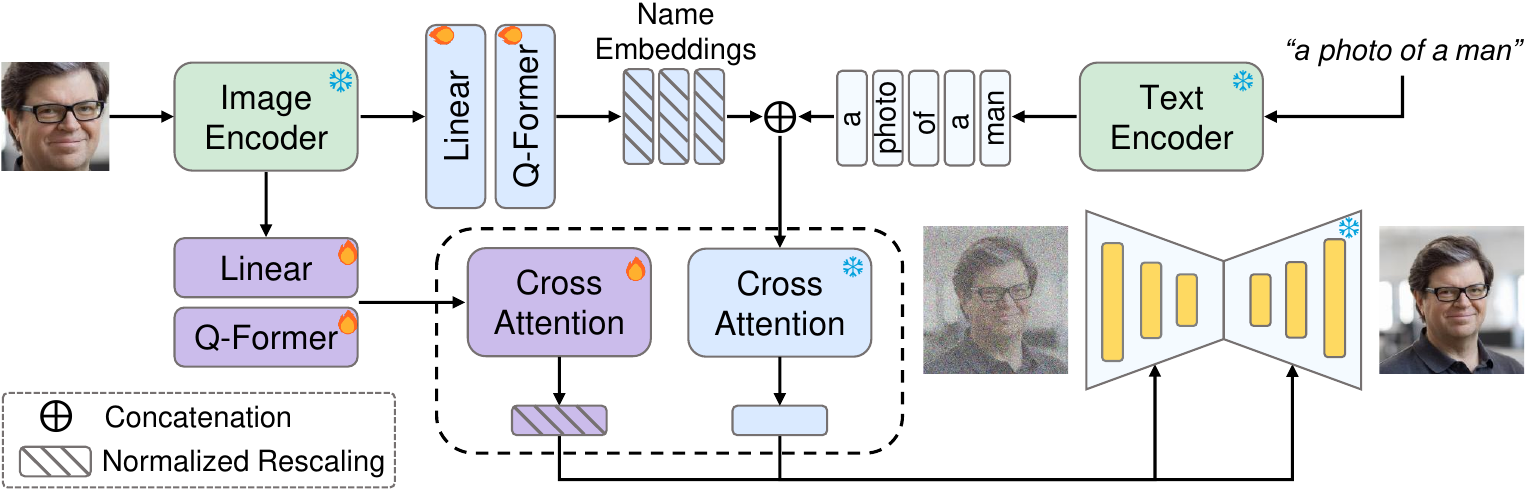} 
    \caption{\textbf{Overview of UniID.} (Top) We map the facial features extracted by the image encoder into the output embeddings of the text encoder. The predicted embeddings are concatenated with those of the given prompt. (Bottom) The extracted facial features are also injected into the pre-trained diffusion model via auxiliary cross-attention layers. At inference time, we apply the proposed normalized rescaling strategy to both branches to recover the text controllability of the original diffusion model.}
    \label{fig:framework} 
    \vspace{-10pt}
\end{figure*}
\section{Method}

Our objective is to synergize text embedding and adapter approaches to enhance identity fidelity while preserving text controllability. Our method achieves this integration through a key design principle: when merging the two branches (see our dual-branch architecture in Section~\ref{sec:architecture}), they should mutually reinforce identity information exclusively, while compositional generation remains governed by the diffusion model's prior knowledge. We realize this principle through a strategic training-inference paradigm. During training (Section~\ref{sec:training}), we employ an identity-focused learning scheme that ensures both branches capture identity-relevant features while deliberately avoiding the learning of scene composition, pose, or other non-identity attributes. At inference (Section~\ref{sec:inference}), we enable identity signals from both branches to mutually reinforce each other while preserving the original model's text controllability.

\subsection{Dual-Branch Architecture}
\label{sec:architecture}
Figure~\ref{fig:framework} illustrates our dual-branch architecture. The input image is first fed into a pre-trained face recognition model~\cite{an2021partial} to extract identity-relevant facial features. Then, these features are mapped to two branches: the text embedding branch and the adapter branch. This mapping is performed through a linear layer followed by a Q-Former~\cite{q_former}, as the learnable queries in Q-Former is able to capture distinct facial features~\cite{nested,UniPortrait}. In general, the text embedding branch maps the facial features into learnable name embeddings, while the adapter branch injects facial features into the diffusion model through auxiliary cross-attention layers.

\vspace{5pt}
\noindent\textbf{Text Embedding Branch.}
Unlike previous approaches~\cite{textual-inversion,gal2023encoderbased,basis} that typically inject personalized representations through the input text embeddings of the text encoder, we propose to map facial features directly to the \textit{output embeddings} of the text encoder. The reason for this design choice is that output embeddings provide significantly higher expressiveness than input embeddings~\cite{neti}. Although output embeddings are known to provide weaker text controllability~\cite{neti}, we address this limitation through a rescaling strategy applied during inference (Section~\ref{sec:inference}), which leverages the disentanglement properties of name embeddings discovered in~\cite{MagicNaming}. Through empirical analysis, we find that mapping facial features to a sequence of three token embeddings, corresponding to the target individual's full name, achieves optimal performance. These predicted name embeddings are then concatenated with the remaining token embeddings from the text prompt.

\vspace{5pt}
\noindent\textbf{Adapter Branch.}
For the adapter branch, we largely adopt the design of IP-Adapter~\cite{ipa}, where facial features are injected into the diffusion model through auxiliary cross-attention layers parallel to the original text cross-attention layers. The key distinction in our approach lies in the use of the Q-Former to map facial features, as Q-Former has been proven to capture more discriminative facial information~\cite{UniPortrait,nested}.

\subsection{Training for Identity Preservation}
\label{sec:training}
As discussed previously, our training objective is to exclusively maximize identity learning for each branch while deliberately avoiding the learning of non-identity attributes such as scene composition or background elements. This focused learning strategy ensures that when the two branches are merged during inference, only identity-relevant features mutually reinforce each other, without introducing conflicting information about non-identity elements, which remain governed by the diffusion model's prior knowledge.

Based on this insight, our training strategy comprises two key designs: 1) We separately train each branch using the standard diffusion loss (Eq.~\ref{eq:diffusion_loss}), thereby maximizing the identity preservation of each branch. 2) Identity-focused learning is achieved by training each branch on portrait images (where faces occupy most of the image space) with minimal prompts (``a photo of a man/woman''). This training paradigm fundamentally differs from prior work \cite{li2023photomaker,pulid,nested}, which must simultaneously optimize for both identity fidelity and text controllability during training, consequently requiring large-scale datasets of diverse in-the-wild images paired with compositional text prompts. By decoupling identity learning from scene composition, our approach achieves superior identity preservation while significantly reducing data requirements.

\subsection{Inference for Controllability}
\label{sec:inference}
At inference time, our objective is to integrate the identity information learned by each branch while preserving the text controllability of the original diffusion model. However, since both branches are trained on simple portrait images with minimal text prompts, directly using either branch for personalized generation produces nearly exact reconstructions of the input images while disregarding the given text prompts, as illustrated in Figure~\ref{fig:rescaling_weight}. This occurs because the newly introduced representations (i.e., the new cross-attention outputs or predicted text embeddings) dominate the generation process, overfitting to the training portraits and suppressing the model's ability to respond to novel text prompts.

\begin{figure}[t]
    \centering
    \begin{minipage}{0.47\linewidth}
        \centering
        \includegraphics[width=\textwidth]{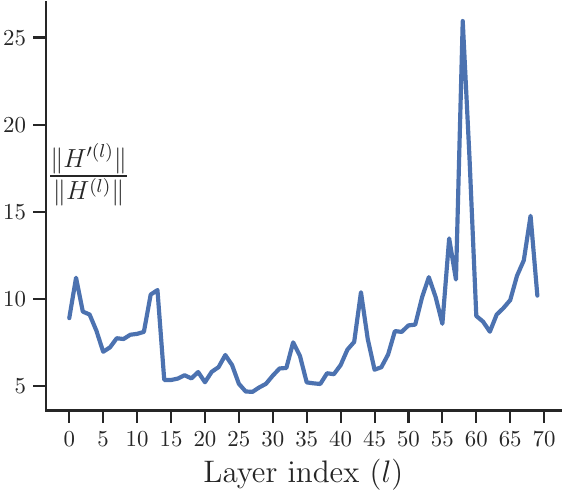}
        \caption{Layer-wise output magnitude ratios.}
        \label{fig:ratio}
    \end{minipage}
    \hspace{5pt}
    \begin{minipage}{0.47\linewidth}
        \centering
        \vspace{0.75cm}
        \includegraphics[width=\textwidth]{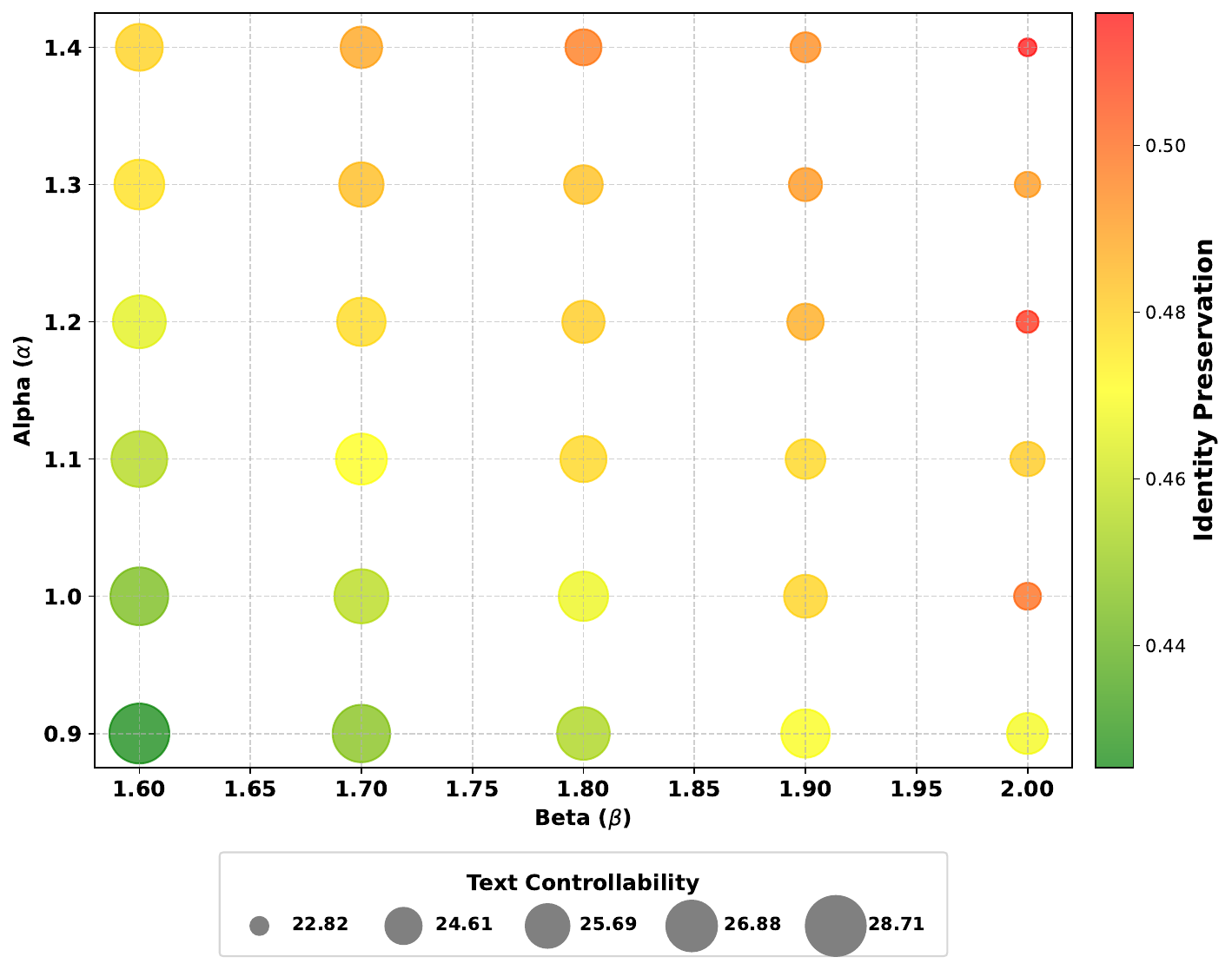}
        \caption{Grid search results for hyperparameters $\alpha$ and $\beta$. Zoom in for a better view.}
        \label{fig:heatmap}
    \end{minipage}
    \vspace{-8pt}
\end{figure}

Through systematic investigation, we identify that this domination manifests through dramatically inflated magnitudes of these new representations. Figure~\ref{fig:ratio} visualizes the ratios between the output norms of new cross-attention layers and their corresponding original layers. The new cross-attention outputs exhibit significantly larger norms, reaching up to 26$\times$ the magnitude of the original layers. Notably, the peak occurs around Layer 58, which aligns with findings from prior work~\cite{blora} indicating that layers in this region are particularly effective at capturing image content. Similarly, in the text embedding branch, the predicted name embeddings demonstrate norms approximately 4$\times$ larger than those of the original name tokens in the vocabulary.

A straightforward solution to mitigate these inflated magnitudes is to apply rescaling. Existing adapter-based methods~\cite{ipa} typically employ a global rescaling weight uniformly across all layers, as illustrated in Eq.~\ref{eq:ipa}. However, as shown in Figure~\ref{fig:ratio}, the magnitude variations across layers are highly non-uniform in our scenarios. Applying a uniform weight leads to a fundamental dilemma: layers with extreme magnitudes (e.g., Layer 58) remain over-dominant even after rescaling, while layers with moderate magnitudes become excessively suppressed. 

\vspace{5pt}
\noindent\textbf{Layer-wise Normalized Rescaling.}
To address this limitation, we propose a layer-wise normalized rescaling strategy that adapts to the magnitude distribution of each layer. The key insight is to first normalize each new layer's output to unit norm, then rescale it proportionally to its corresponding original layer's magnitude, before applying a global weight. For the adapter branch, this is formalized as:
\begin{align}
Z_\text{new}^{(l)} = H^{(l)} + \alpha \cdot \frac{H'^{(l)}}{\lVert H'^{(l)} \rVert} \cdot \lVert H^{(l)} \rVert,
\label{eq:rescaling_adapter}
\end{align}
where $H^{(l)}$ and $H'^{(l)}$ denote the outputs of the original and new cross-attention layers at layer $l$, respectively, and $\alpha$ is the global rescaling weight. This layer-wise normalization ensures that each new layer's contribution is proportional to the magnitude of its corresponding original layer, preventing any single layer from dominating while maintaining balanced influence across the architecture. 

Similarly, we apply this normalized rescaling principle to the text embedding branch:
\begin{align}
e^* = \beta \frac{e'}{\lVert e' \rVert} \lVert \bar{e} \rVert,
\label{eq:rescaling_textual}
\end{align}
where $e'$ represents the predicted name embedding, $\bar{e}$ denotes the average embedding computed from hundreds of common names in the vocabulary, and $\beta$ is the rescaling weight. This normalization aligns the predicted embeddings with the typical magnitude of name tokens, enabling seamless integration with other textual tokens during inference.

Notably, previous text embedding approaches~\cite{MagicNaming,neti} typically perform rescaling during training to balance identity preservation and text controllability. In contrast, we focus exclusively on identity preservation during training (Section~\ref{sec:training}), and thus do not apply any rescaling operations. Instead, the proposed normalized rescaling strategy is applied at the inference stage to restore the pre-trained diffusion model's text controllability. When merging the two branches, the complementary identity signals from both branches are combined, enabling strong identity preservation while maintaining the diffusion model's text controllability.

\subsubsection{Determining Rescaling Weights}
As shown in Figure~\ref{fig:appendix_rescaling_weight} (Appendix), fusing our dual branches achieves effective identity reinforcement while maintaining the text controllability of the weaker individual branch. To determine optimal values for $\alpha$ and $\beta$, we perform a grid search over a broad range of values. Figure~\ref{fig:heatmap} visualizes the results, revealing parameter regions that yield strong performance. We select values that achieve the best trade-off between identity preservation and text controllability.
\begin{figure*}
    \centering
    \includegraphics[width=1\textwidth]{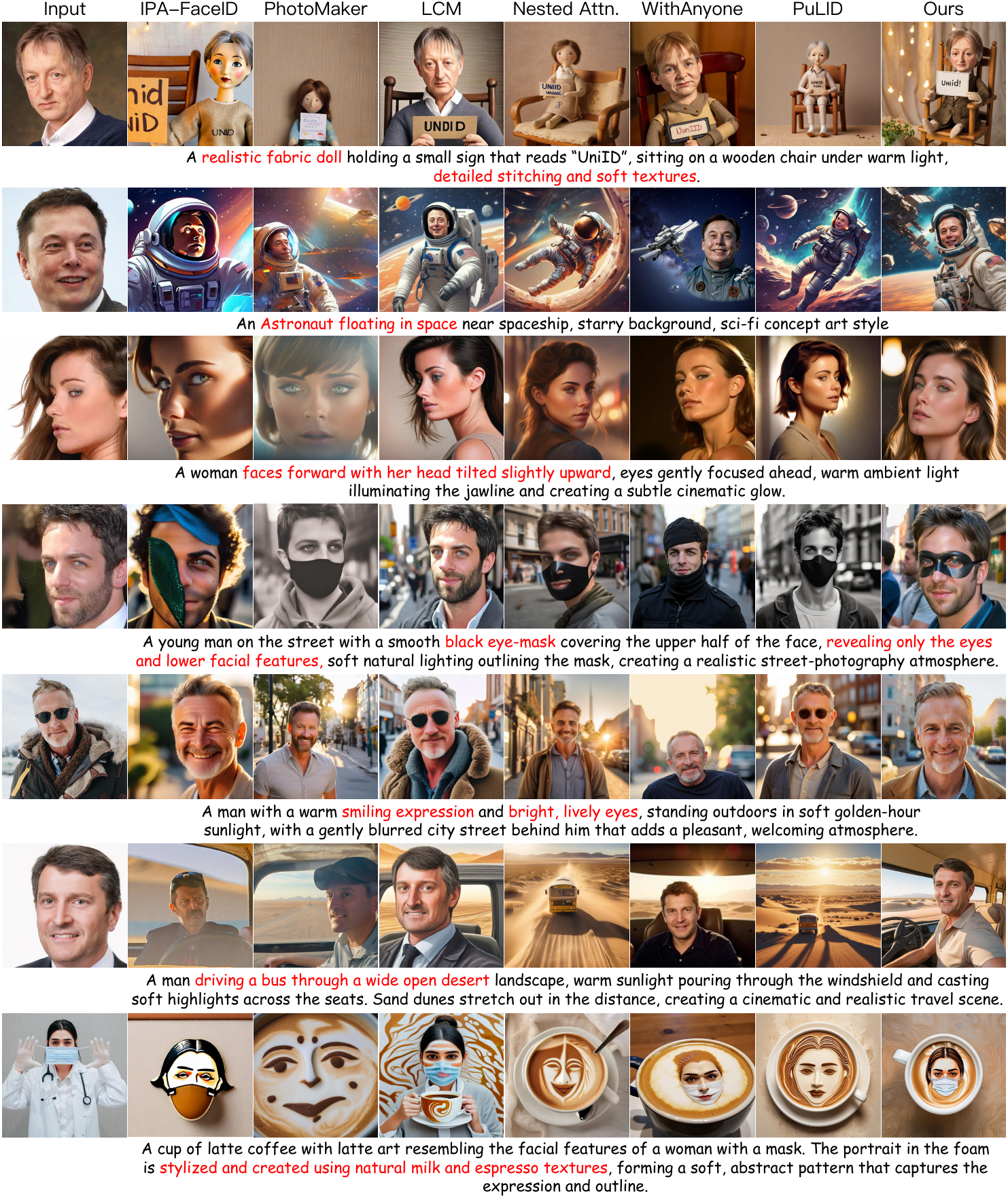} 
    \caption{\textbf{Qualitative comparison}. We compare our method with six baseline methods, including IPA-FaceID~\cite{ipa}, PhotoMaker~\cite{li2023photomaker}, LCM~\cite{lcm}, Nested Attention~\cite{nested}, WithAnyone~\cite{xu2025withanyone}, and PuLID~\cite{pulid}. Our method demonstrates superior performance in identity preservation and text controllability compared to these baselines. Please zoom in for a better view.}
    \label{fig:qualitative_comparison} 
\end{figure*}

\section{Experiments}
\subsection{Implementation and Evaluation Setup}
\label{sec:implementation}
\paragraph{Implementation Details.}
We build our implementation on Stable Diffusion XL~\cite{sdxl}. Our model is trained on approximately $500,000$ images from three datasets: FFHQ-Portrait~\cite{stylegan}, CelebA-HQ~\cite{celeba}, and filtered FaceID-6M~\cite{wang2025faceid}. Facial features are extracted using a face recognition model~\cite{an2021partial}. The text embedding branch employs a 4-layer Q-Former and is trained for $12$ epochs with a learning rate of $10^{-4}$ and batch size of $92$. The adapter branch utilizes a 6-layer Q-Former and is trained for $16$ epochs with a learning rate of $10^{-5}$ and batch size of $76$. All training is conducted on four NVIDIA A100 80GB GPUs using the AdamW optimizer. At inference time, we employ $30$ denoising steps with rescaling weights $\alpha=1.2$ and $\beta=1.8$. For all baselines, we use their official implementations with default hyperparameters.

\begin{figure}[htbp]
    \centering
    \begin{minipage}{0.49\linewidth}
        \centering
        \includegraphics[width=\textwidth]{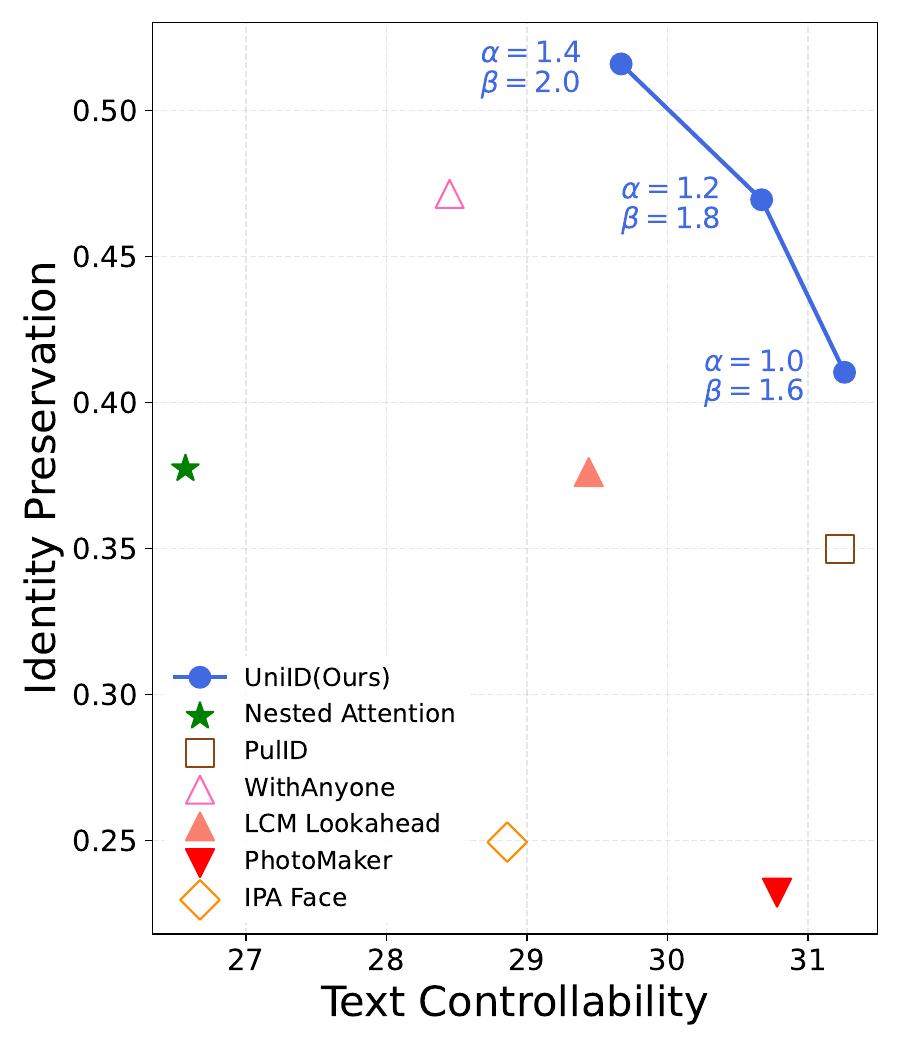}
    \end{minipage}
    \hspace{0\linewidth}
    \begin{minipage}{0.49\linewidth}
        \centering
        \includegraphics[width=\textwidth]{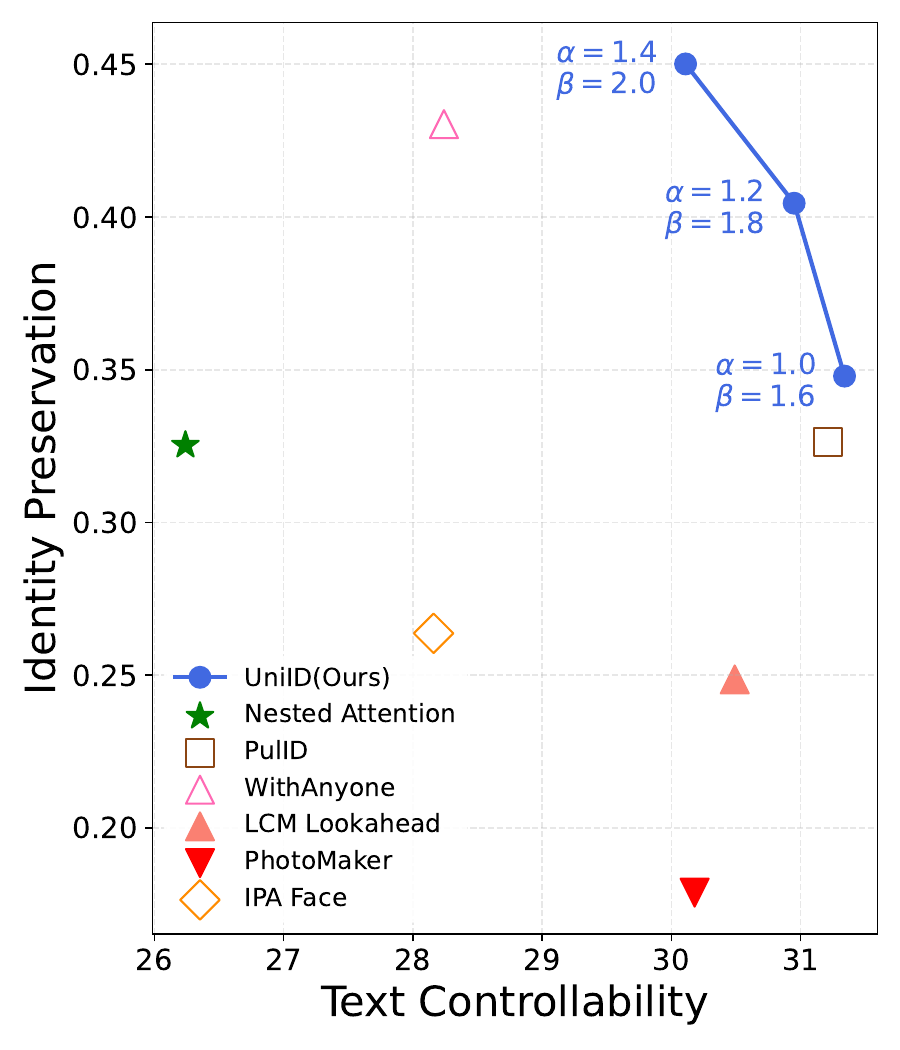}
    \end{minipage}
    \caption{\textbf{Quantitative comparisons.} We evaluate all methods on two test sets: 50 synthetic portraits generated by StyleGAN3 (left) and 50 real portrait photographs from Unsplash (right). Our method achieves superior performance across both identity preservation and text controllability metrics.}
    \label{fig:quantitative_comparison}
    \vspace{-0.5cm}
\end{figure}

\vspace{5pt}
\noindent\textbf{Evaluation Setup.}
We compare our method against six state-of-the-art face personalization approaches: IP-Adapter-FaceID~\cite{ipa}, PhotoMaker~\cite{li2023photomaker}, LCM~\cite{lcm}, Nested Attention~\cite{nested}, PuLID~\cite{pulid}, and WithAnyone~\cite{xu2025withanyone}. It is important to note that these methods employ different base models as their backbone architectures. Specifically, PuLID builds upon SDXL-Lightning~\cite{sdxl-lightning}, LCM utilizes SDXL-Turbo~\cite{sdxl-turbo}, and WithAnyone employs the more powerful FLUX model~\cite{flux}. To ensure competitive visual quality in our main comparisons, we adopt RealVisXL as our generation backbone. For a fair comparison under identical conditions, we provide additional results where all methods use the standard SDXL backbone in Appendix~\ref{sec:appendix_all_sdxl}. We also include a visual comparison to the state-of-the-art image editing model FLUX.1 Kontext~\cite{kontext} in Appendix~\ref{sec:appendix_kontext}.
For quantitative evaluation, we deliberately avoid using celebrity portraits, as such images are highly likely to appear in the training datasets of existing models. Instead, our test set consists of 100 identity images: 50 portrait photographs recently uploaded to Unsplash by individual users, and 50 synthetic faces generated using StyleGAN3~\cite{stylegan3}. This combination ensures diversity while minimizing potential data leakage. Each method is evaluated across 20 diverse text prompts. The complete list of prompts is provided in Appendix~\ref{sec:appendix_prompt}.

\subsection{Results}
\label{sec:results}

\paragraph{Qualitative Evaluation.}
Figure~\ref{fig:qualitative_comparison} presents a visual comparison between our method and baseline approaches. As shown, IP-Adapter-FaceID and PhotoMaker exhibit significant limitations in identity preservation, failing to maintain faithful facial features of the input reference. LCM and WithAnyone demonstrate limited text controllability, particularly when handling complex generation tasks such as adding occlusions or applying stylistic transformations. Furthermore, LCM produces outputs that appear blurry with noticeable artifacts, compromising overall image quality. Nested Attention similarly suffers from weak text-prompt alignment, especially in style transfer scenarios. PuLID shows inconsistent identity preservation, particularly when generating full-body images. Notably, both PuLID and Nested Attention occasionally fail to incorporate the reference face into the generated images (e.g., row 6).
In contrast, our method consistently generates high-quality results that effectively preserve identity while accurately aligning with the provided text prompts. Notably, for challenging examples such as occlusions at specific facial locations (row 4) and stylization of occluded faces (bottom row), our method is the only approach that successfully produces the desired personalized images. Additional qualitative results are provided in Appendix~\ref{sec:appendix_qualitative}.

\begin{table}
    \centering
    \caption{\textbf{User study results.} Participants were asked to select the image that best preserves the reference identity while accurately matching the text prompt.}
    \vspace{-3pt}
    \begin{tabular}{@{\hspace{0.6cm}}l@{\hspace{0.5cm}}c@{\hspace{0.6cm}}c@{\hspace{0.6cm}}}
      \toprule
      Methods & Baseline    & Ours  \\
      \midrule
      IPA-FaceID~\cite{ipa}                 & 15.0\%            & 85.0\%    \\
      PhotoMaker~\cite{li2023photomaker}    & 15.8\%            & 84.2\%    \\
      LCM~\cite{lcm}                        & 23.3\%            & 76.7\%    \\
      Nested Attention~\cite{wang2024instantid}    & 17.5\%     & 82.5\%    \\
      WithAnyone~\cite{wang2024instantid}    & 33.3\%            & 66.7\%    \\
      PuLID~\cite{pulid}                    & 28.3\%            & 71.7\%     \\
      \bottomrule
          \label{tab:user_study}
    \end{tabular}
    \vspace{-25pt}
\end{table}

\vspace{3pt}
\noindent\textbf{Quantitative Evaluation.}
We conduct quantitative evaluation using two metrics: identity preservation and text controllability. Identity preservation is measured by computing the cosine similarity between CurricularFace~\cite{huang2020curricularface} embeddings of the generated images and the reference portraits. Text controllability is evaluated using the cosine similarity between CLIP~\cite{clip} embeddings of the generated images and their corresponding text prompts. Figure~\ref{fig:quantitative_comparison} presents the quantitative comparison across all methods. The results align with our qualitative observations. IP-Adapter-FaceID and PhotoMaker achieve notably low identity preservation scores, indicating limitations in maintaining facial characteristics from the reference portraits. Conversely, Nested Attention and WithAnyone exhibit weaker text controllability, demonstrating difficulty in generating images that accurately reflect the input prompts. Among the baseline methods, PuLID and LCM demonstrate stronger performance, achieving a favorable balance between identity preservation and text controllability. However, our method achieves further improvements on both synthetic and real-world test sets, demonstrating superior identity preservation and text controllability.

\begin{figure*}
    \centering
    \renewcommand{\arraystretch}{0.3}
    \setlength{\tabcolsep}{1pt}

    {\small
    \begin{tabular}{cc c c c c c c c cc}
        Input && $\alpha=0.4$ & $\alpha=0.6$ & $\alpha=0.8$ & $\alpha=1.0$ & $\alpha=1.2$ & $\alpha=1.4$ & $\alpha=1.6$ & $\alpha=1.8$ &w/o Resc.\\
        \includegraphics[width=0.09\textwidth]{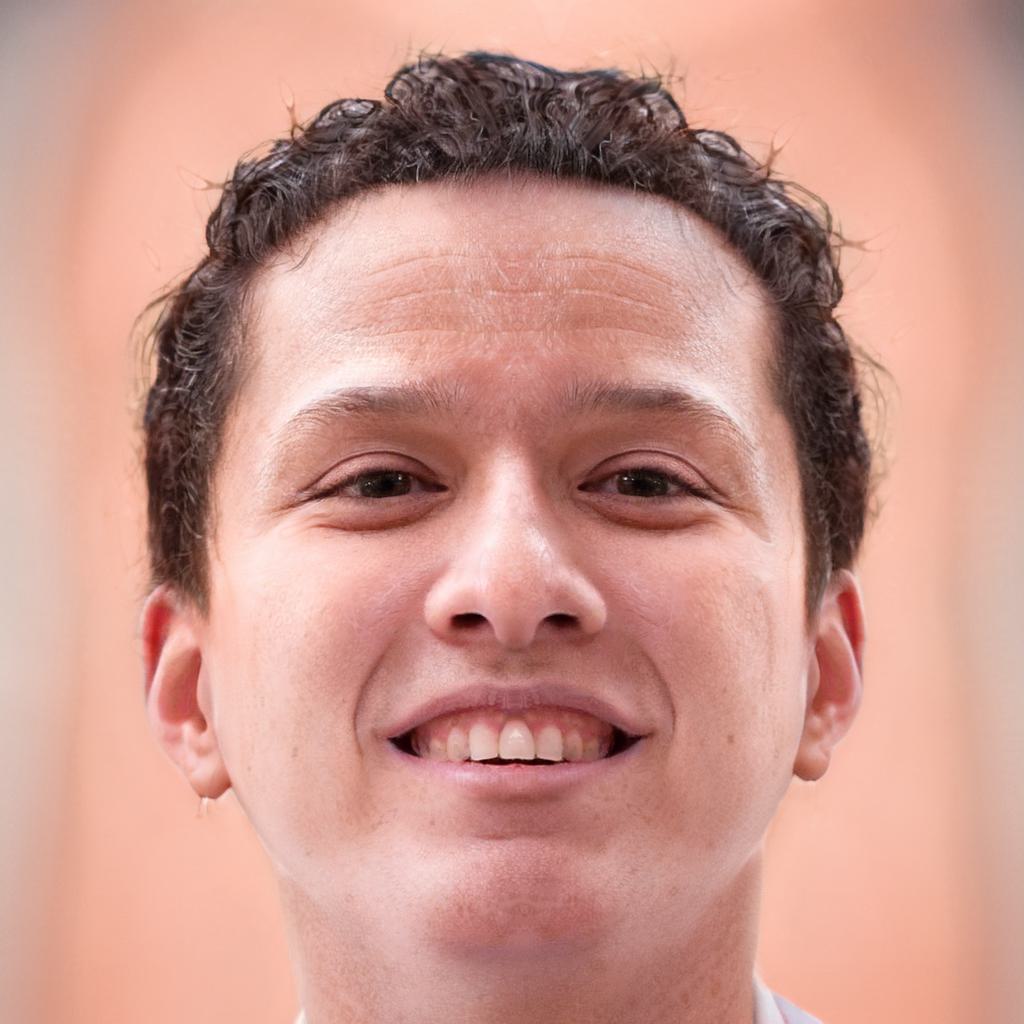} & 
        \raisebox{0.22in}{\rotatebox[origin=t]{90}{Adapter}}&
        \includegraphics[width=0.09\textwidth]{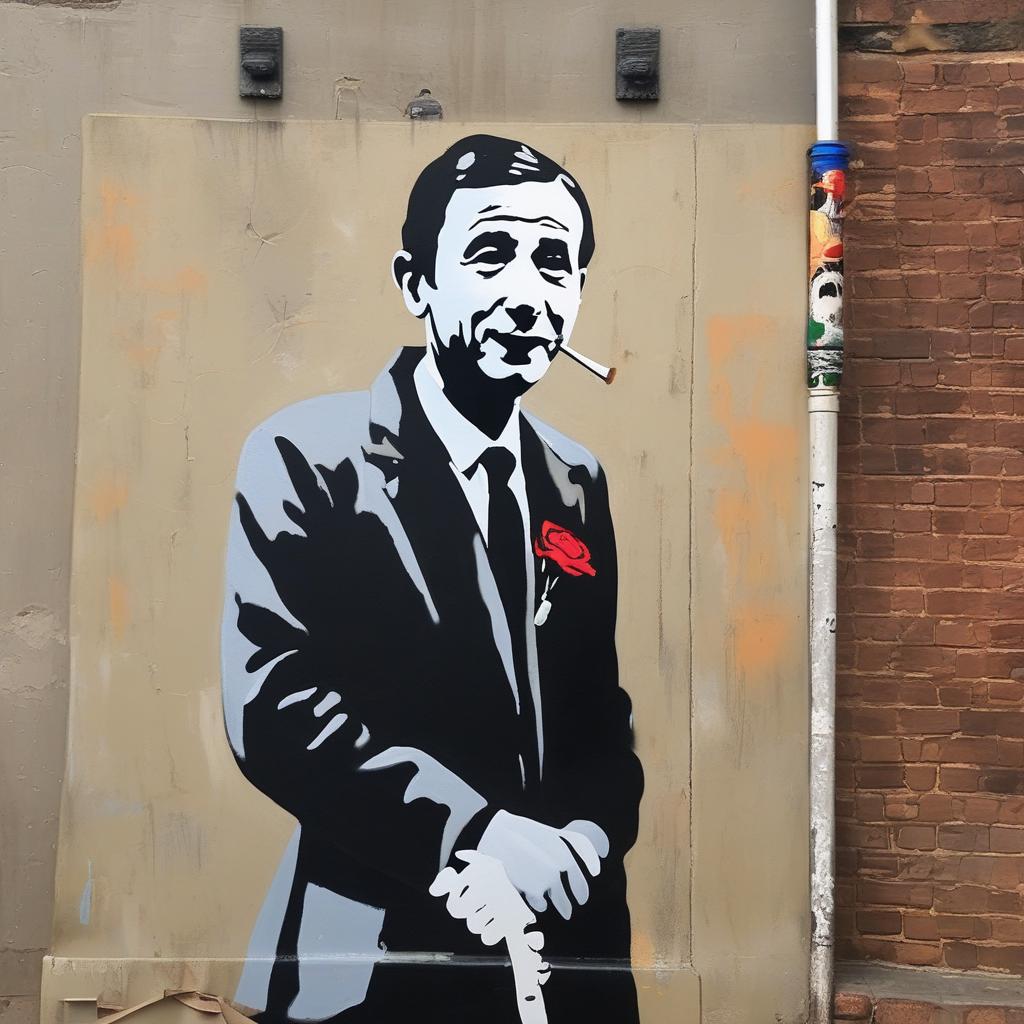} & 
        \includegraphics[width=0.09\textwidth]{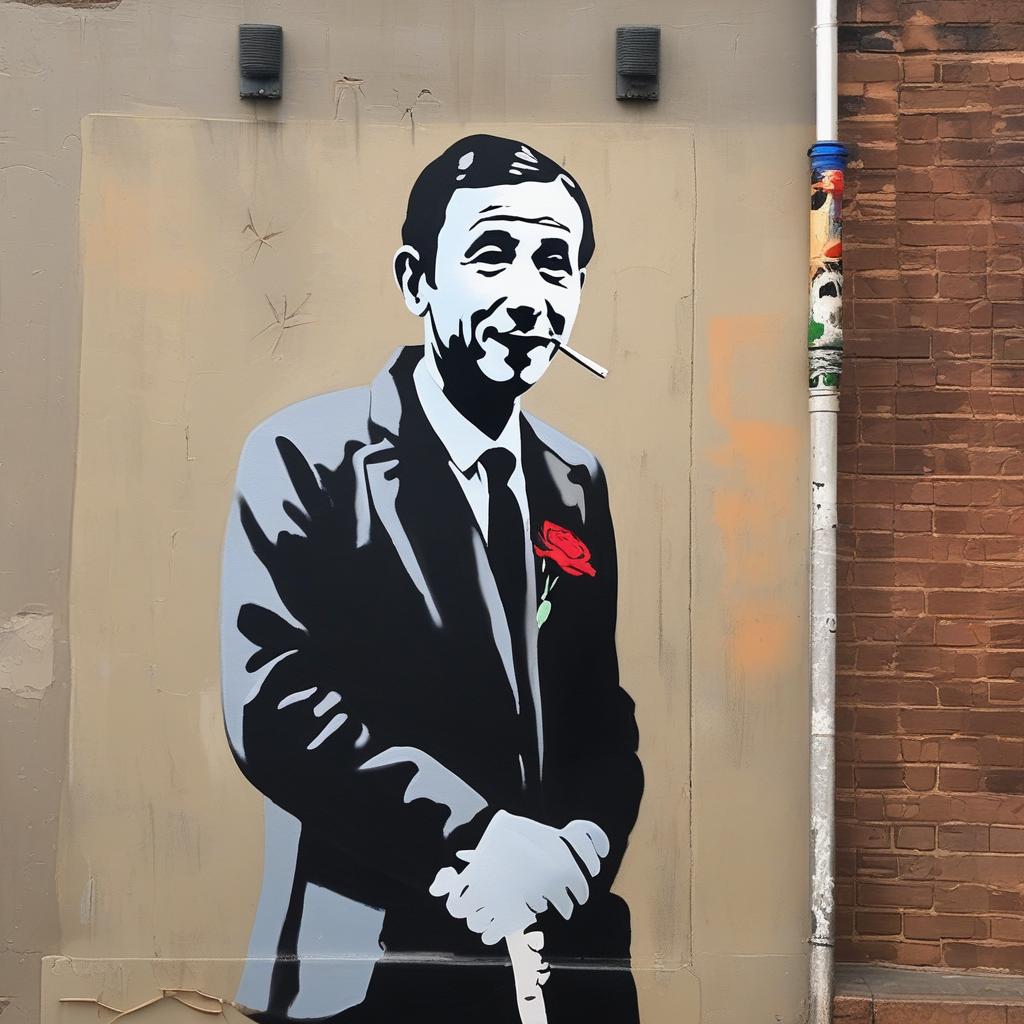} & 
        \includegraphics[width=0.09\textwidth]{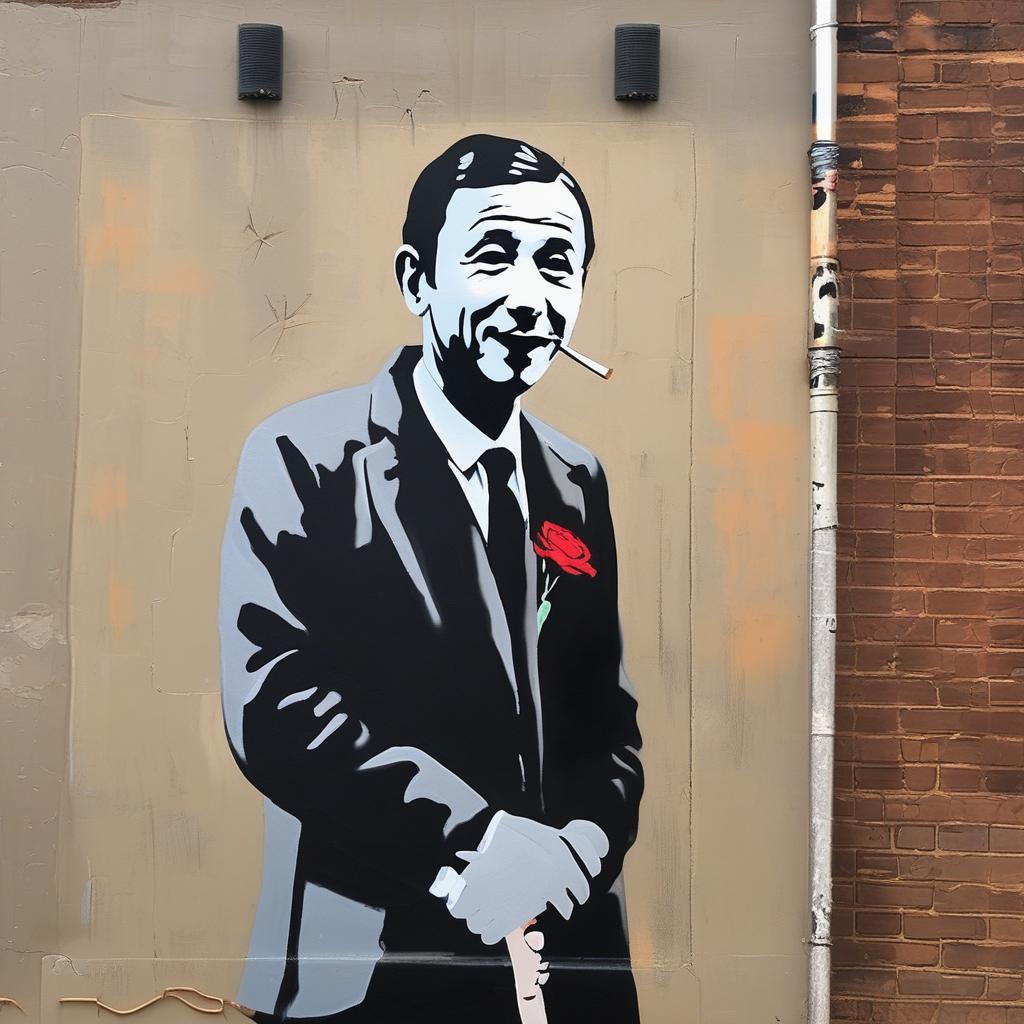} & 
        \includegraphics[width=0.09\textwidth]{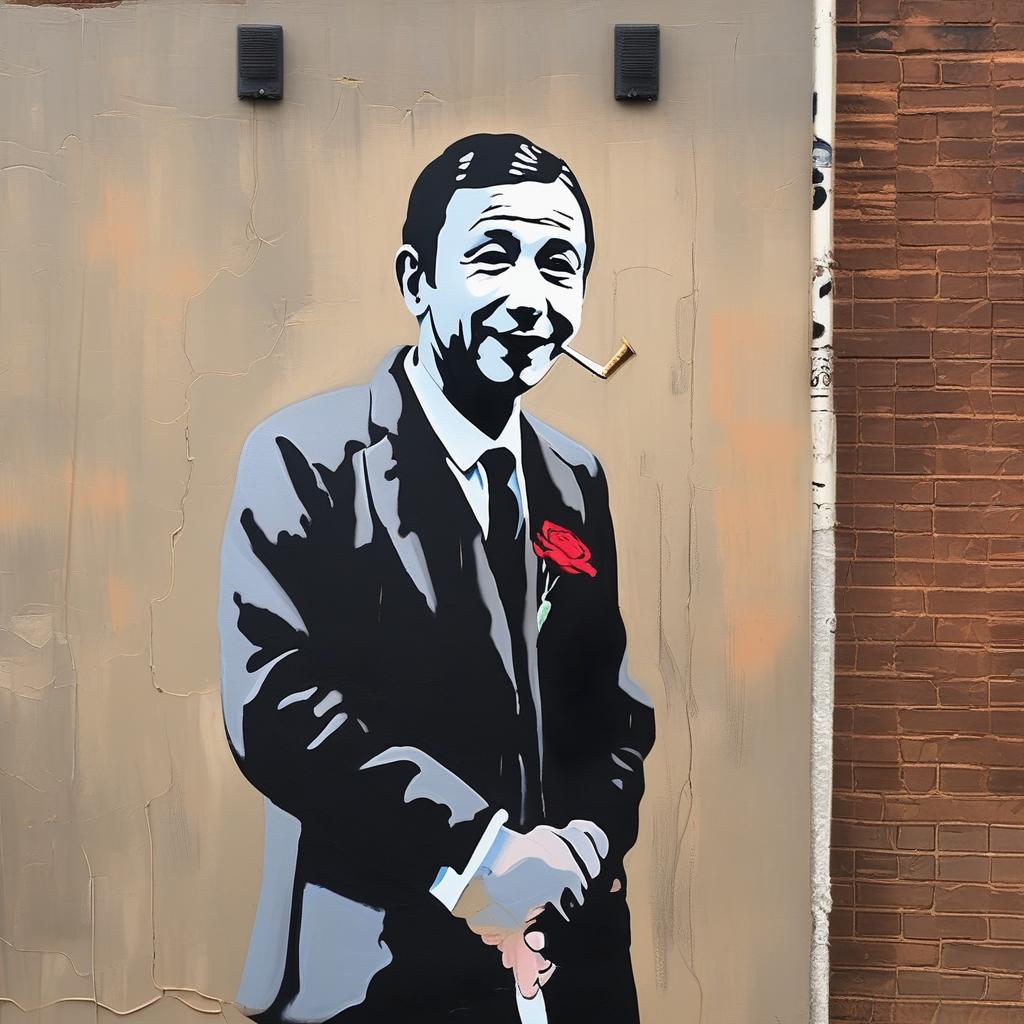} & 
        \includegraphics[width=0.09\textwidth]{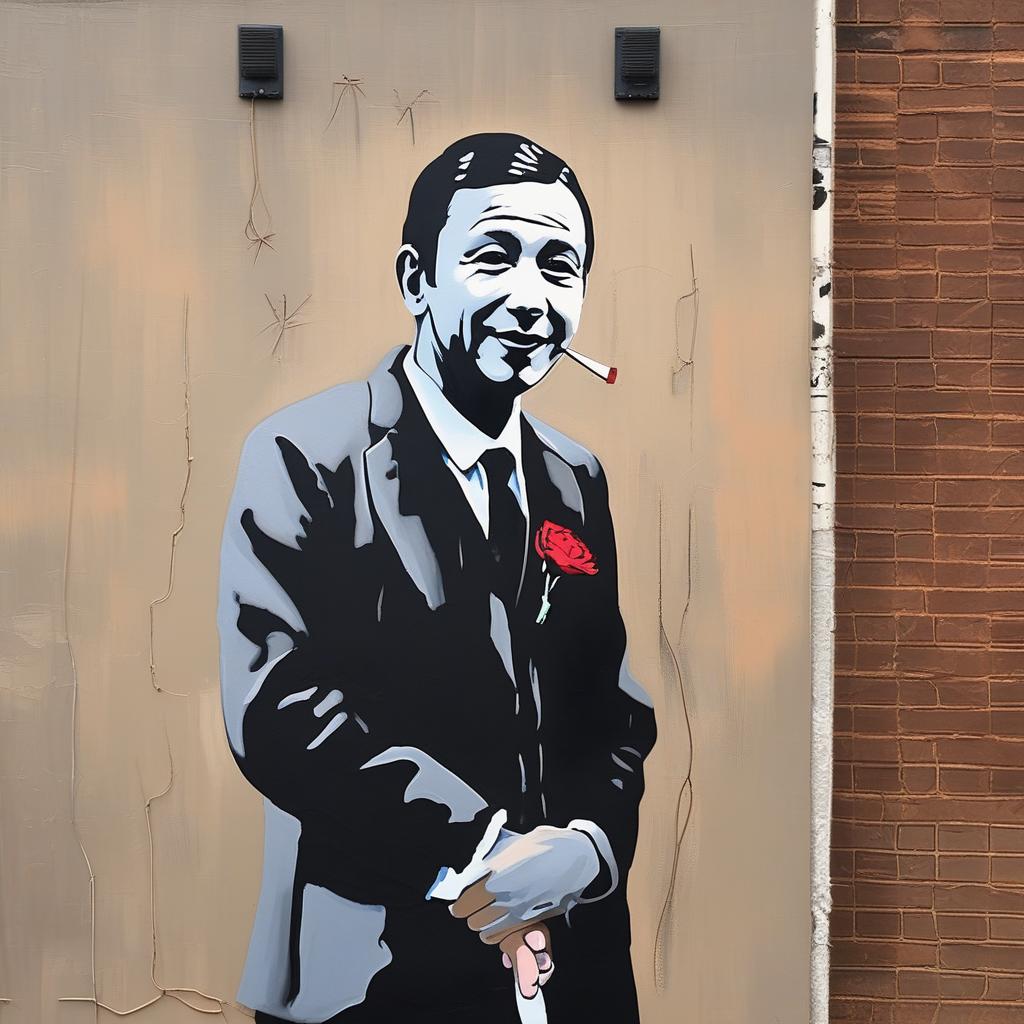} & 
        \includegraphics[width=0.09\textwidth]{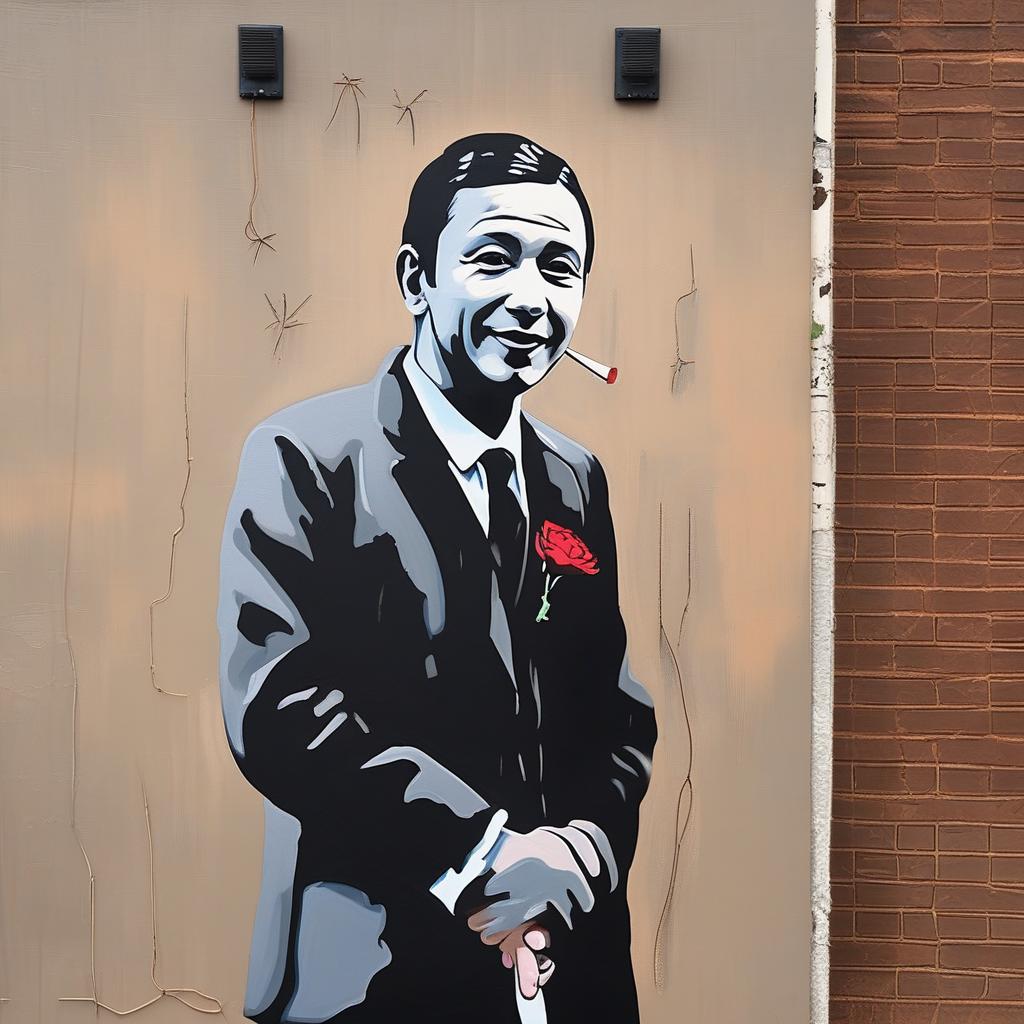} & 
        \includegraphics[width=0.09\textwidth]{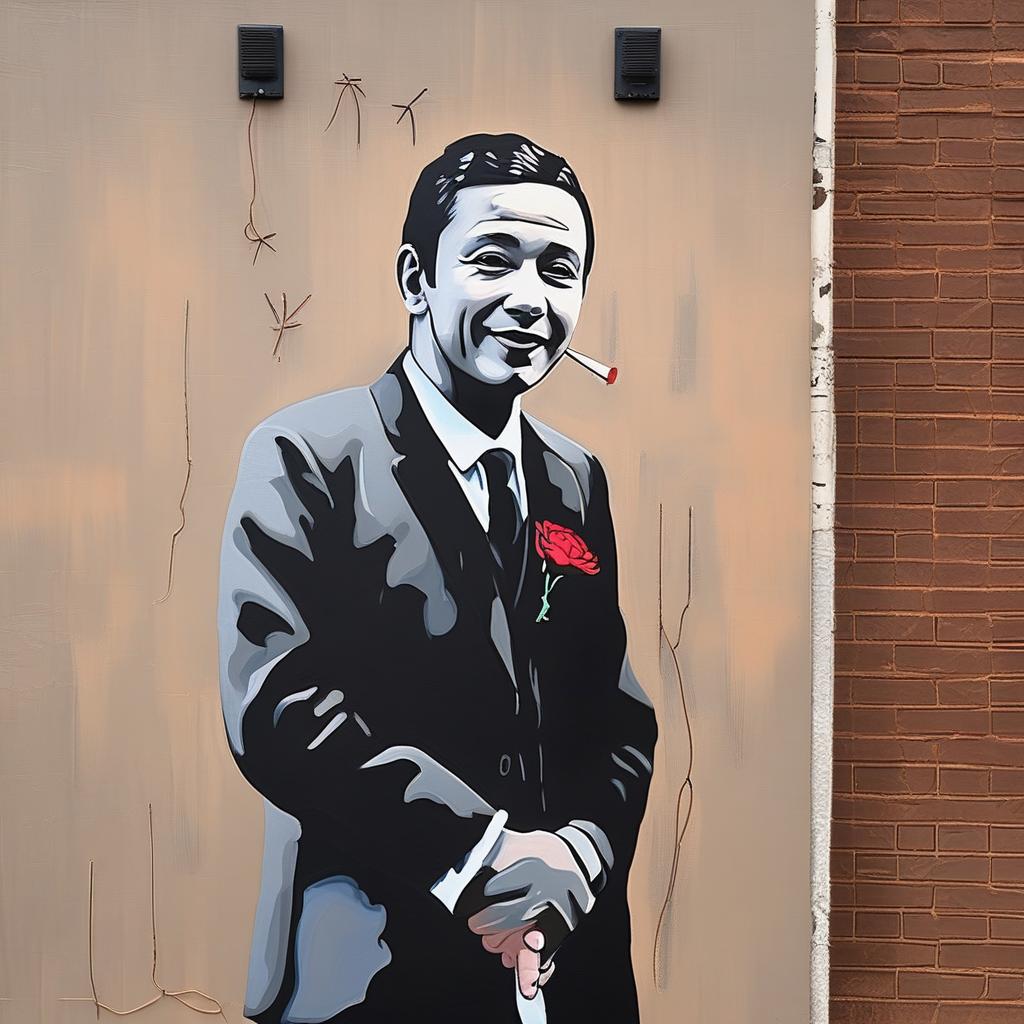} & 
        \includegraphics[width=0.09\textwidth]{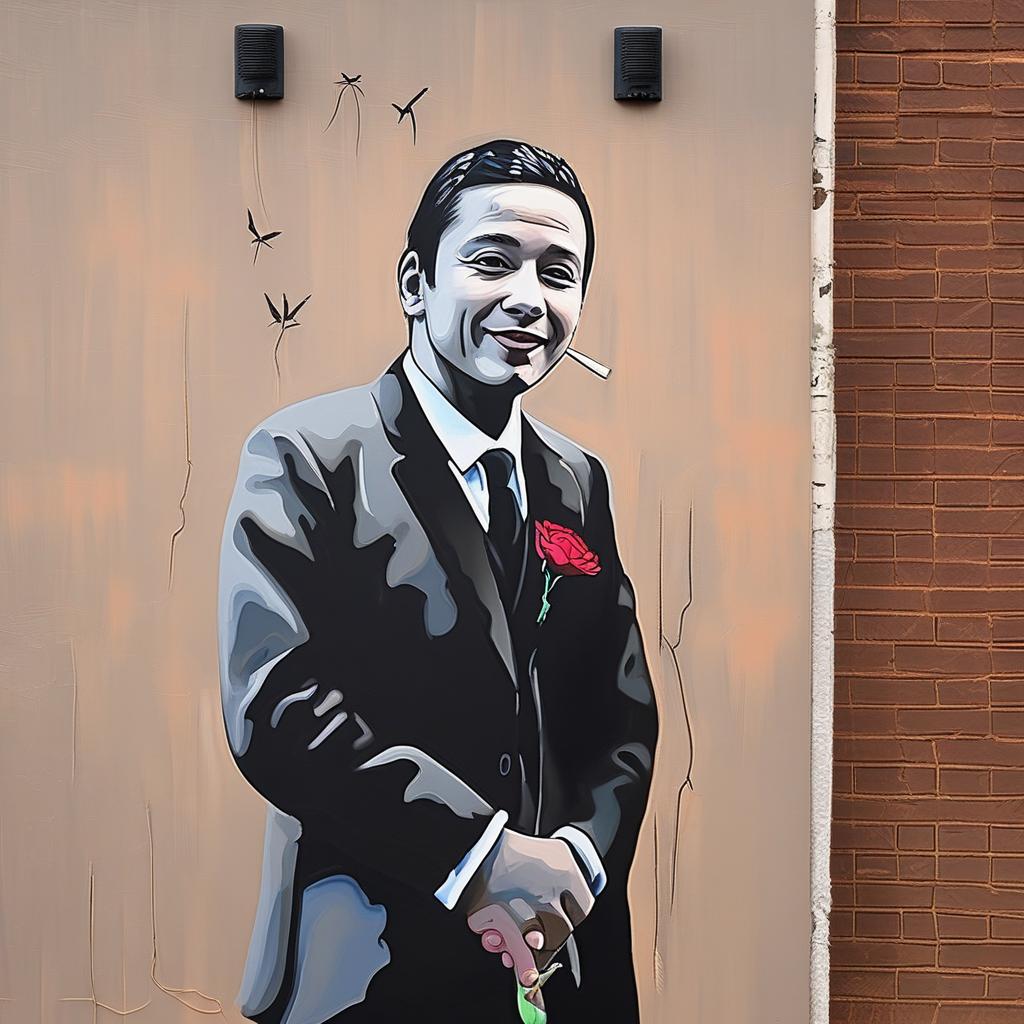} & 
        \includegraphics[width=0.09\textwidth]{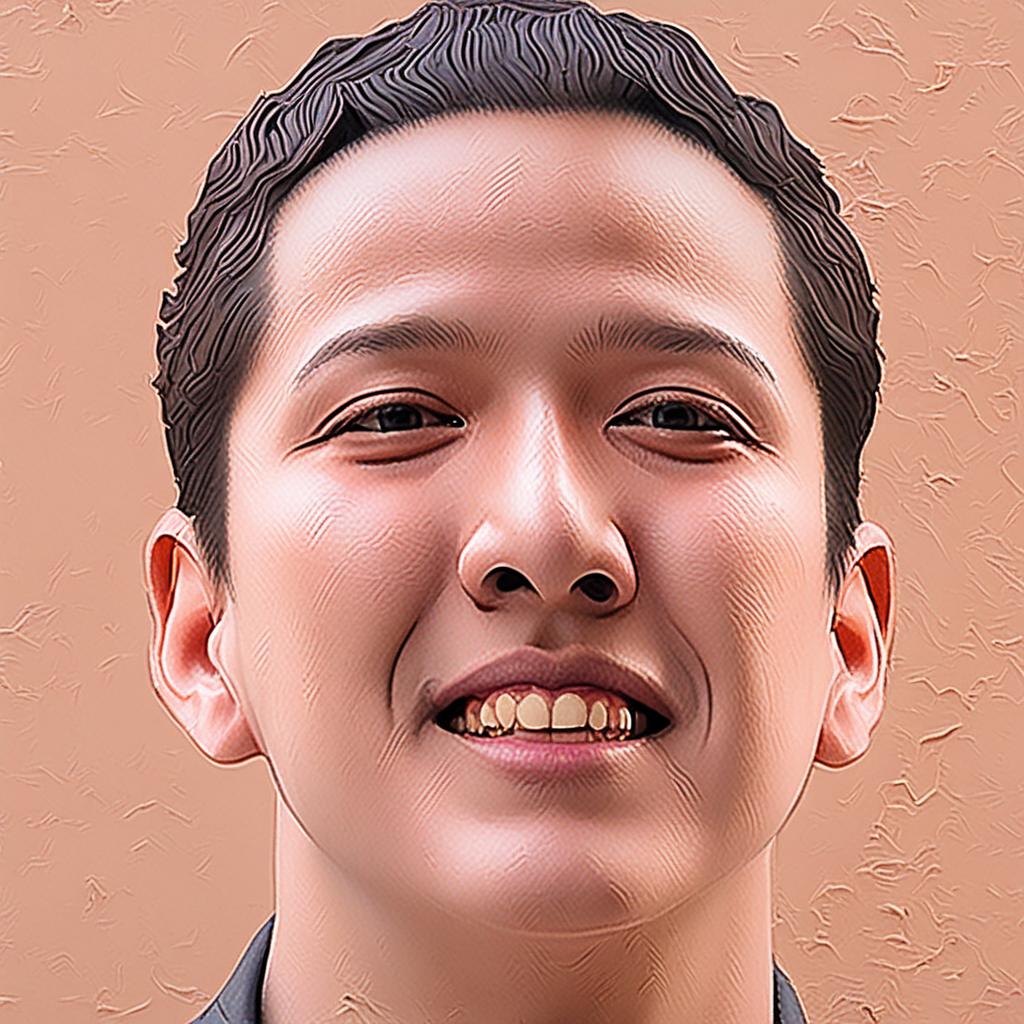} \\
        \quad && $\beta=1.0$ & $\beta=1.2$ & $\beta=1.4$ & $\beta=1.6$ & $\beta=1.8$ & $\beta=2.0$ & $\beta=2.2$ & $\beta=2.4$ &w/o Resc.\\
        &
        \raisebox{0.22in}{\rotatebox[origin=t]{90}{Text}}&
       \includegraphics[width=0.09\textwidth]{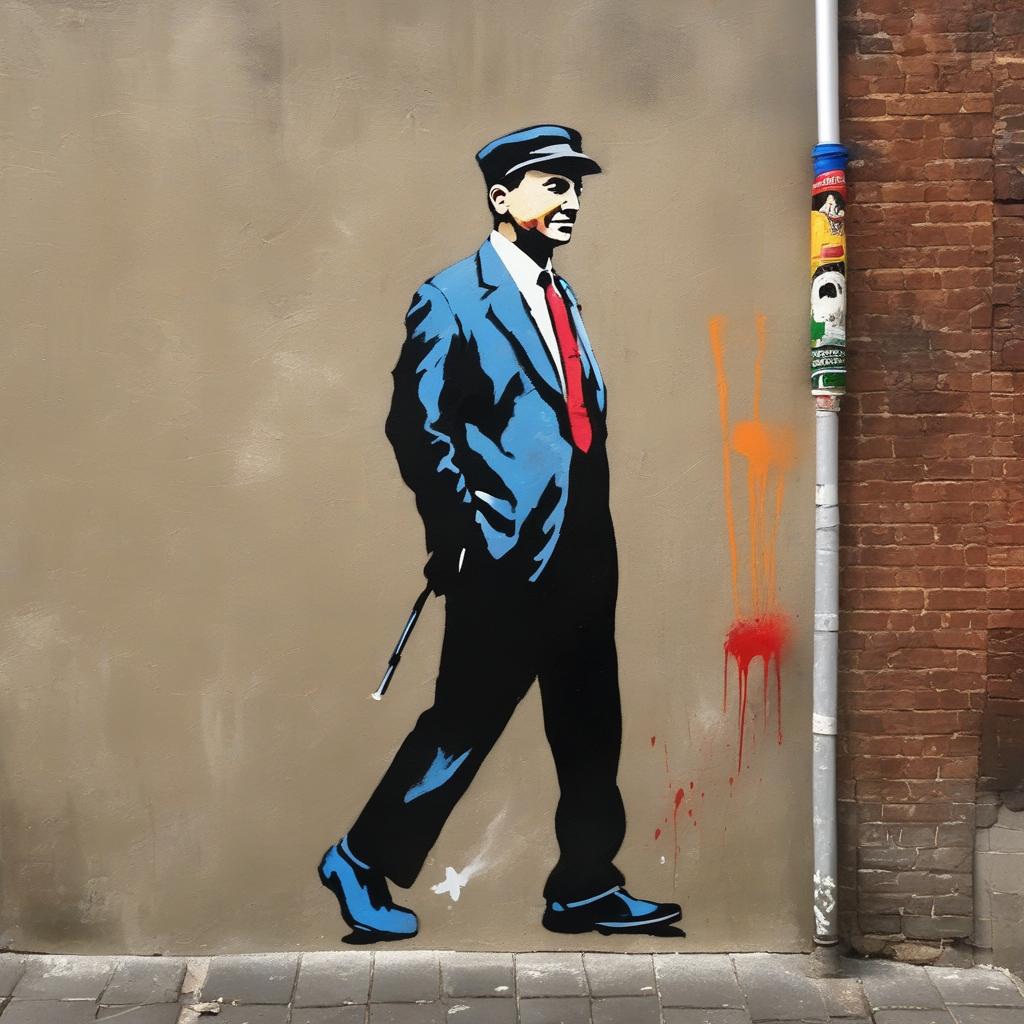} & 
        \includegraphics[width=0.09\textwidth]{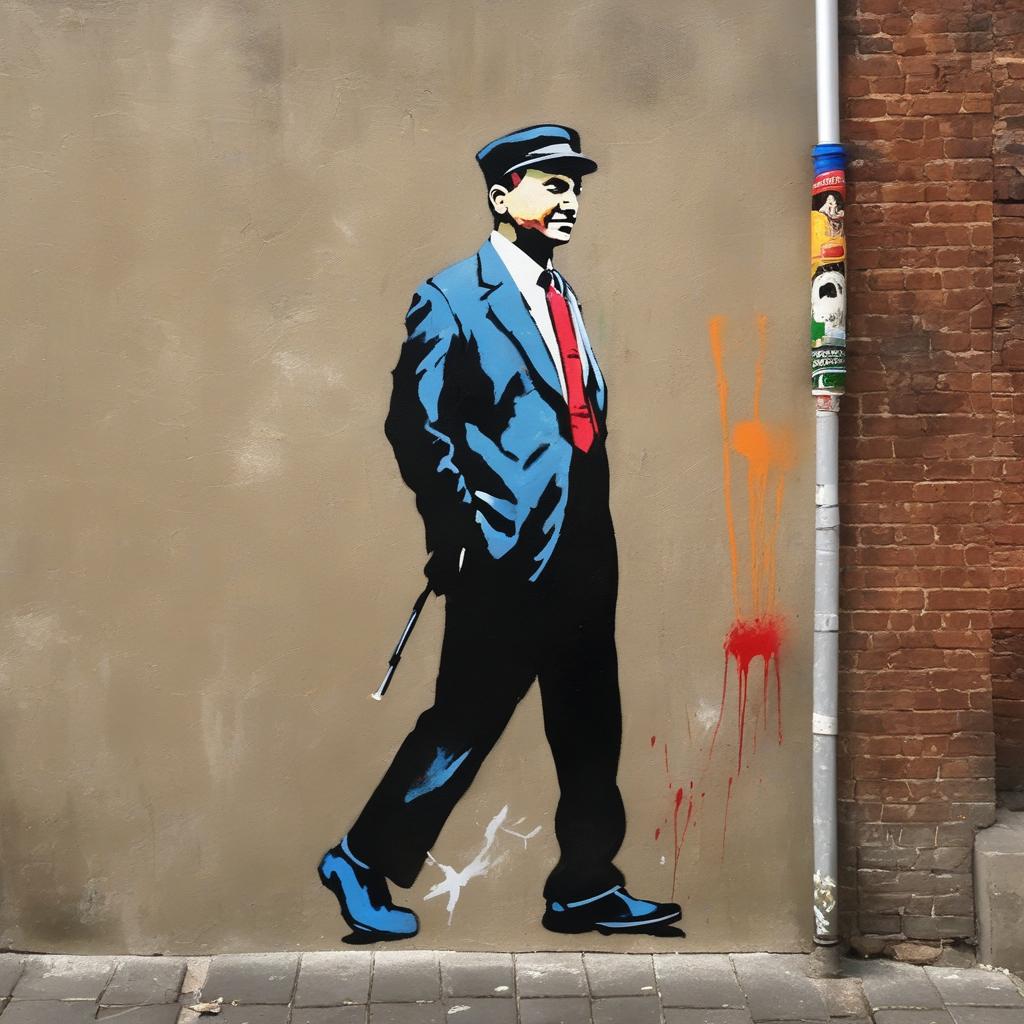} & 
        \includegraphics[width=0.09\textwidth]{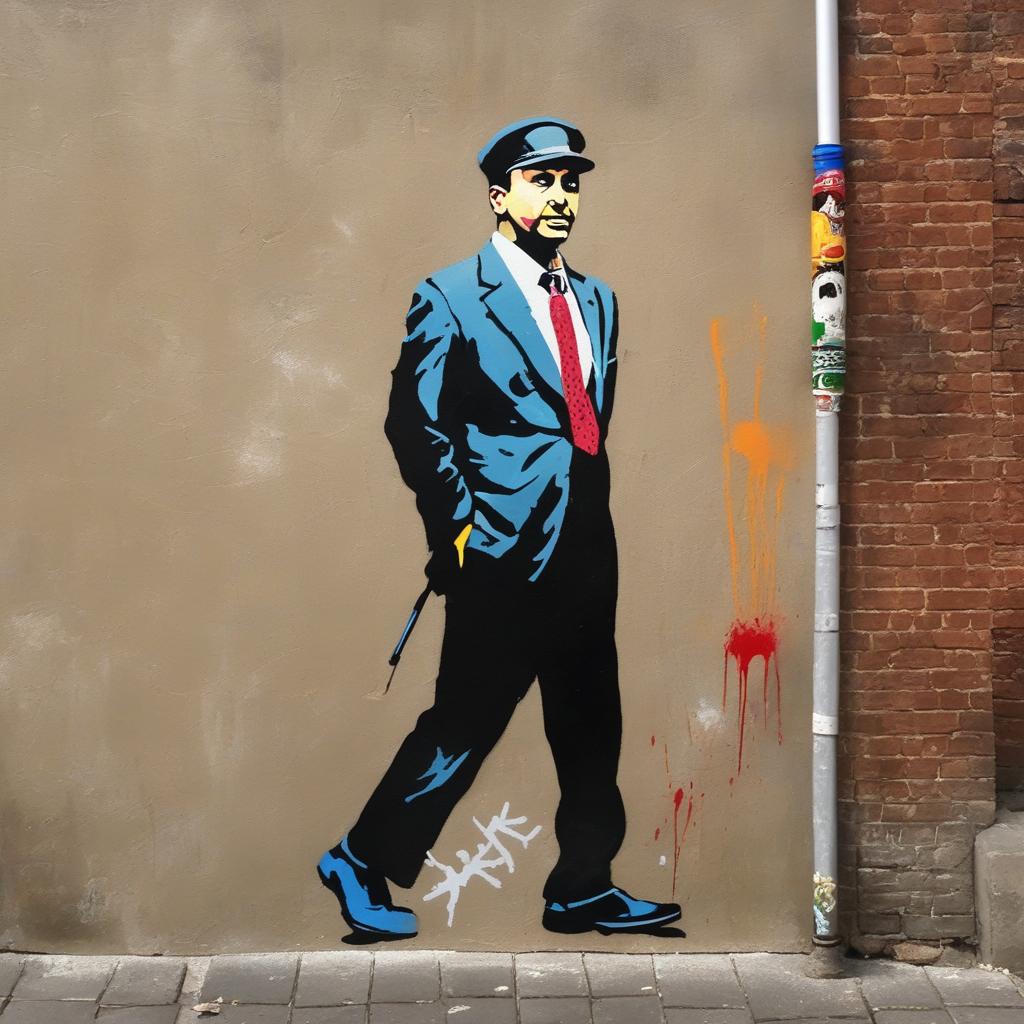} & 
        \includegraphics[width=0.09\textwidth]{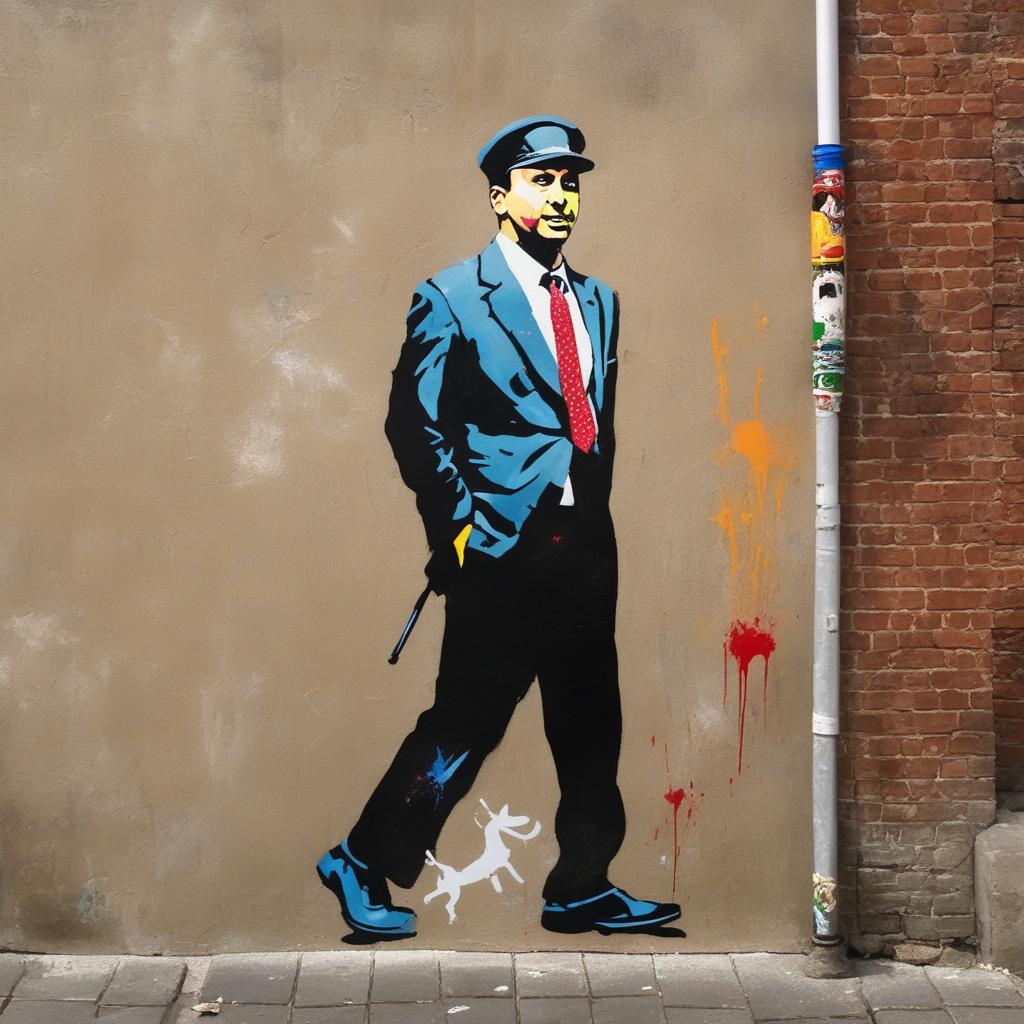} & 
        \includegraphics[width=0.09\textwidth]{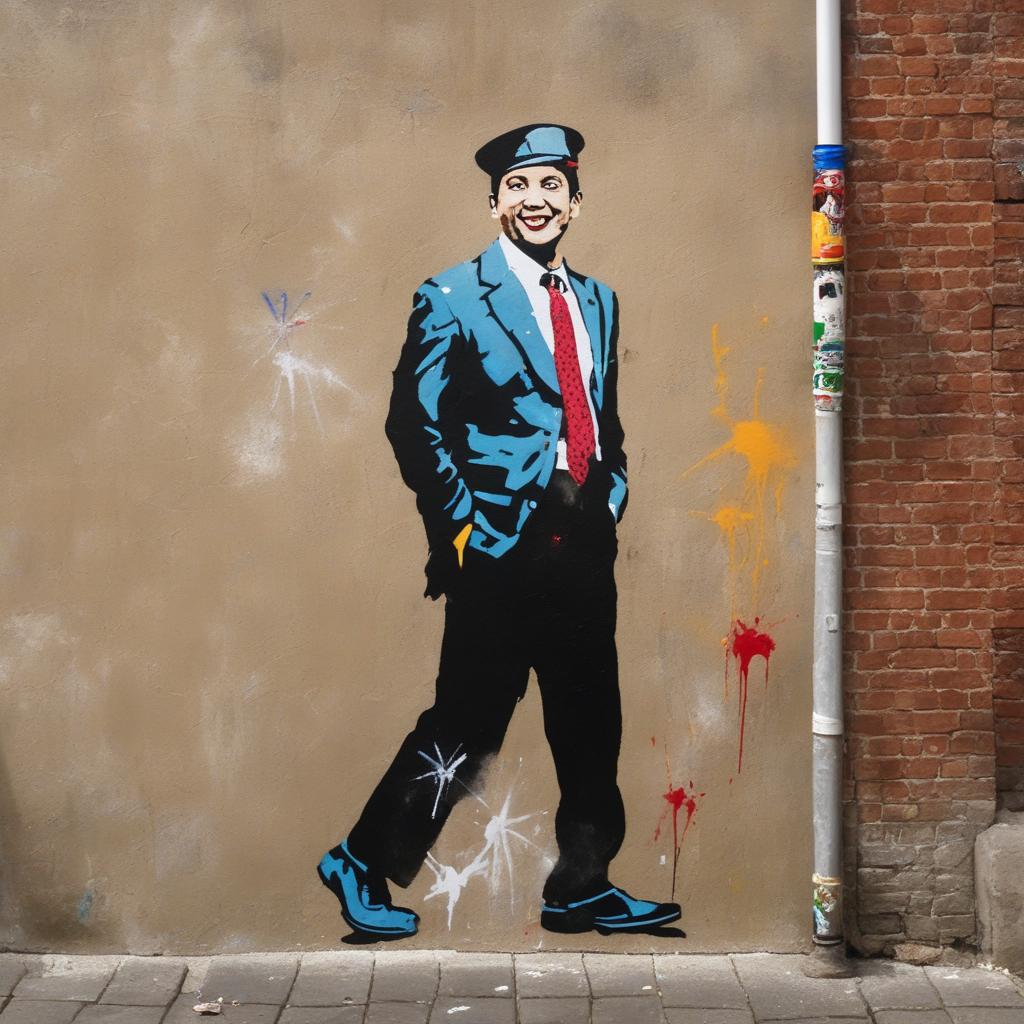} & 
        \includegraphics[width=0.09\textwidth]{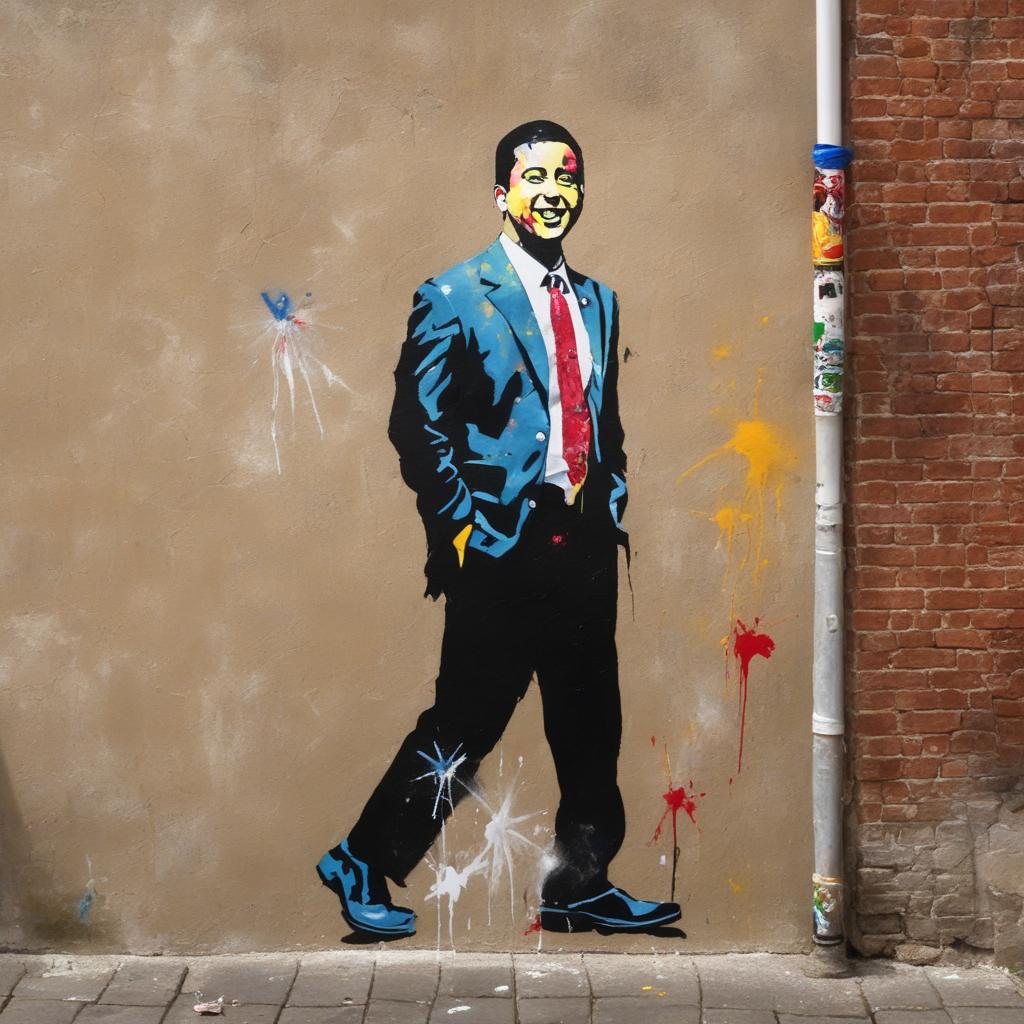} & 
        \includegraphics[width=0.09\textwidth]{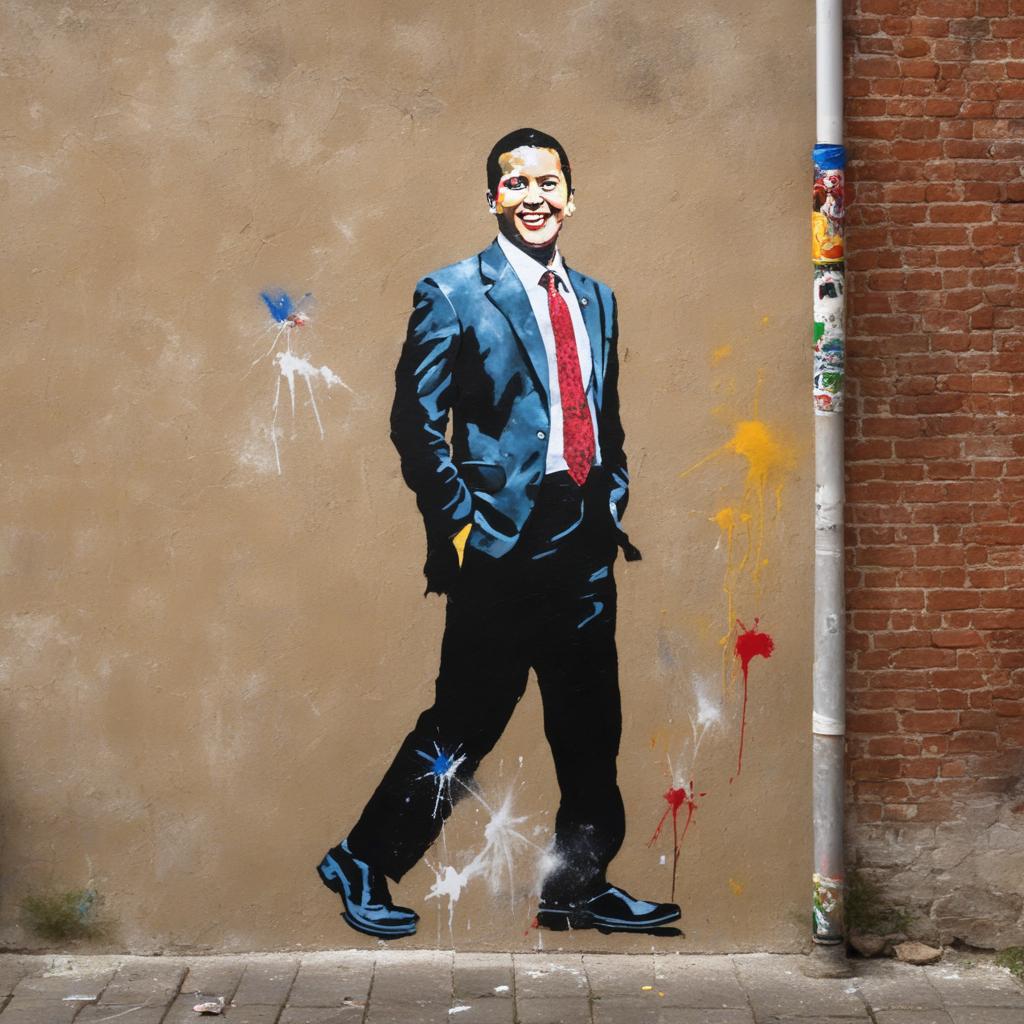} & 
        \includegraphics[width=0.09\textwidth]{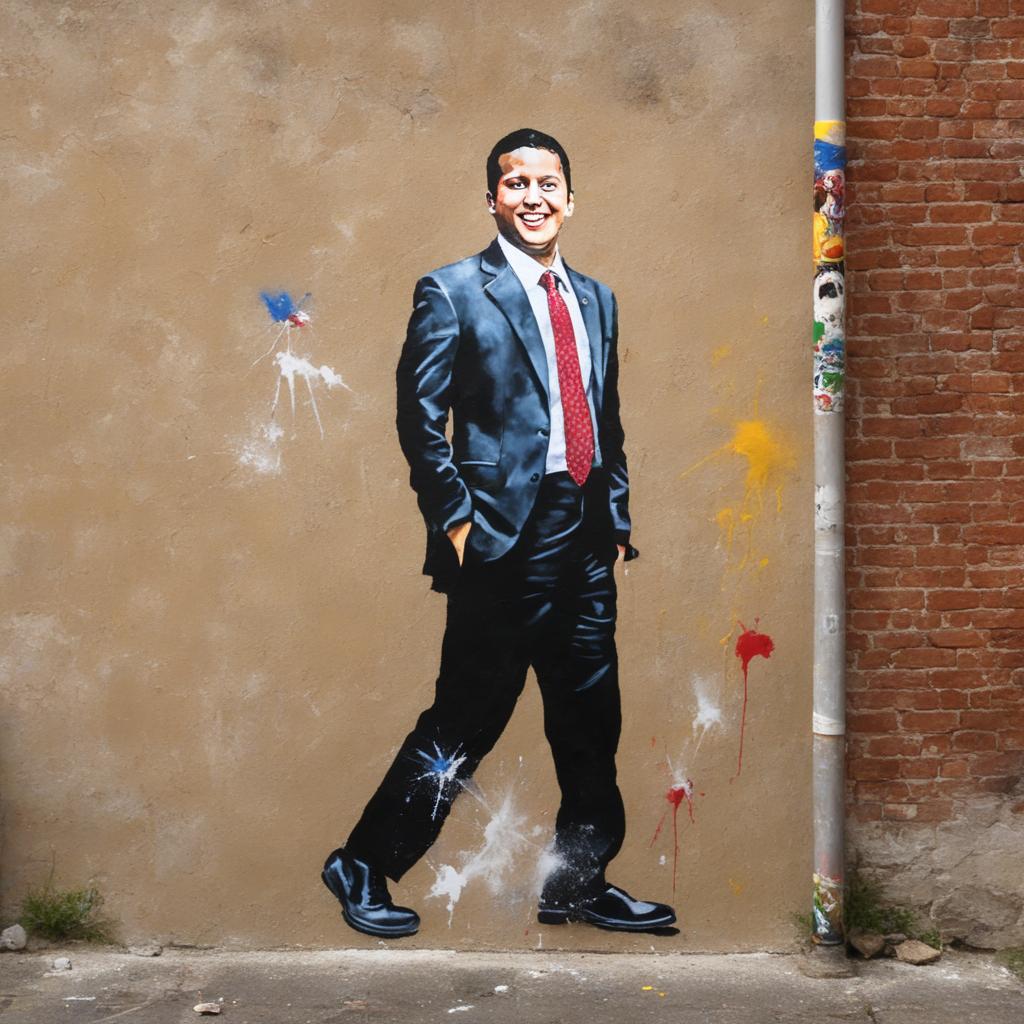} & 
        \includegraphics[width=0.09\textwidth]{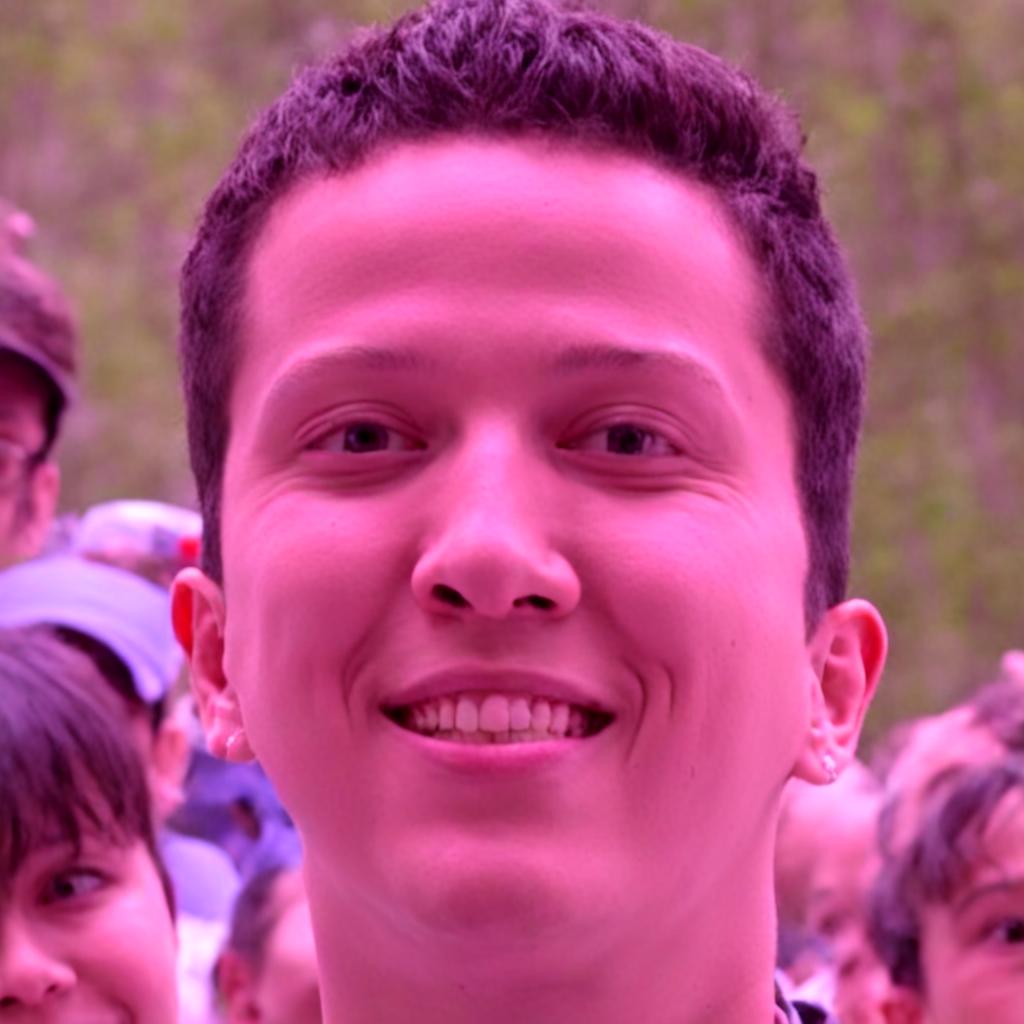} \\
        \vspace{3pt} \\[-4pt]
        \cline{3-11}
        \vspace{3pt} \\[-3pt]
        &
        \raisebox{0.22in}{\rotatebox[origin=t]{90}{Merged}}&
       \includegraphics[width=0.09\textwidth]{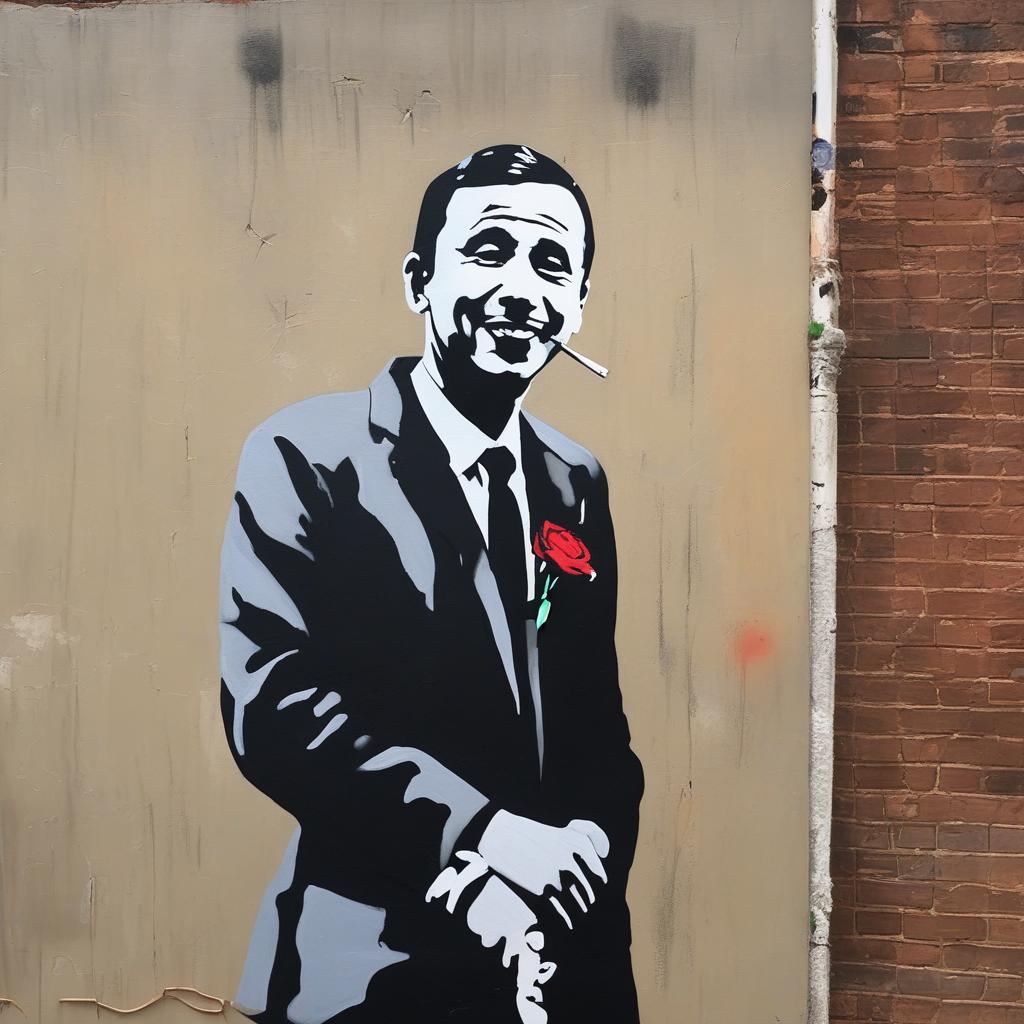} & 
        \includegraphics[width=0.09\textwidth]{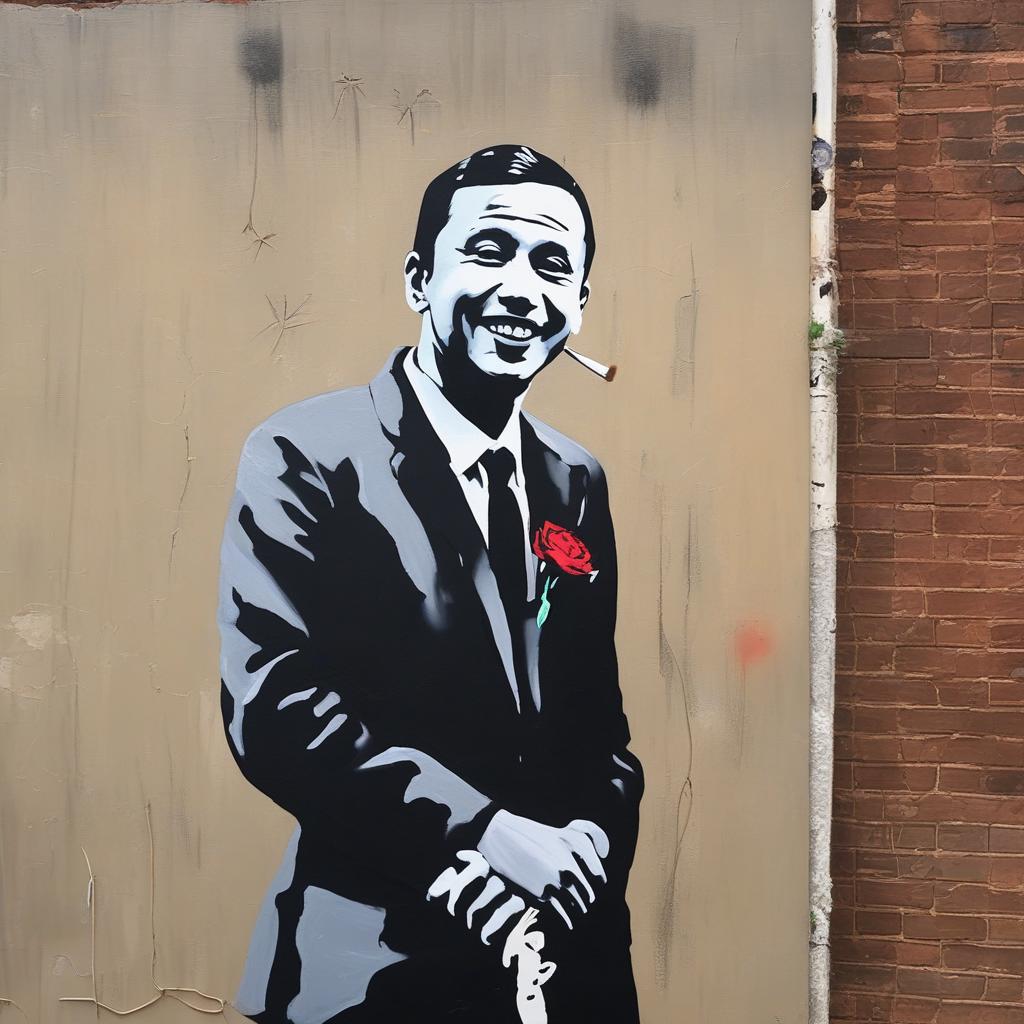} & 
        \includegraphics[width=0.09\textwidth]{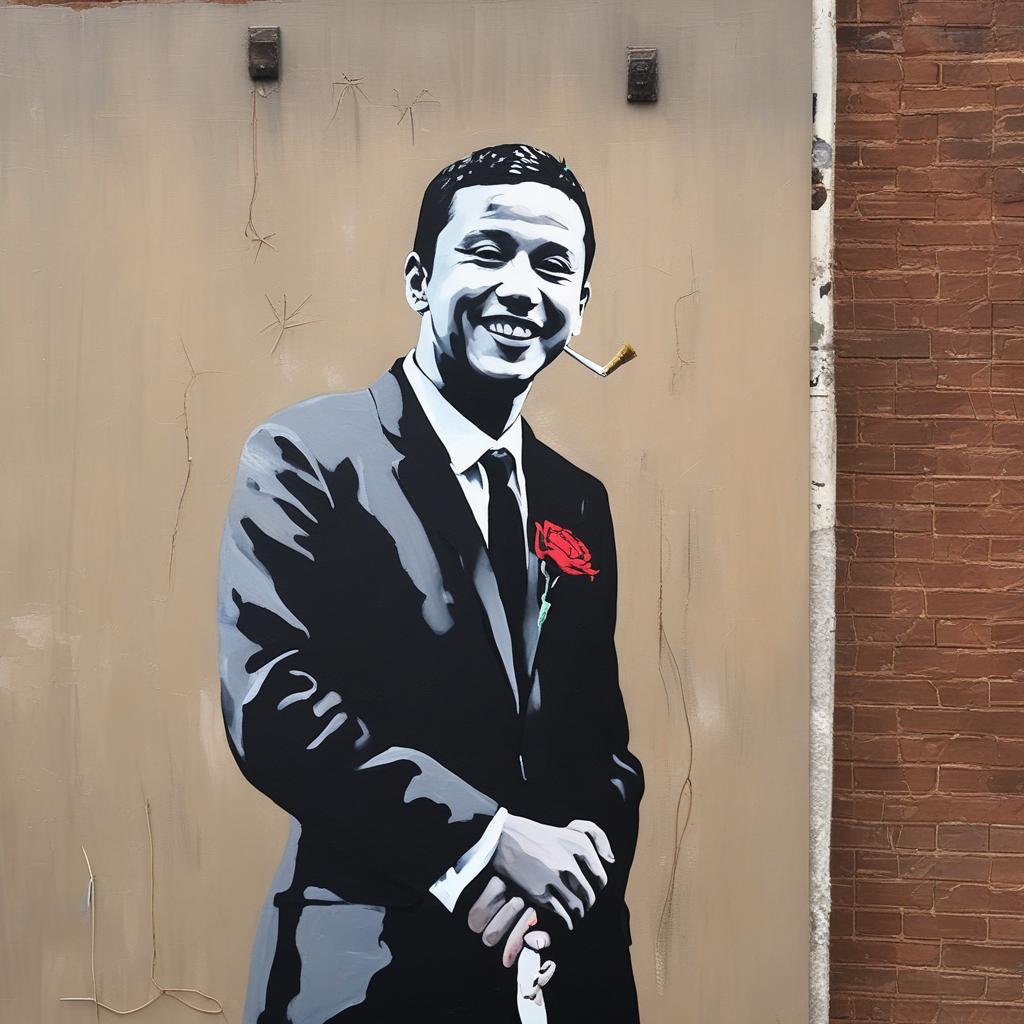} & 
        \includegraphics[width=0.09\textwidth]{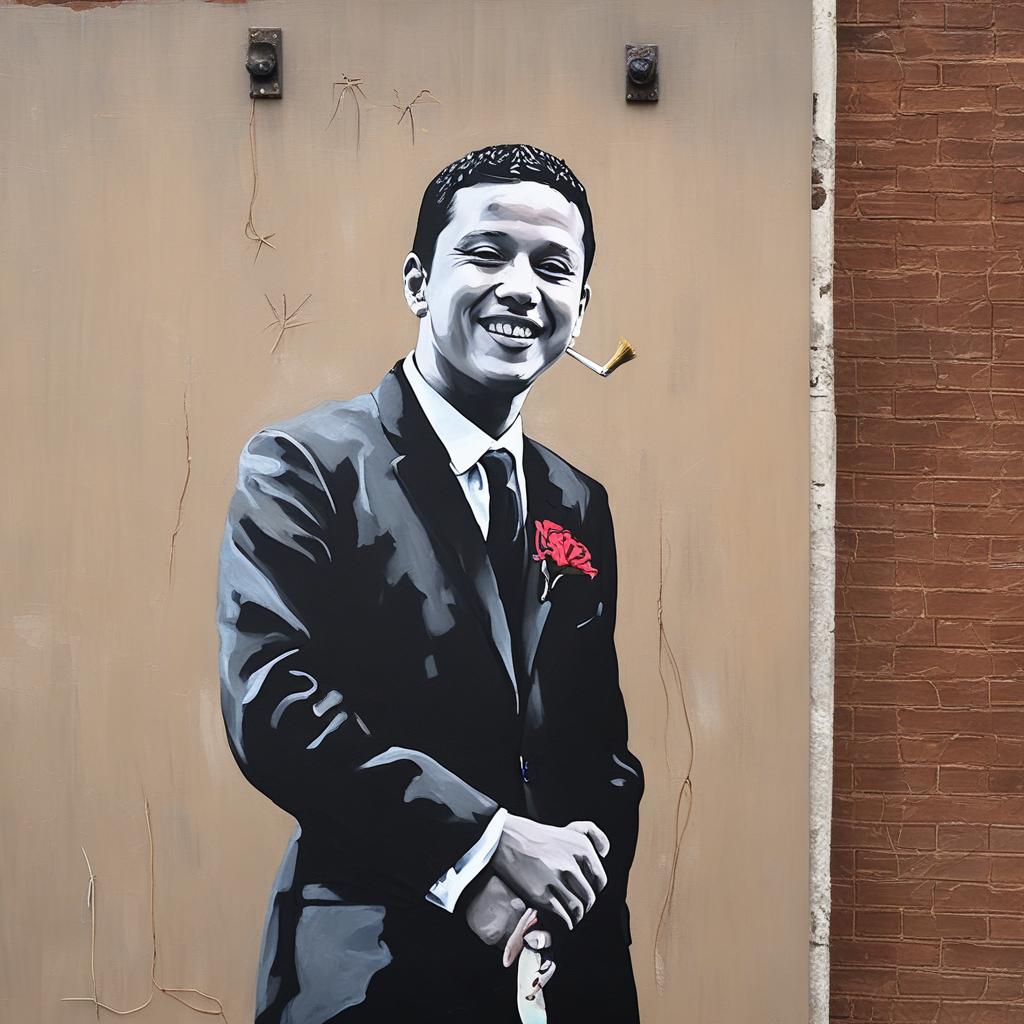} & 
        \includegraphics[width=0.09\textwidth]{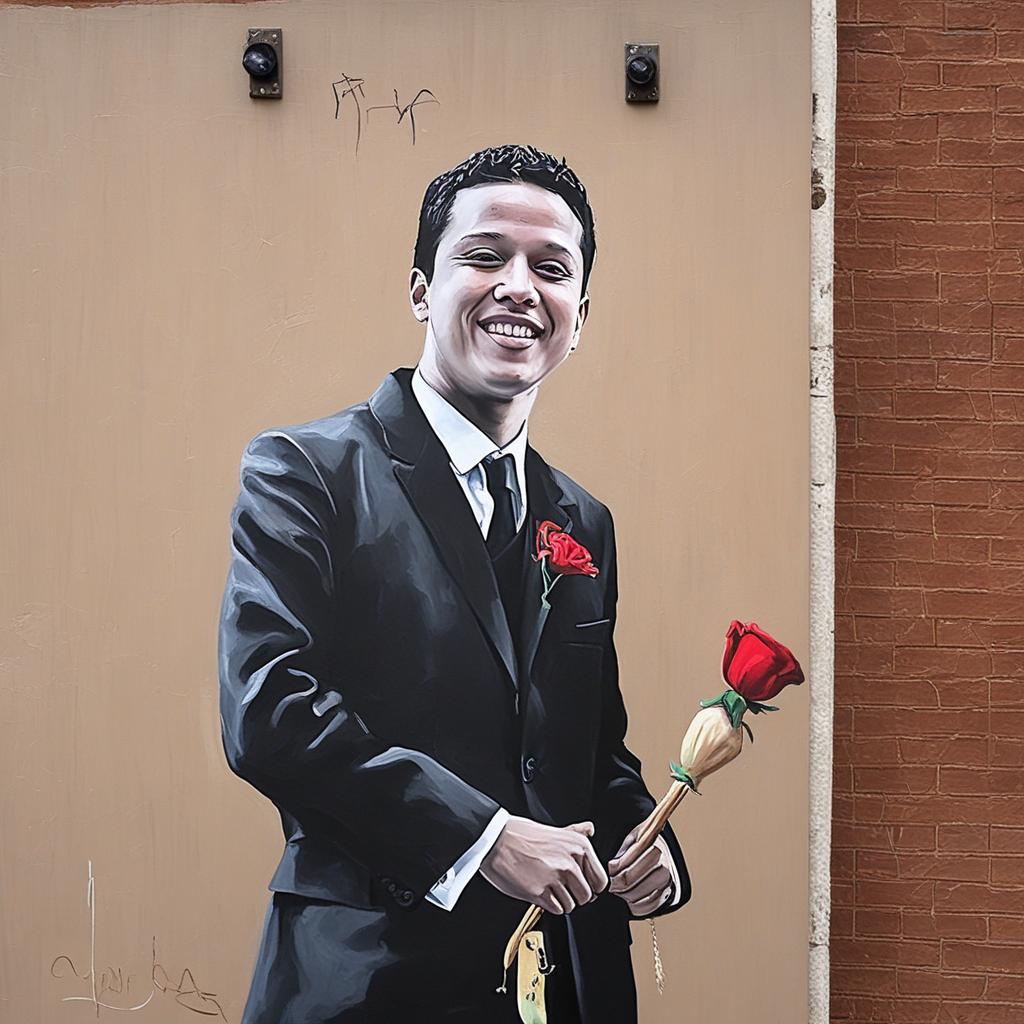} & 
        \includegraphics[width=0.09\textwidth]{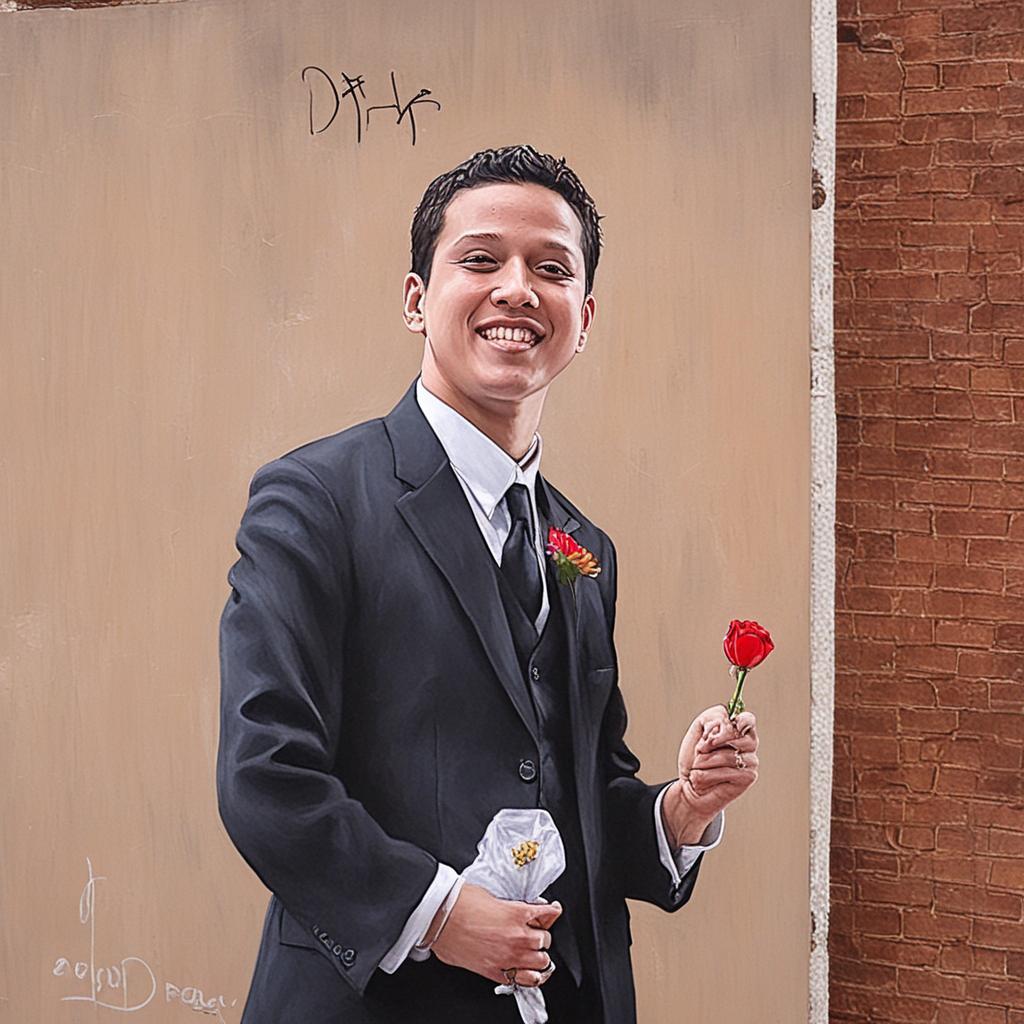} & 
        \includegraphics[width=0.09\textwidth]{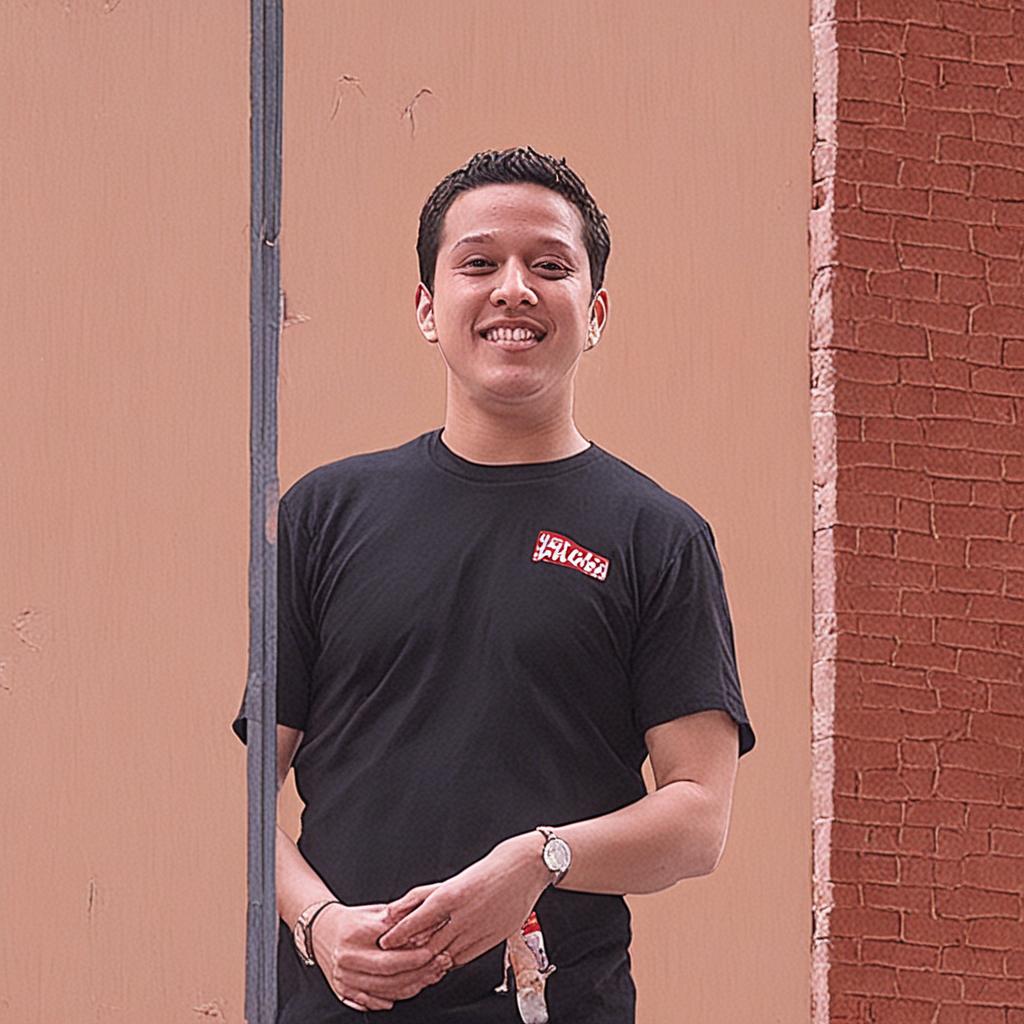} & 
        \includegraphics[width=0.09\textwidth]{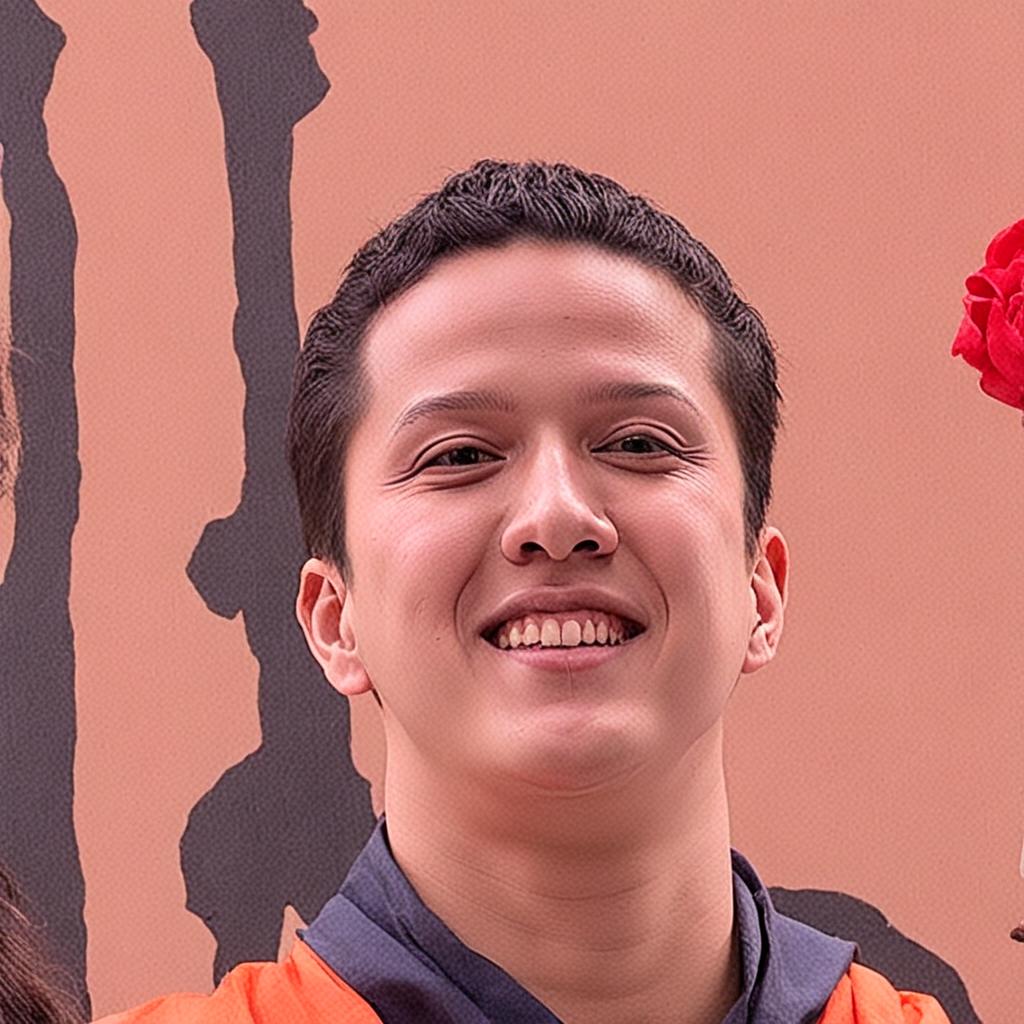} & 
        \includegraphics[width=0.09\textwidth]{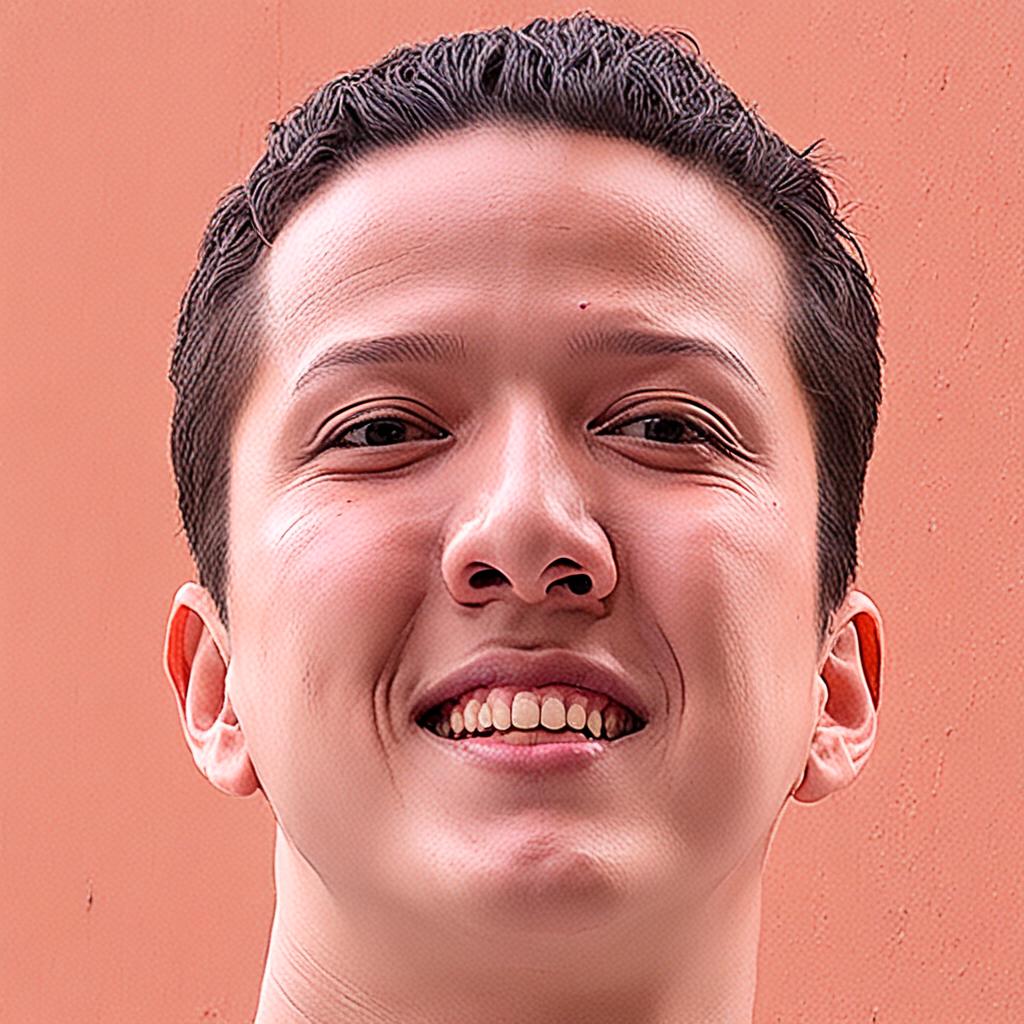} \\
    \end{tabular}
    }
    \vspace{-3pt}
\caption{\textbf{Effect of varying rescaling weights.} Merging the two branches substantially improves identity preservation while maintaining text controllability comparable to the individual branches when rescaling weights are set to relatively small values. Text prompt: ``a painting of a man in the style of Banksy''. Please zoom in for a better view.}

\label{fig:rescaling_weight}
\vspace{-10pt}
\end{figure*}

\vspace{3pt}
\noindent\textbf{User Study.}
\label{sec:user_study}
To further validate the effectiveness of our method, we conducted a user study with human evaluators. For each comparison, participants were presented with a randomly selected test image, a corresponding textual prompt, and two personalized outputs generated by our method and a baseline method, respectively. Participants were asked to select the image that better preserves the individual's identity while more accurately reflecting the prompt description. We collected 720 responses from 40 participants, with results summarized in Table~\ref{tab:user_study}. The results demonstrate a clear preference for images generated by our method.

\noindent\textbf{Effect of Varying Rescaling Weights.}
Figure~\ref{fig:rescaling_weight} illustrates the effect of varying rescaling weights applied to the two branches, which enable flexible control over the trade-off between identity preservation and text controllability. We observe that merging the two branches yields substantial improvements in identity preservation compared to employing either branch independently. Notably, when rescaling weights are set to relatively small values, the merged model maintains text controllability comparable to that of the individual branches. Additionally, without the proposed rescaling strategy, both branches exhibit severe overfitting to the input image, as we employ an identity-focused learning scheme during training. Additional results with varying rescaling weights are provided in Appendix~\ref{sec:appendix_rescaling}.

\begin{table}[t]
\centering
\caption{\textbf{Quantitative ablation study.} Removing either branch (Var A, Var B) significantly degrades identity preservation, while removing the rescaling strategy from either branch (Var C, Var D) reduces both identity preservation and text controllability.}
\vspace{-5pt}
        \begin{tabular}{@{\hspace{0.5cm}}lcc@{\hspace{0.5cm}}}
          \toprule
          Methods  & Identity$\uparrow$& Controllability$\uparrow$
          \\
          \midrule
          Var A       & 0.2283            & 31.71   \\
          Var B       & 0.1907            & 32.17            \\
          Var C      & 0.3810            & 21.22            \\
          Var D          & 0.4077            & 21.54            \\
          \midrule
          Full model       & 0.4694            & 30.67 \\
          \bottomrule
        \label{tab:ablation_study}
        \vspace{-0.8cm}
    \end{tabular}
\end{table}

\subsection{Ablation Study}
\begin{figure}
    \centering
    \renewcommand{\arraystretch}{0.3}
    \setlength{\tabcolsep}{0.1pt}

    {\small
    \begin{tabular}{c c c c c c}

         Input &
        \multicolumn{1}{c}{ Var A} &
        \multicolumn{1}{c}{ Var B} &
        \multicolumn{1}{c}{ Var C} &
        \multicolumn{1}{c}{ Var D}  &
        \multicolumn{1}{c}{ Full} \\
        
        \includegraphics[width=0.166\linewidth]{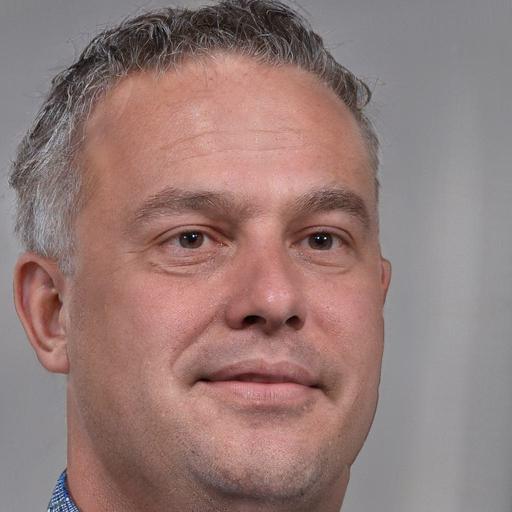} &

        \includegraphics[width=0.166\linewidth]{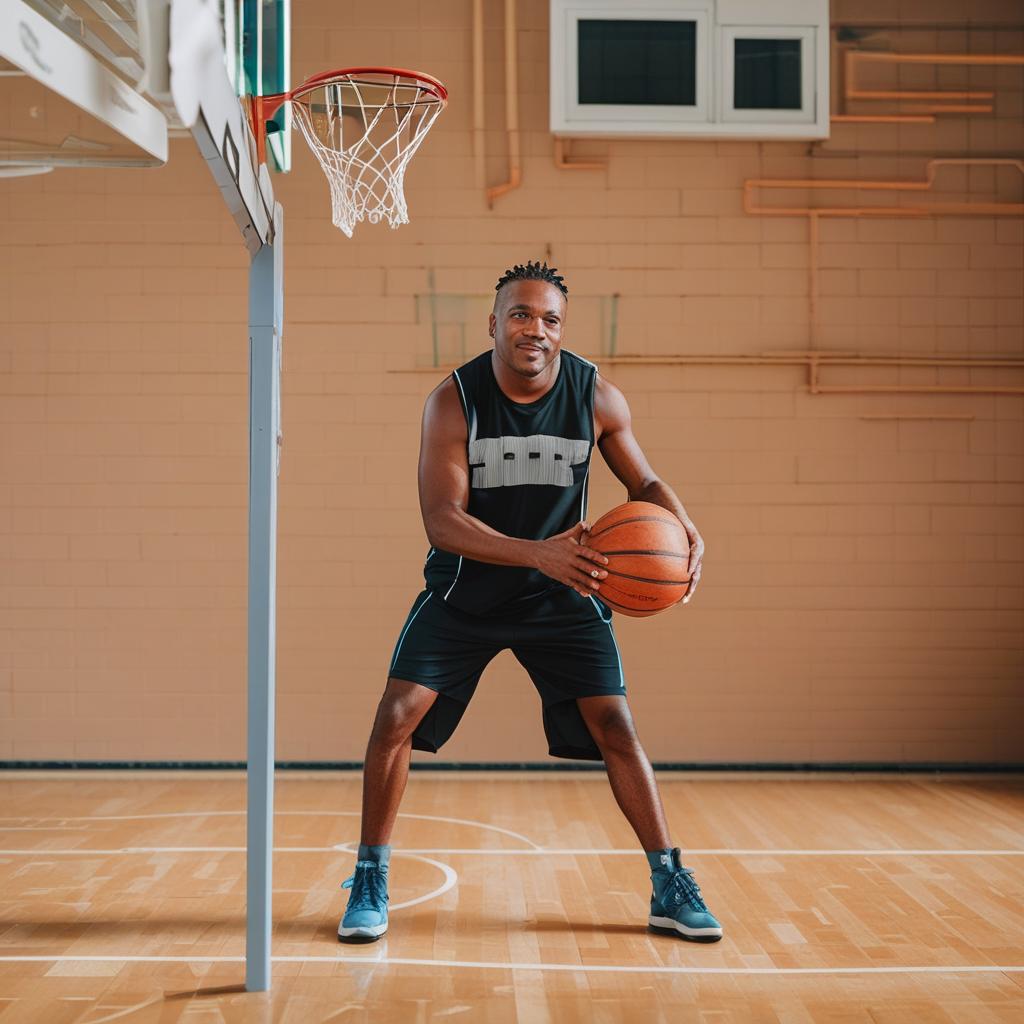} &
        
        \includegraphics[width=0.166\linewidth]{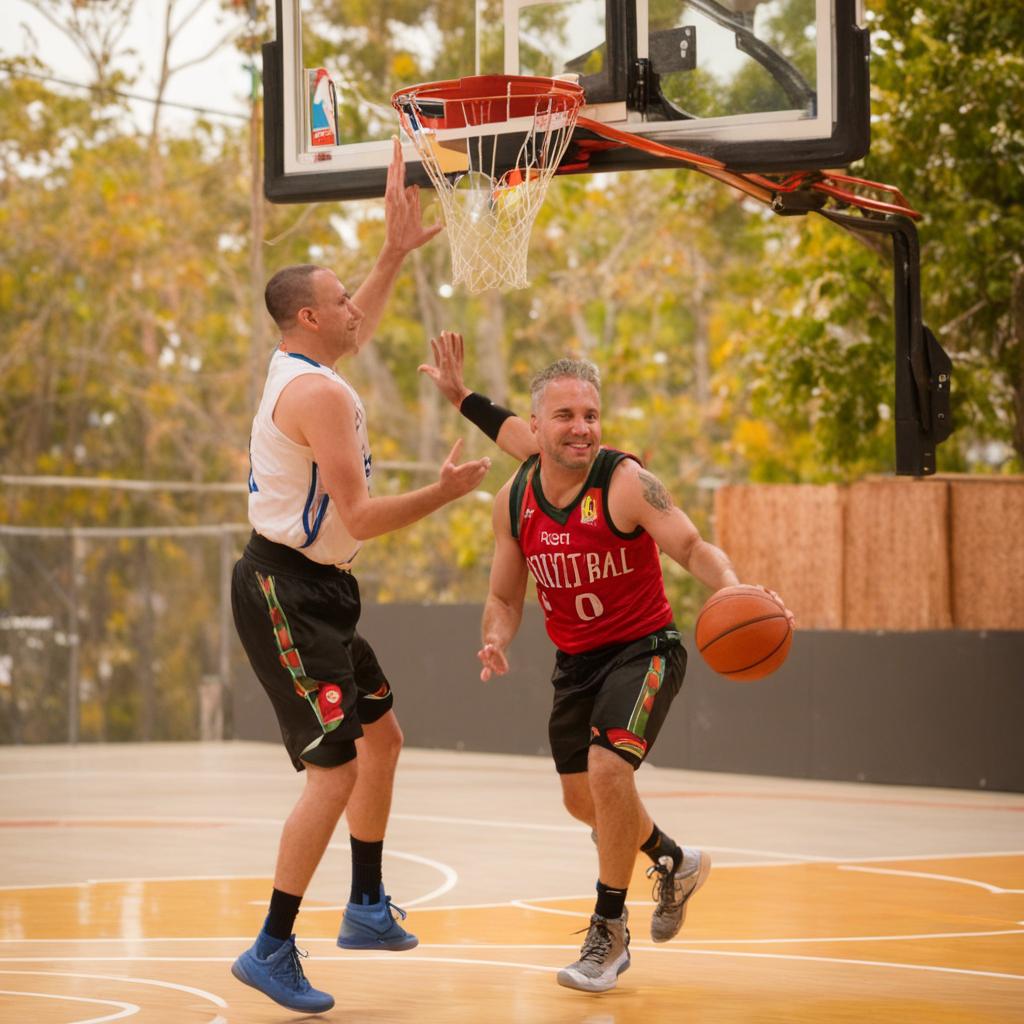} &
        
        \includegraphics[width=0.166\linewidth]{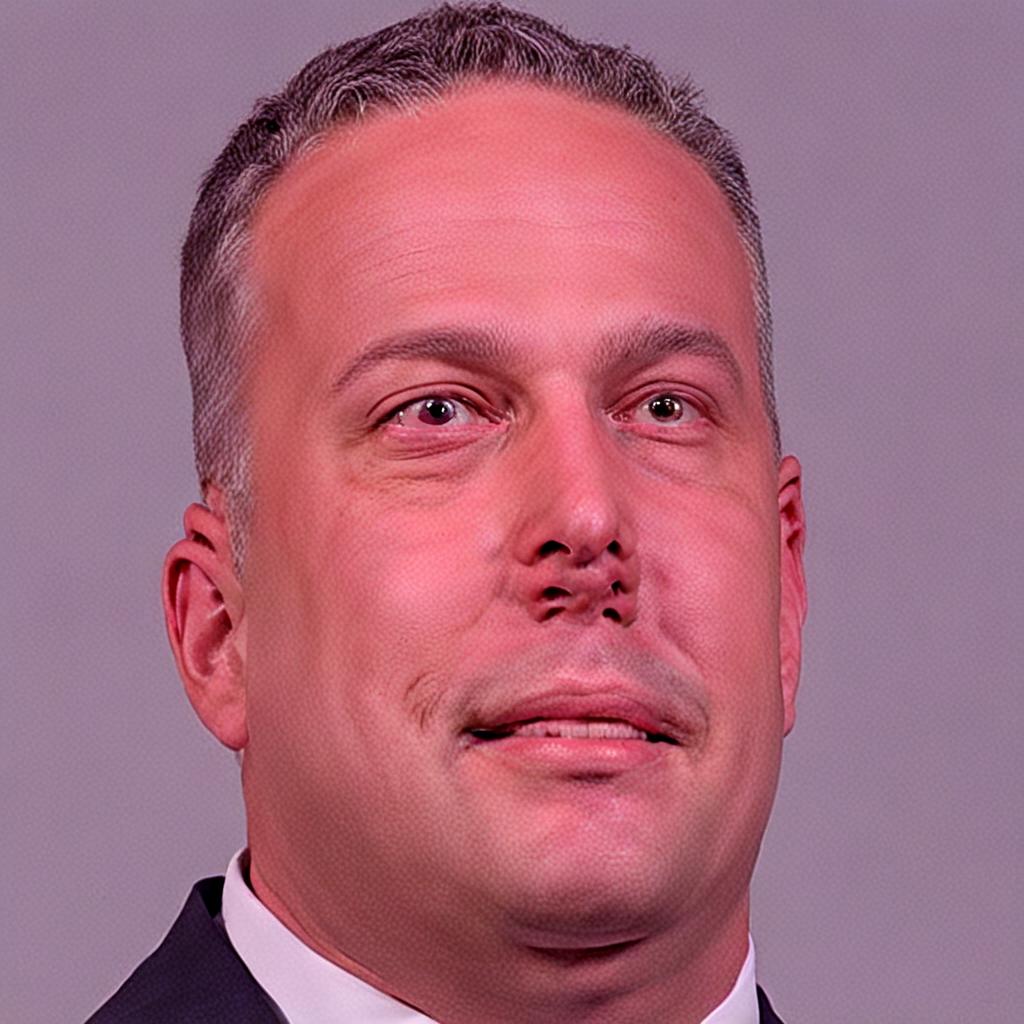} &
        
        \includegraphics[width=0.166\linewidth]{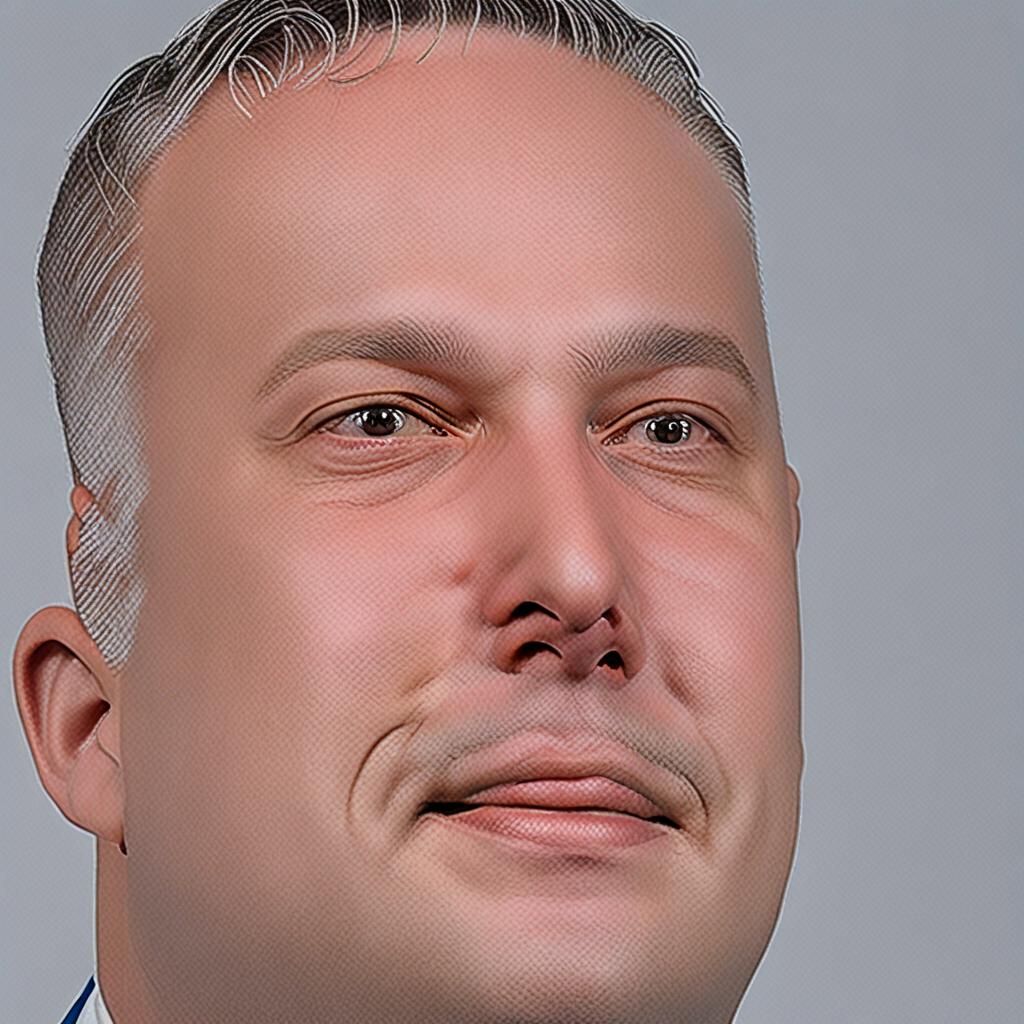} &
        \includegraphics[width=0.166\linewidth]{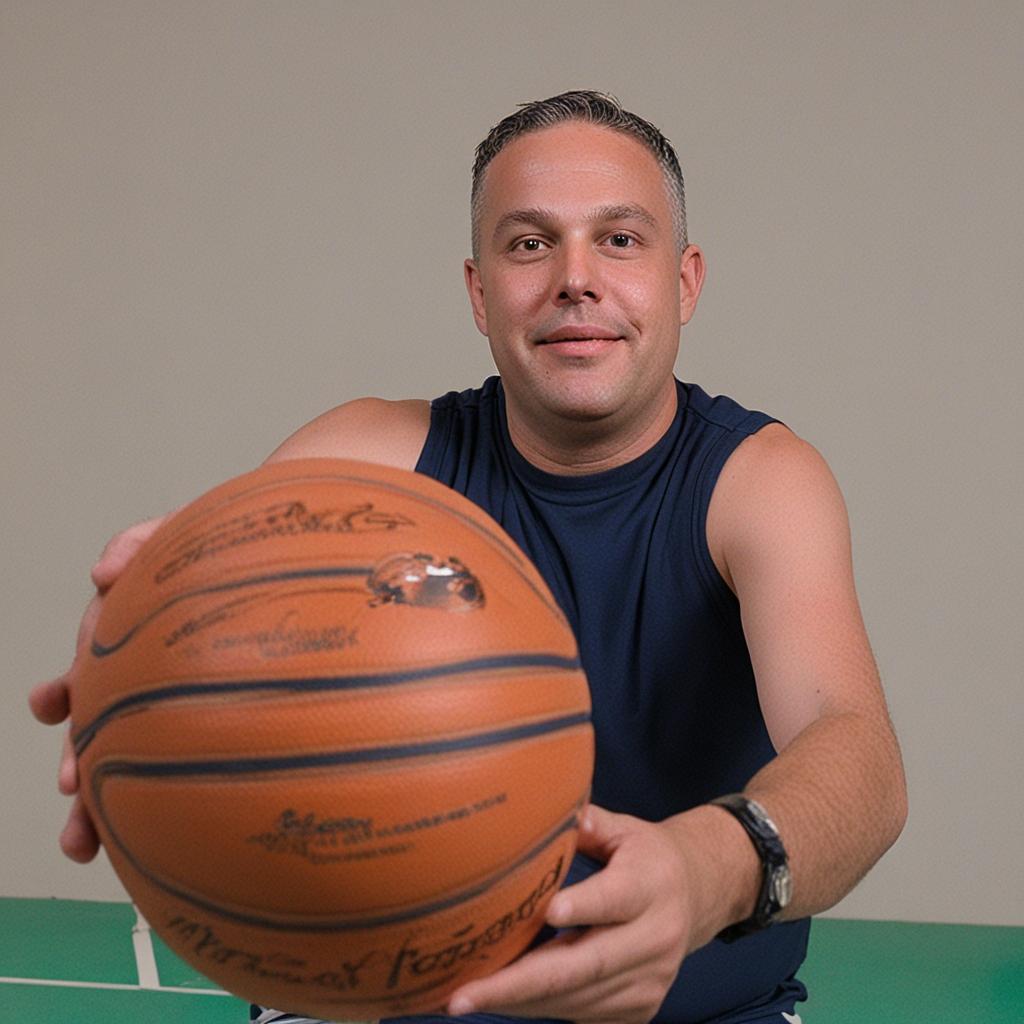} \\

        &\multicolumn{5}{c}{A man playing basketball} \\

        \includegraphics[width=0.166\linewidth]{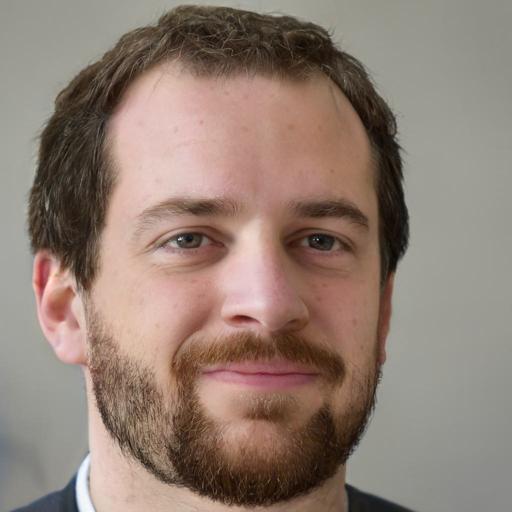} &
        
        \includegraphics[width=0.166\linewidth]{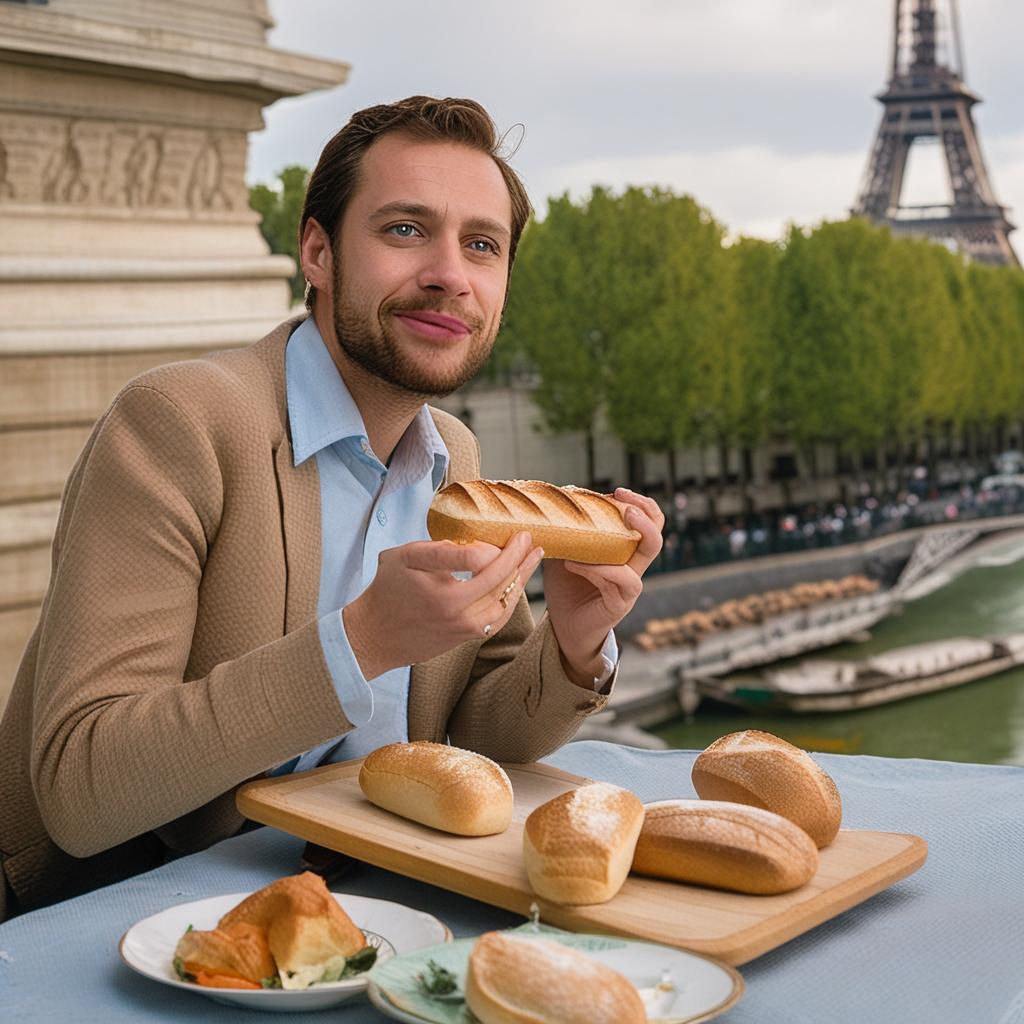} &
        
        \includegraphics[width=0.166\linewidth]{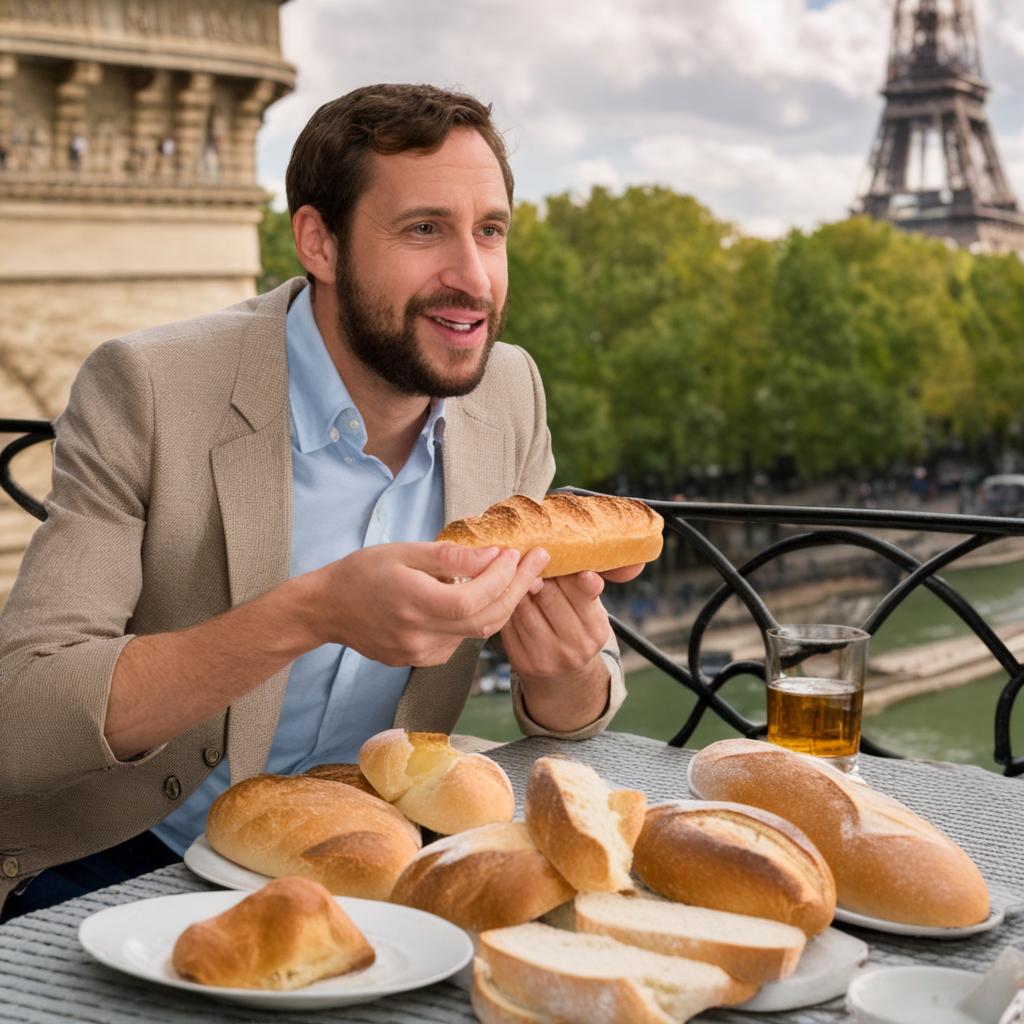} &
        
        \includegraphics[width=0.166\linewidth]{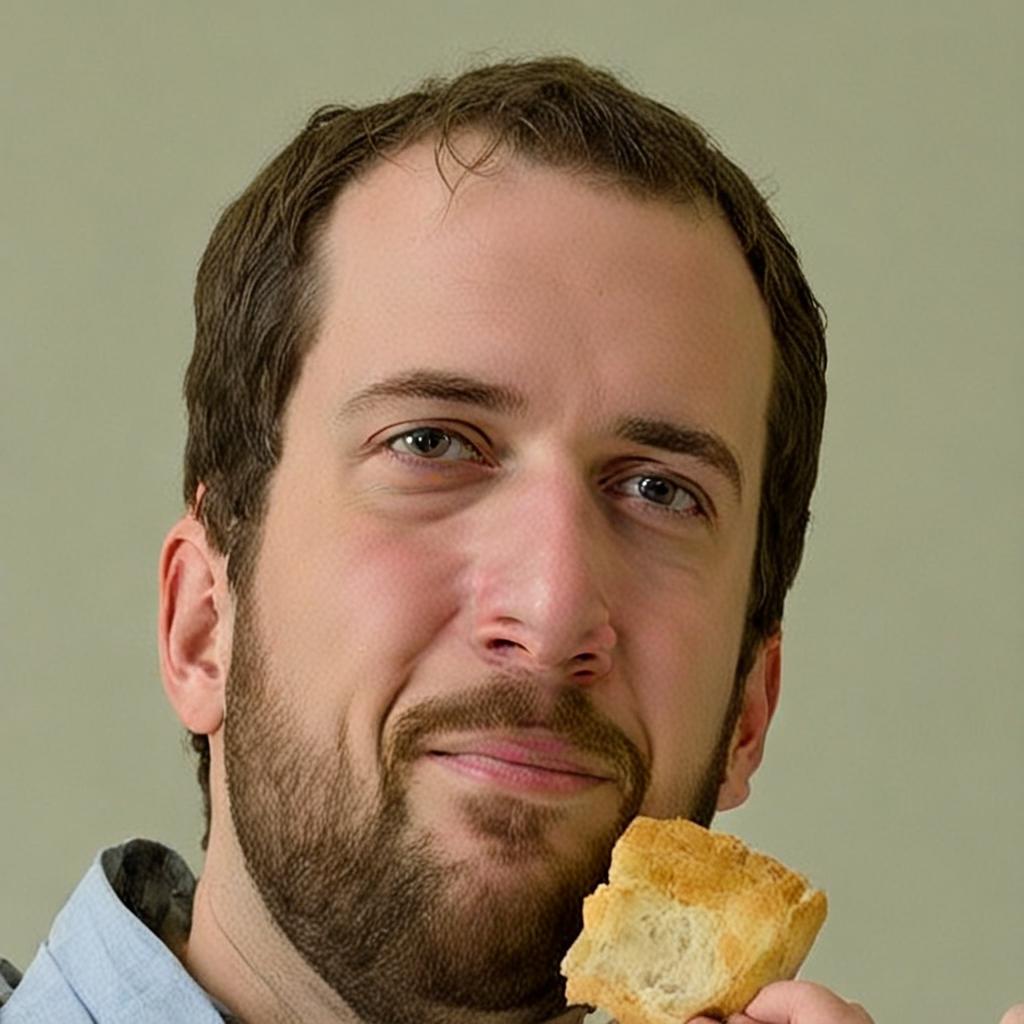} &
        
        \includegraphics[width=0.166\linewidth]{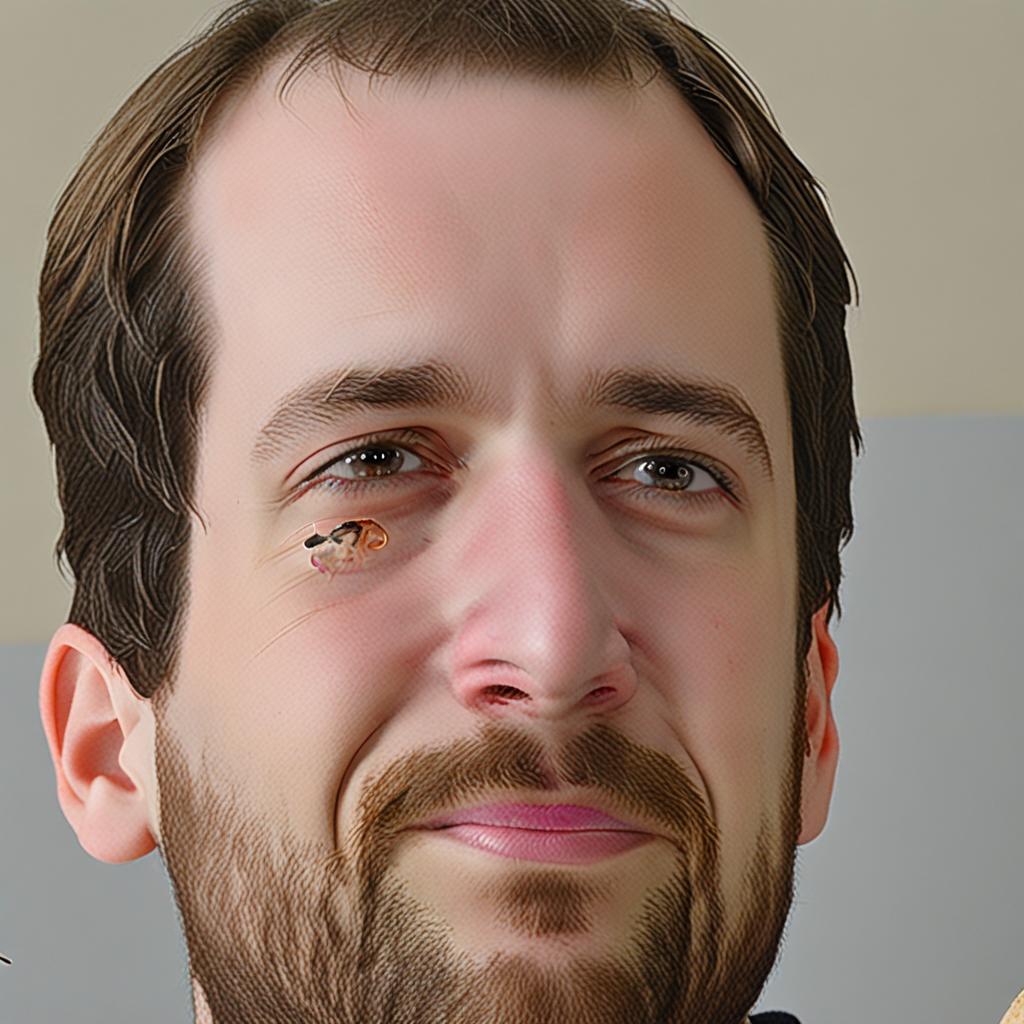} &
        \includegraphics[width=0.166\linewidth]{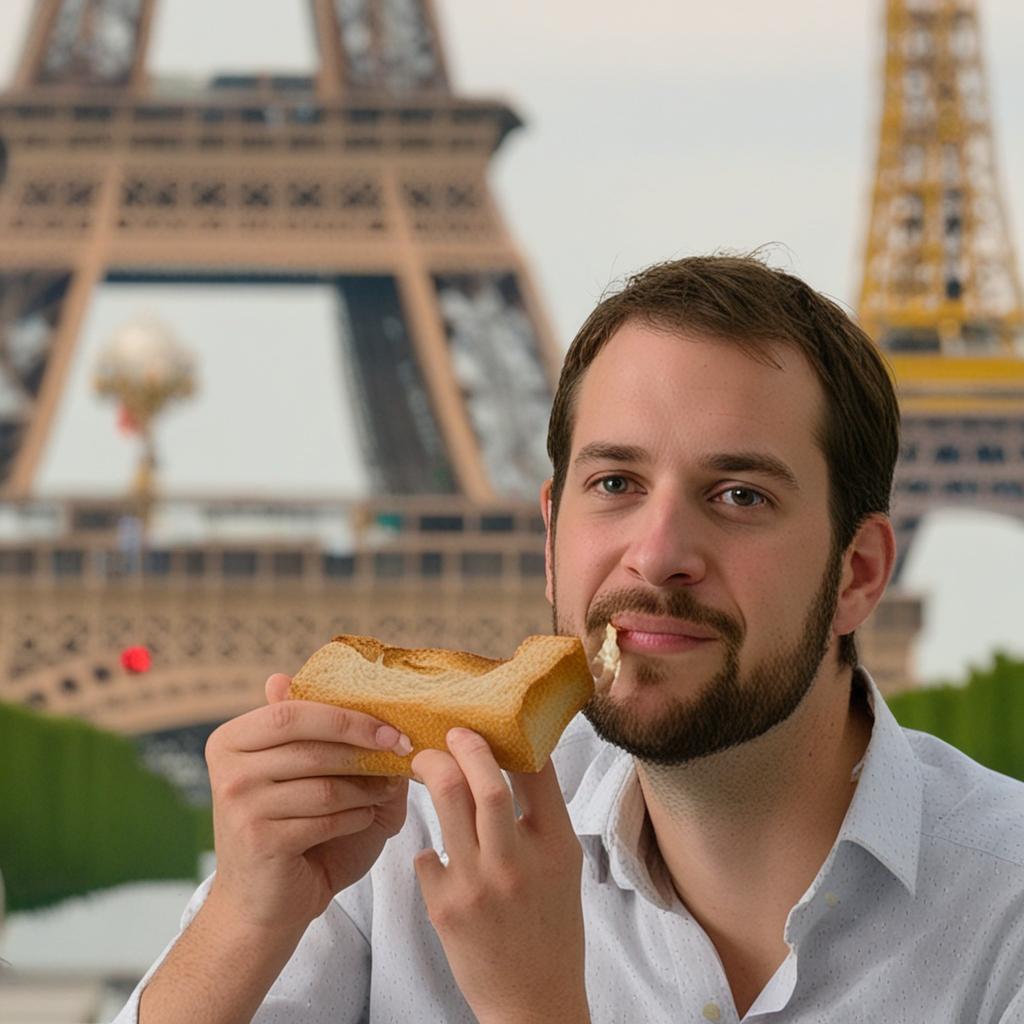} \\

        &\multicolumn{5}{c}{A man eating bread in front of the Eiffel Tower} \\
    \end{tabular}
    }
    \vspace{-5pt}
    \caption{\textbf{Qualitative ablation study.} Removing either branch (Var A, Var B) significantly degrades identity preservation. Removing the rescaling strategy from either branch (Var C, Var D) causes noticeable distortions and artifacts in the generated subjects.}    
\label{fig:ablation}
\vspace{-10pt}
\end{figure}

To validate the effectiveness of each component in our framework, we conduct an ablation study by systematically removing individual sub-modules. Specifically, we evaluate four variants: without the text embedding branch (Var A), without the adapter branch (Var B), without the rescaling strategy in the text embedding branch (Var C), and without the rescaling strategy in the adapter branch (Var D). Qualitative and quantitative comparisons are presented in Figure~\ref{fig:ablation} and Table~\ref{tab:ablation_study}, respectively.
The results demonstrate the critical role of each component. Removing either branch significantly degrades identity preservation, demonstrating that integrating both branches effectively combines their complementary strengths. Furthermore, ablating the rescaling strategy from either branch substantially reduces both identity preservation and text controllability, as the generated subjects exhibit noticeable distortions and artifacts.

\section{Conclusions and Limitations}
\label{sec:conclusion_and_limitation}
We present UniID, a unified framework integrating text embedding and adapter approaches through a principled training-inference paradigm. Our key contributions include an identity-focused learning scheme ensuring both branches capture exclusively identity features, and a normalized rescaling strategy recovering the text controllability of the base diffusion model. Extensive evaluation demonstrates that UniID achieves state-of-the-art performance in both identity fidelity and text alignment. One limitation is the computational overhead from dual-branch processing.

 \small \bibliographystyle{ieeenat_fullname} \bibliography{ref}

\clearpage
\appendix

\section{Additional Qualitative Results}
\label{sec:appendix_qualitative}
\begin{figure*}[t]
    \centering
    \setlength{\tabcolsep}{0.5pt}
    {\scriptsize
    \begin{tabular}{cccccc}

        \includegraphics[width=0.15\textwidth]{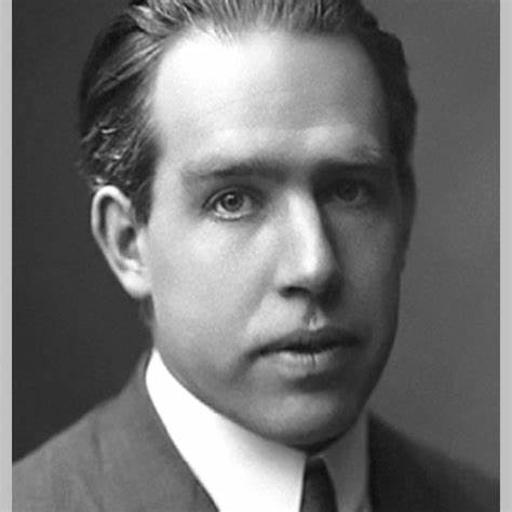} &
        \includegraphics[width=0.15\textwidth]{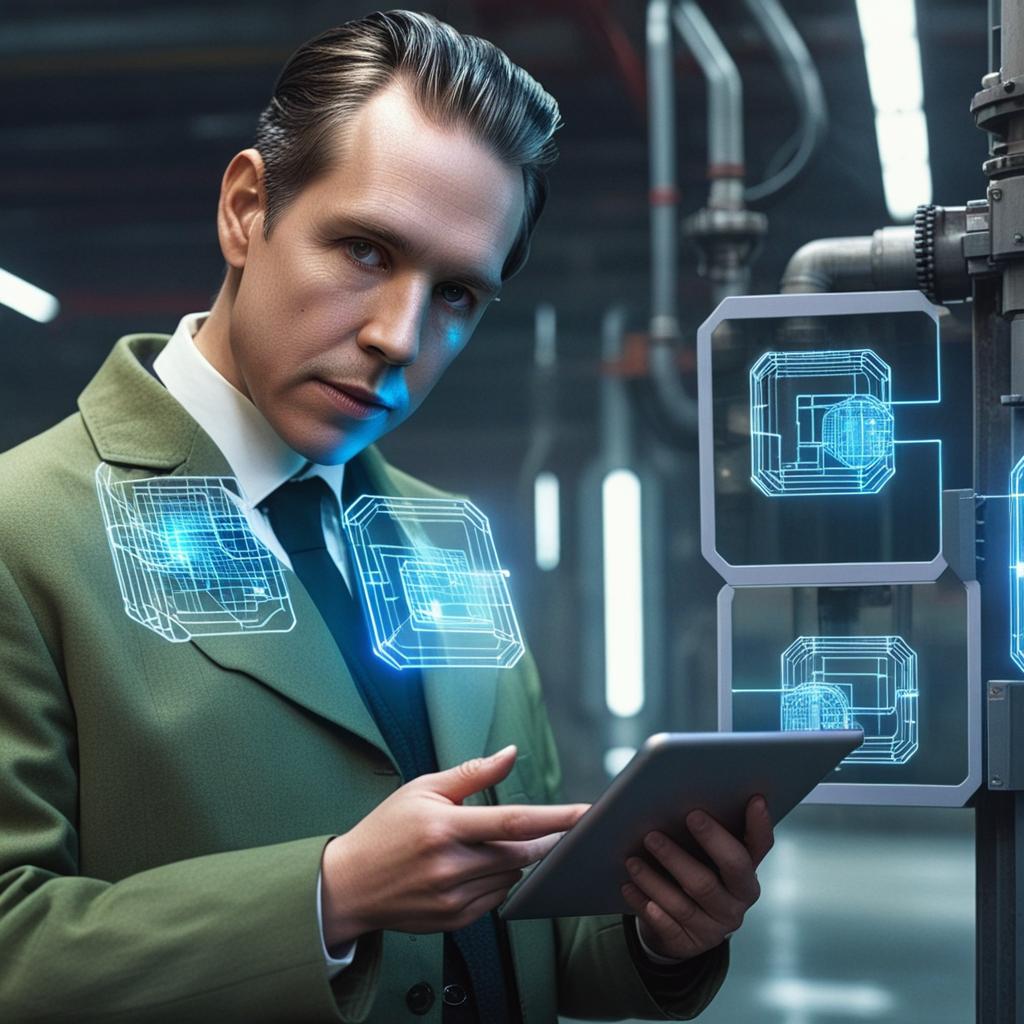} &
        \includegraphics[width=0.15\textwidth]{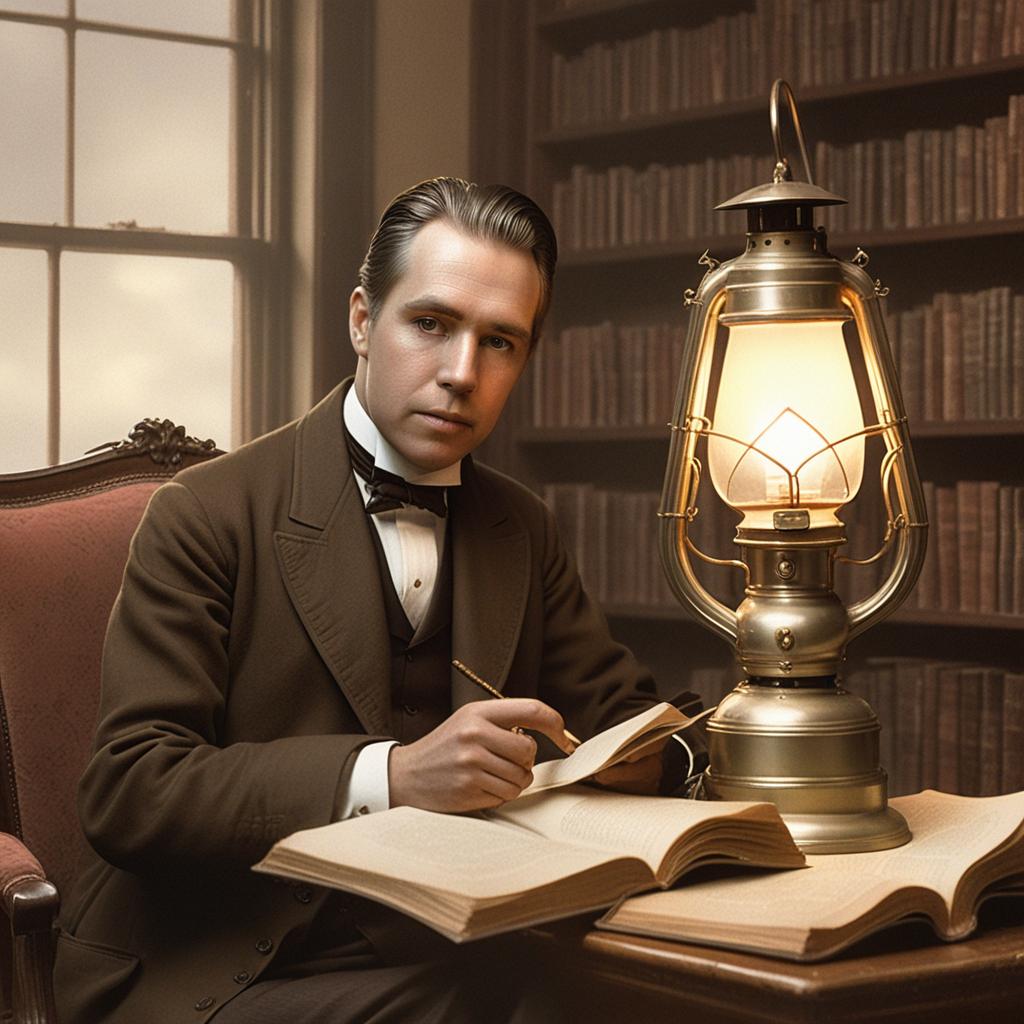} &
        \includegraphics[width=0.15\textwidth]{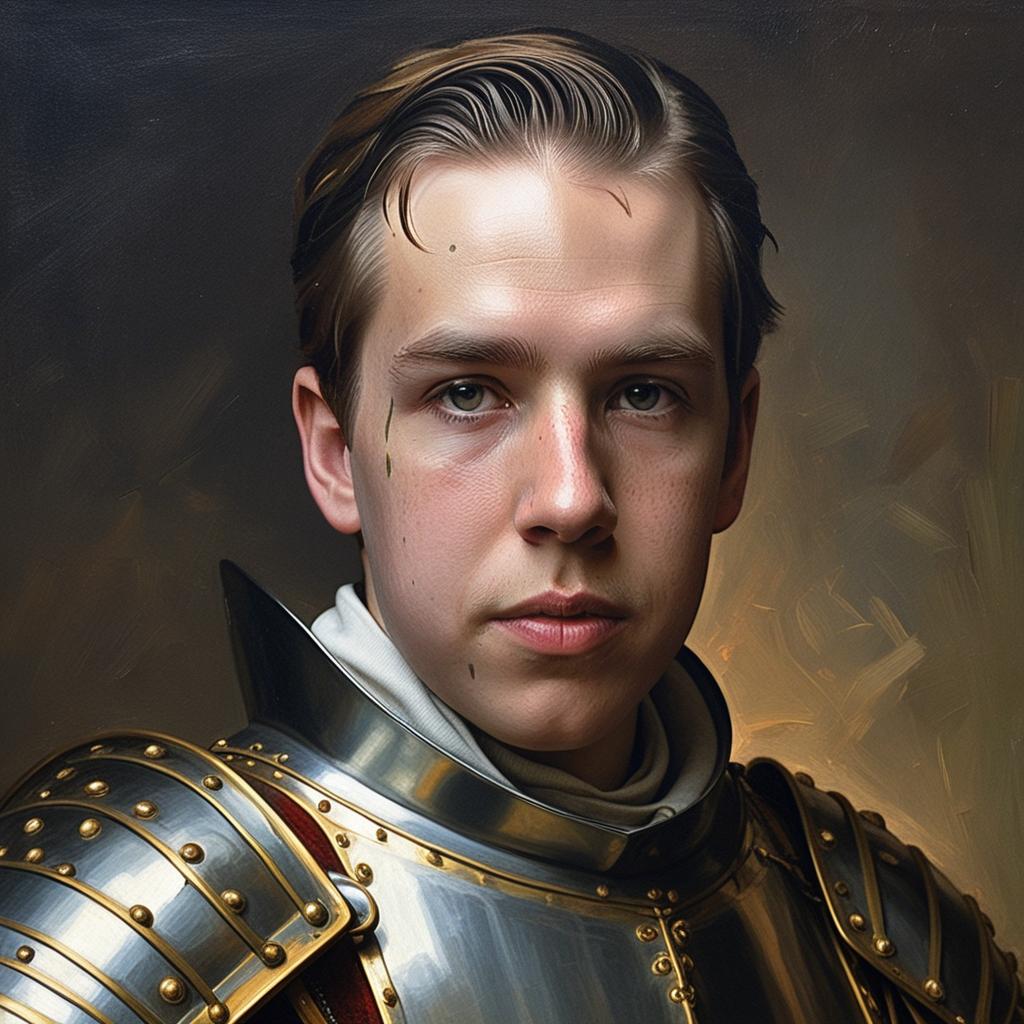} &
        \includegraphics[width=0.15\textwidth]{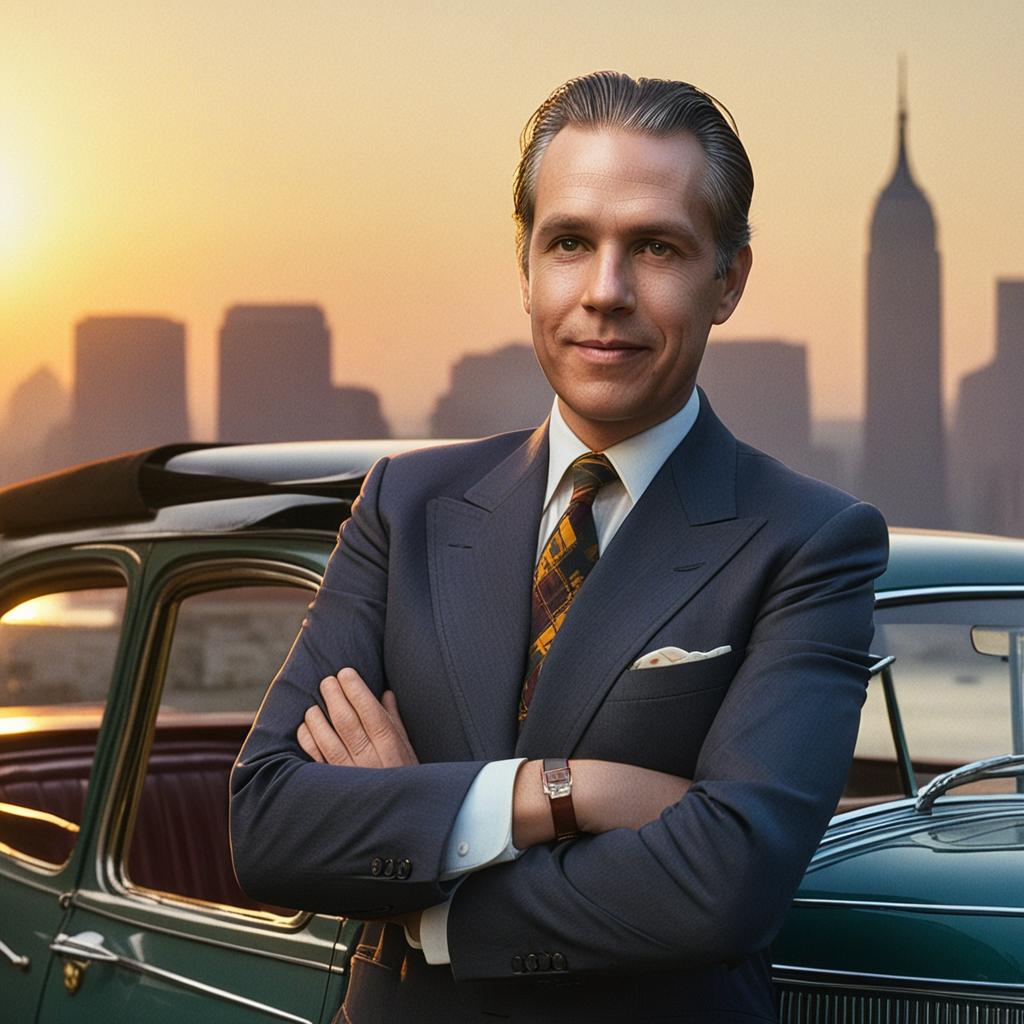} &
        \includegraphics[width=0.15\textwidth]{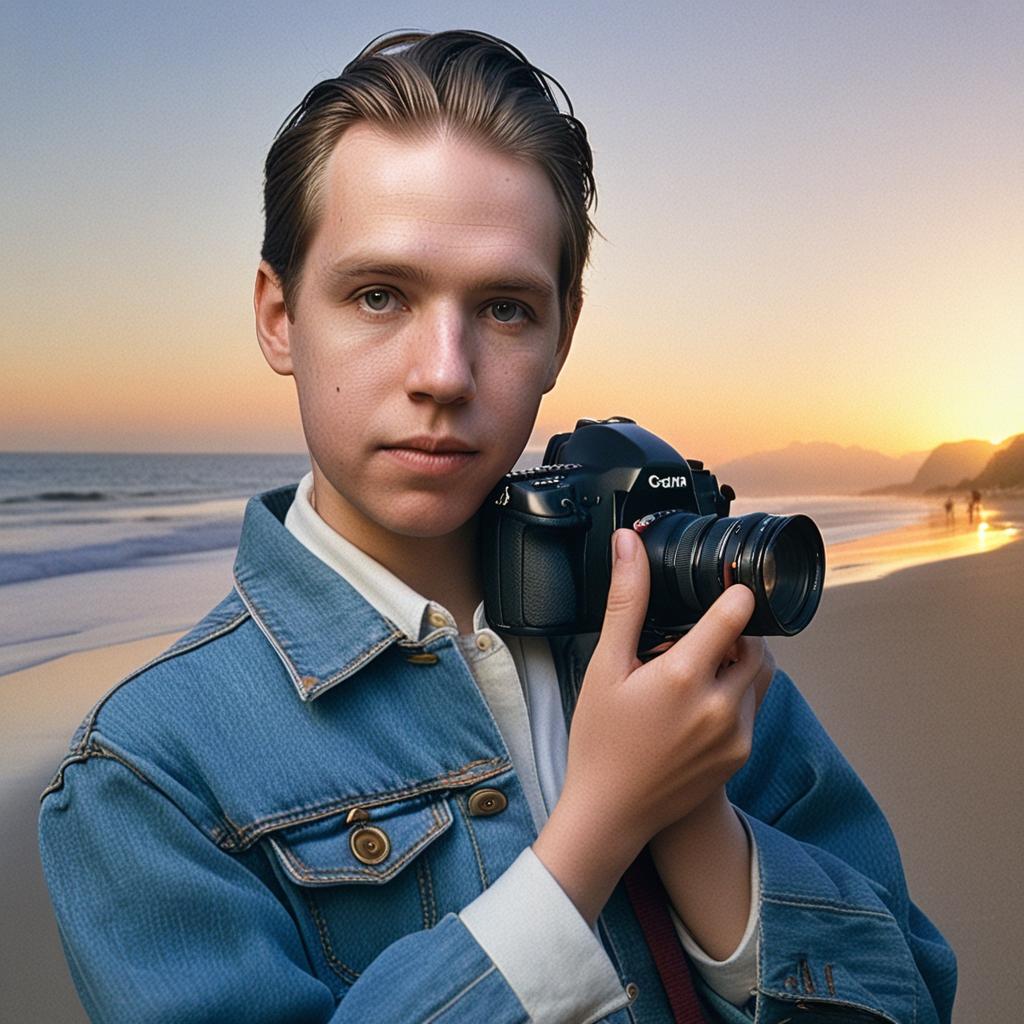} \\

        Input image &
        \begin{tabular}{c}Engineer using\\tablet to inspect\\smart factory,\\futuristic holograms,\\cyberpunk aesthetic\end{tabular}  &
        \begin{tabular}{c}Victorian-era\\gentleman reading\\under gas lamp in\\vintage library,\\sepia tone, oil\\painting strokes\end{tabular} &
        \begin{tabular}{c}An angular portrait\\of a youth in\\medieval armor,\\thick oil‑paint\\brushstrokes, dramatic\\chiaroscuro lighting\end{tabular} &
        \begin{tabular}{c}A middle-aged man\\in a suit leaning\\against a vintage\\car, smiling with\\arms crossed, under\\sunset city backdrop\end{tabular} &
        \begin{tabular}{c}A young photographer\\in a denim jacket,\\camera slung over\\his shoulder, sunset\\beach behind him\end{tabular} \\

        \includegraphics[width=0.15\textwidth]{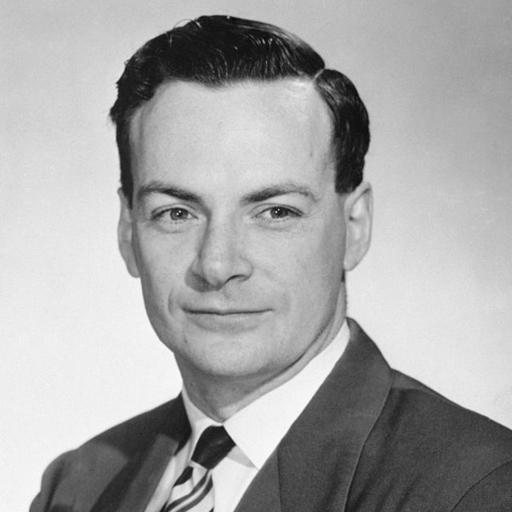} &
        \includegraphics[width=0.15\textwidth]{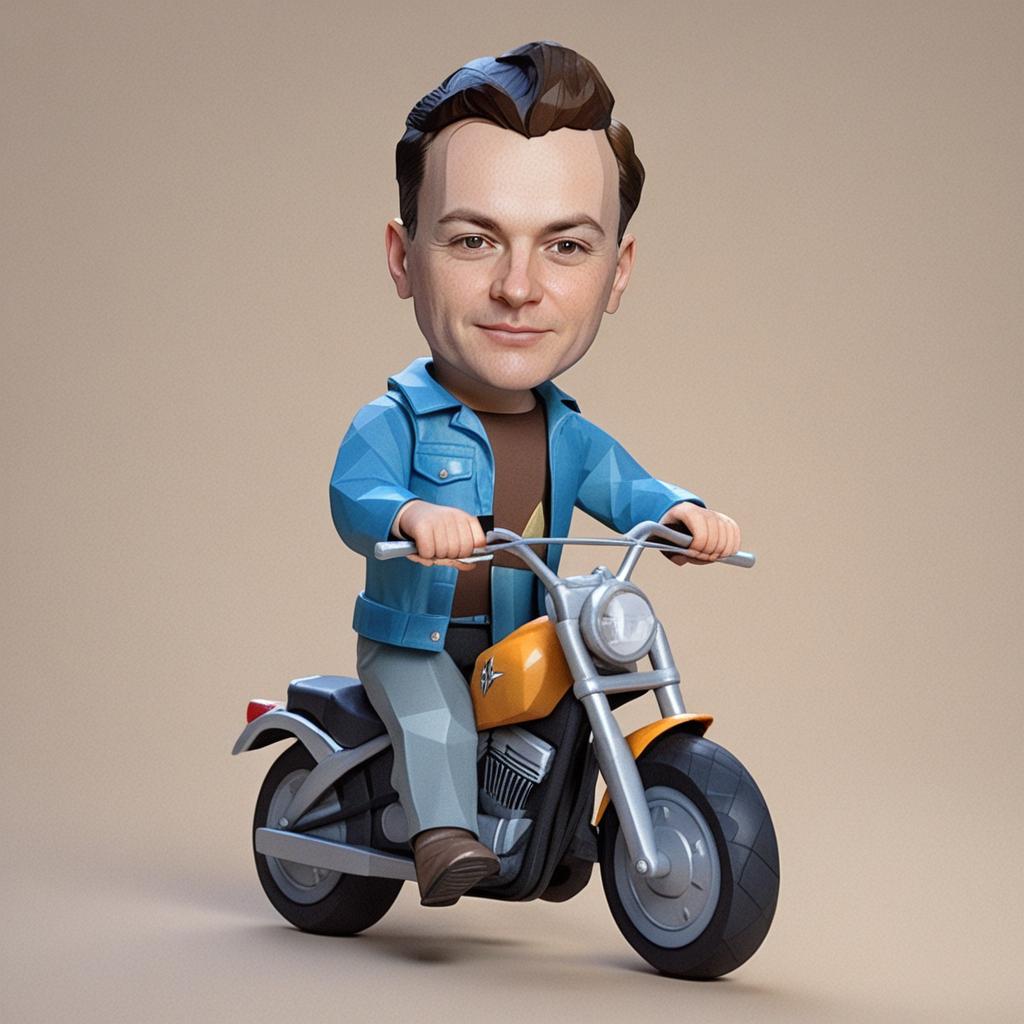} &
        \includegraphics[width=0.15\textwidth]{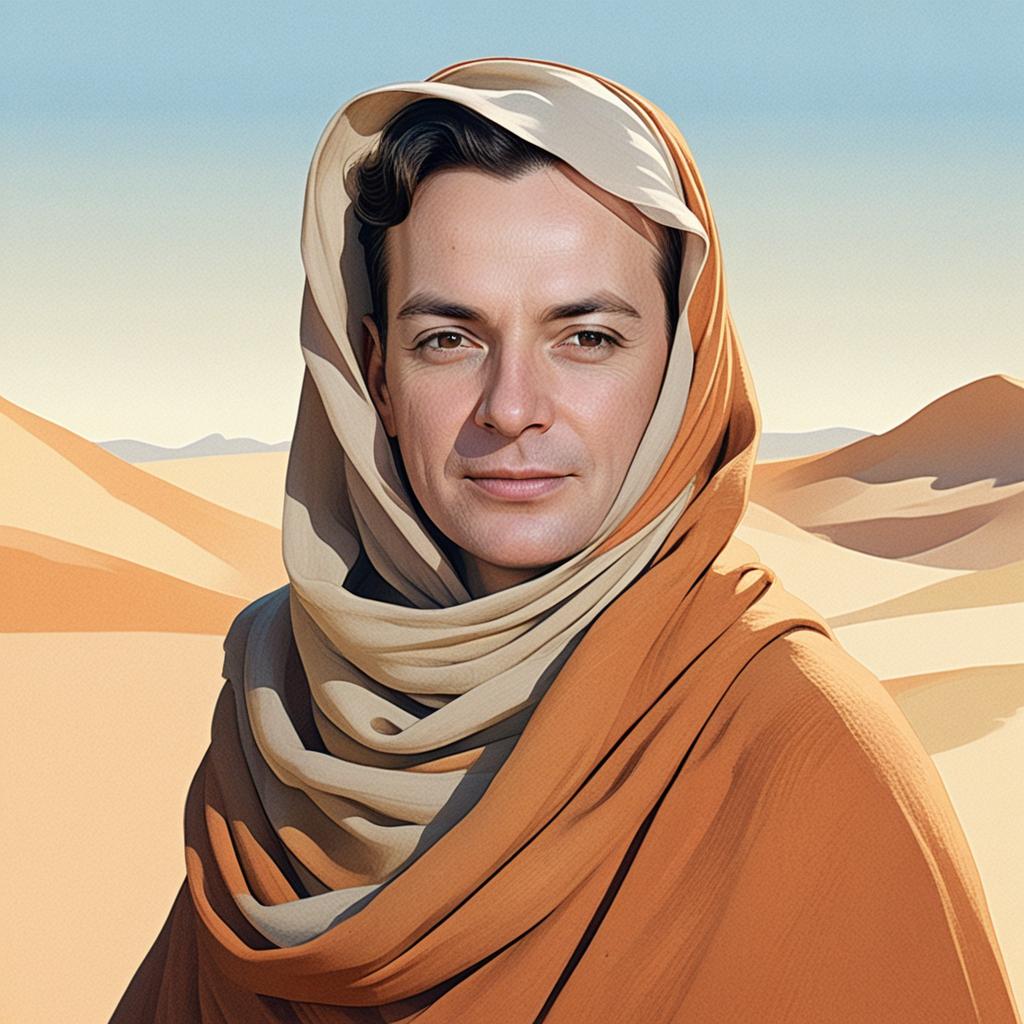} &
        \includegraphics[width=0.15\textwidth]{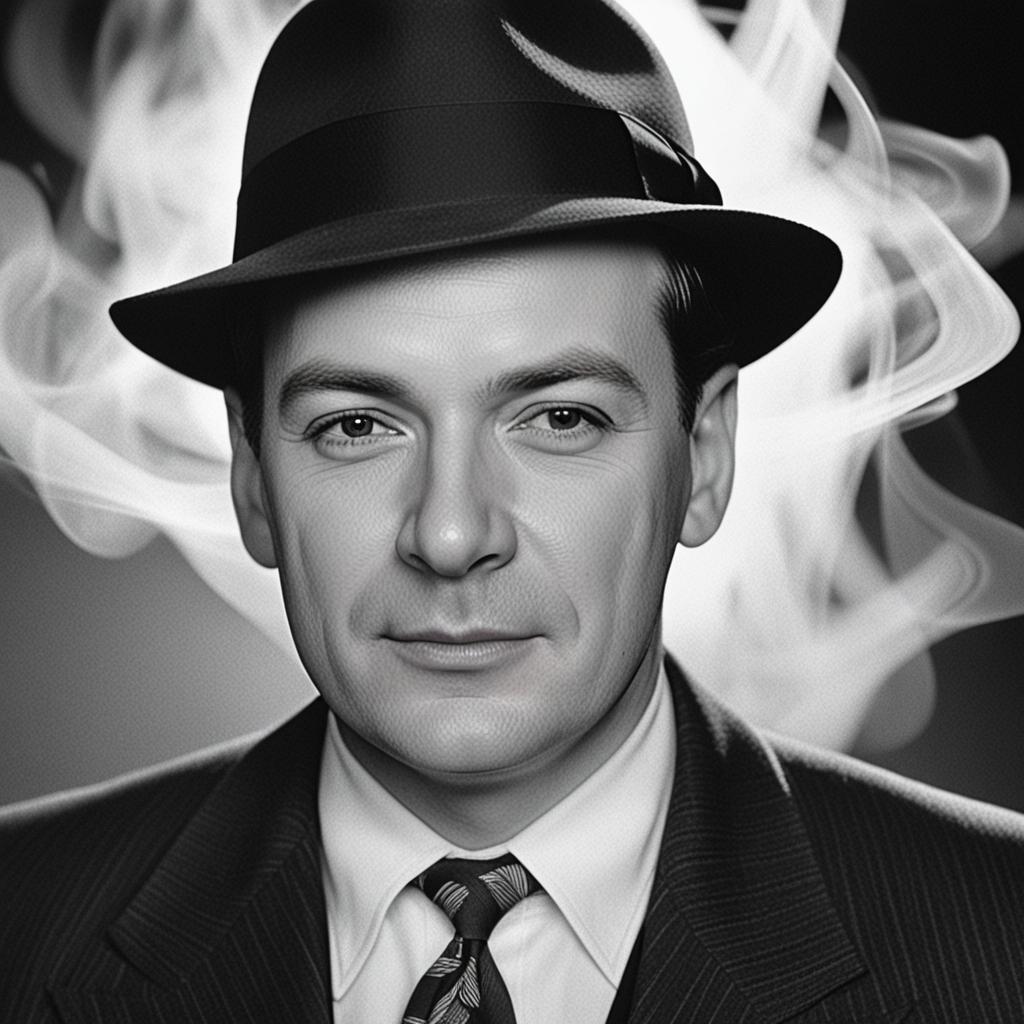} &
        \includegraphics[width=0.15\textwidth]{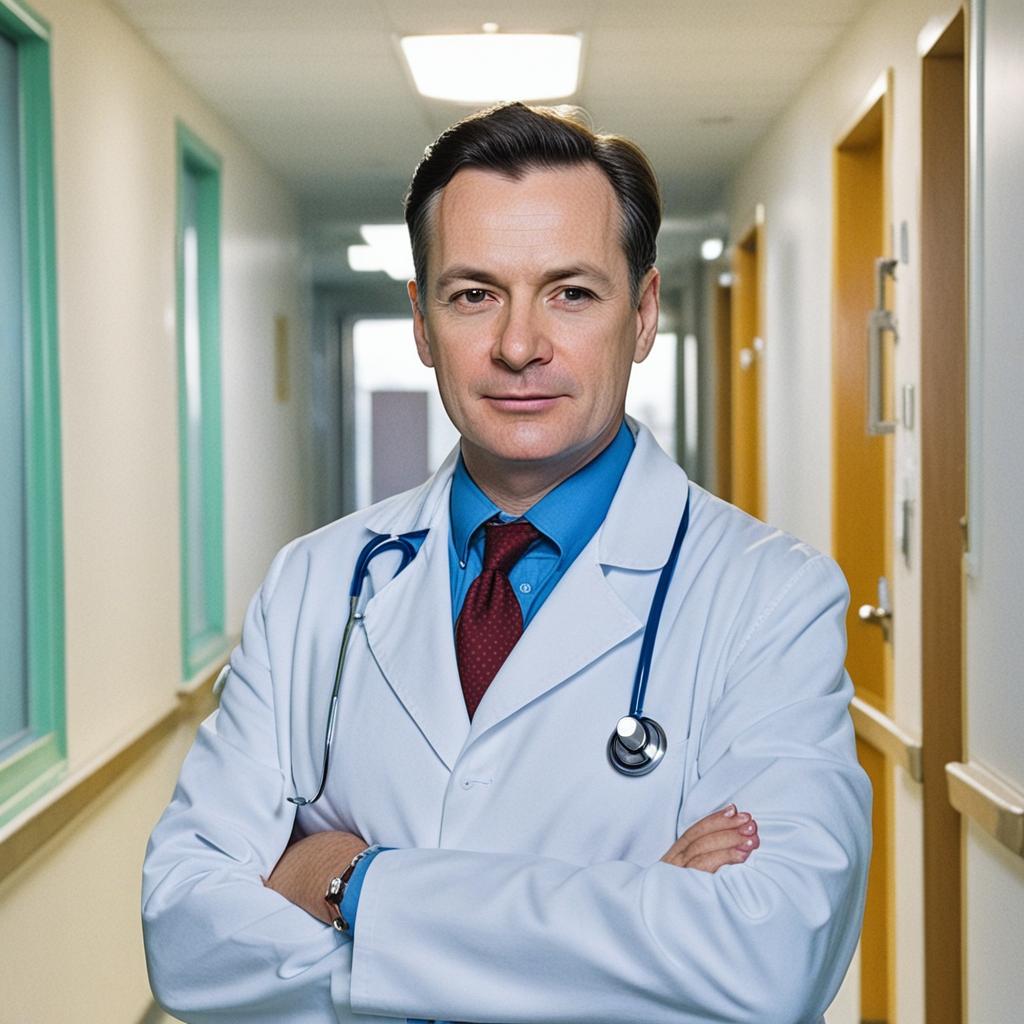} &
        \includegraphics[width=0.15\textwidth]{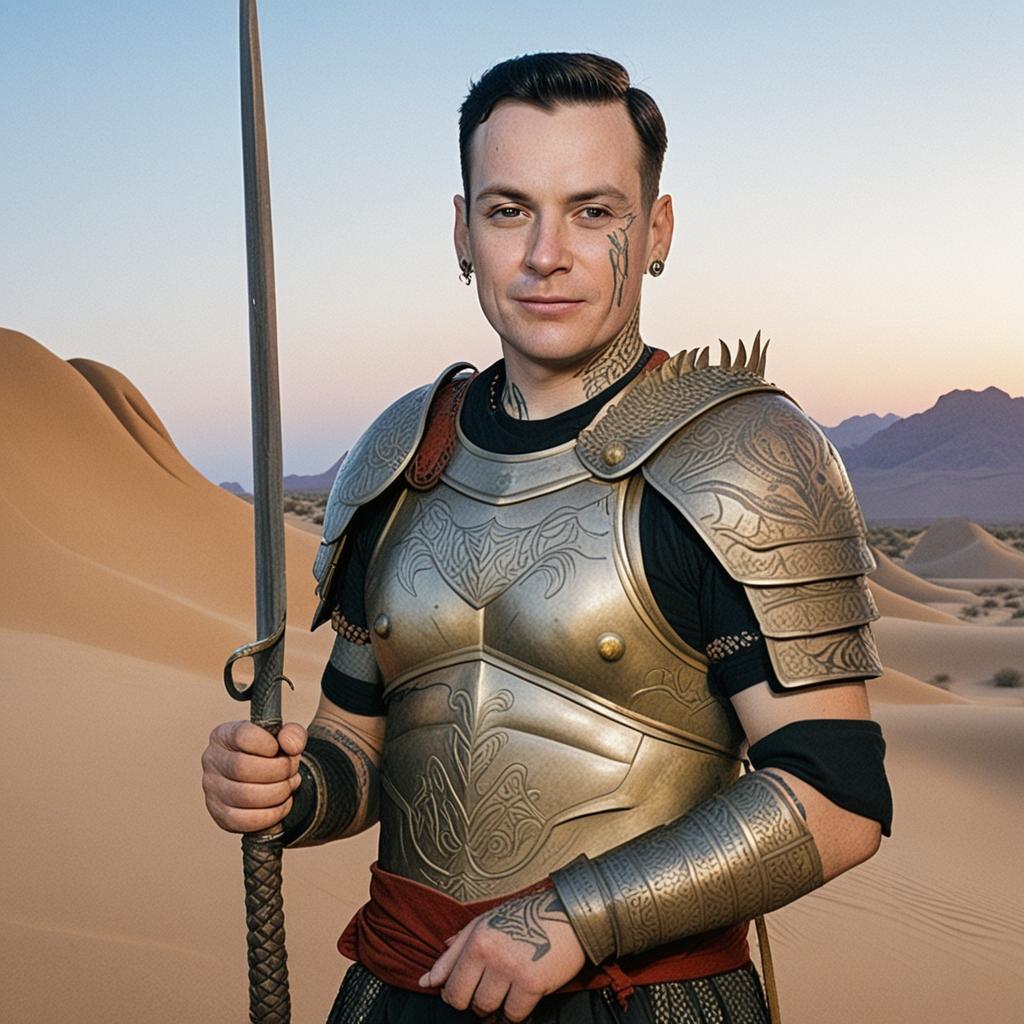} \\

        Input image &
        \begin{tabular}{c}A low poly 3D\\render of a\\young biker\end{tabular} &
        \begin{tabular}{c}A desert wanderer\\in a wind-blown\\cloak and headscarf,\\sun-scorched palette,\\stylized concept\\illustration\end{tabular} &
        \begin{tabular}{c}A film noir portrait\\of a detective in\\a fedora, high‑contrast\\black‑and‑white grain,\\smoke curling around\end{tabular} &
        \begin{tabular}{c}A middle-aged doctor\\in a white coat with\\a stethoscope, calm\\expression, standing\\in a bright hospital\\corridor\end{tabular} &
        \begin{tabular}{c}A warrior with tribal\\tattoos and a battle-\\worn armor, holding\\a sword, standing\\tall in a desert\\at dusk\end{tabular}\\     

        \includegraphics[width=0.15\textwidth]{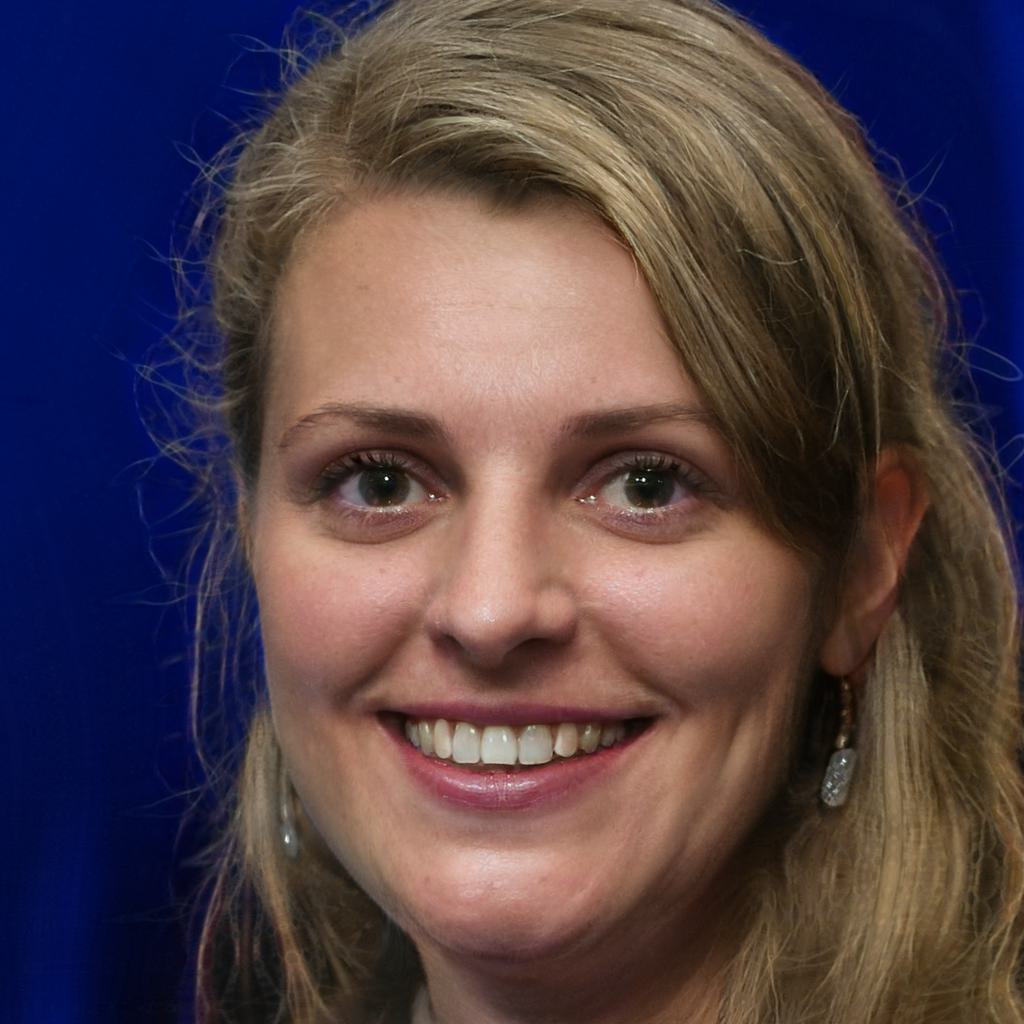} &
        \includegraphics[width=0.15\textwidth]{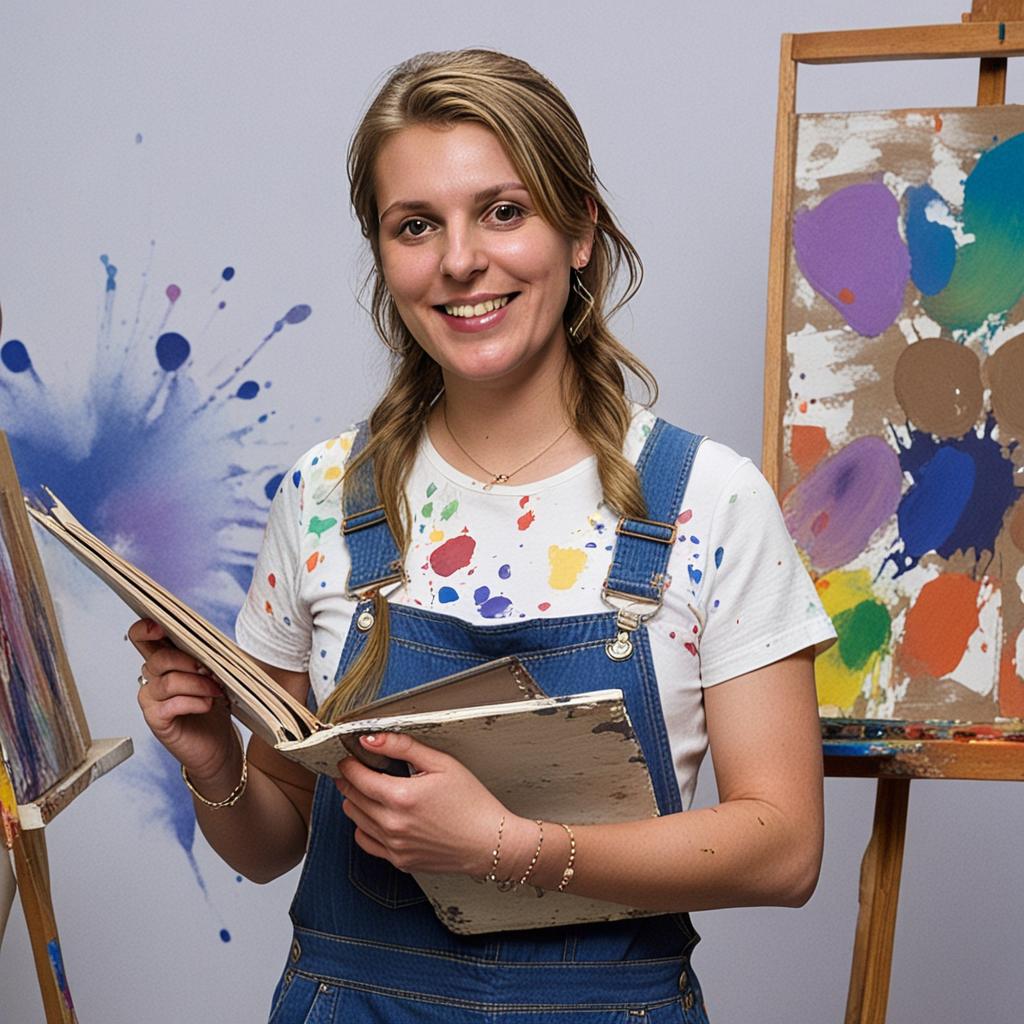} &
        \includegraphics[width=0.15\textwidth]{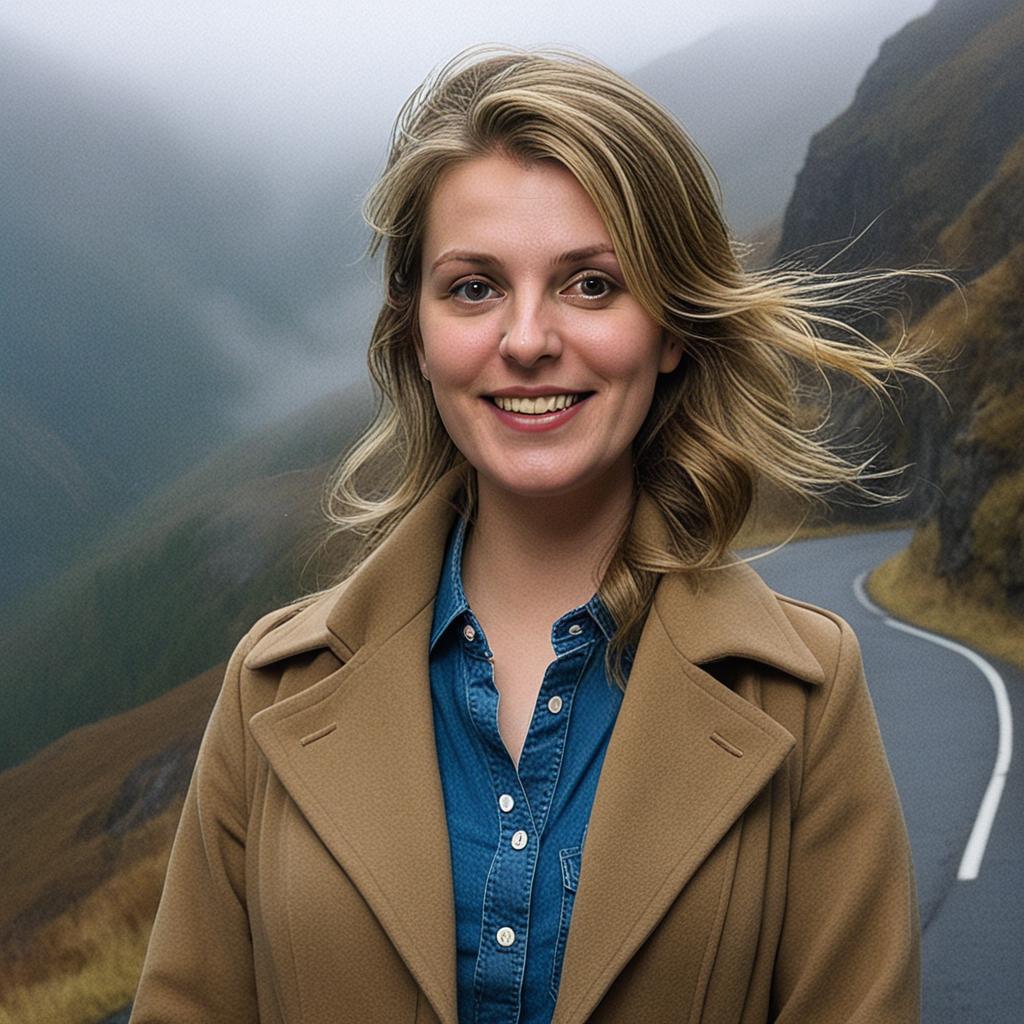} &
        \includegraphics[width=0.15\textwidth]{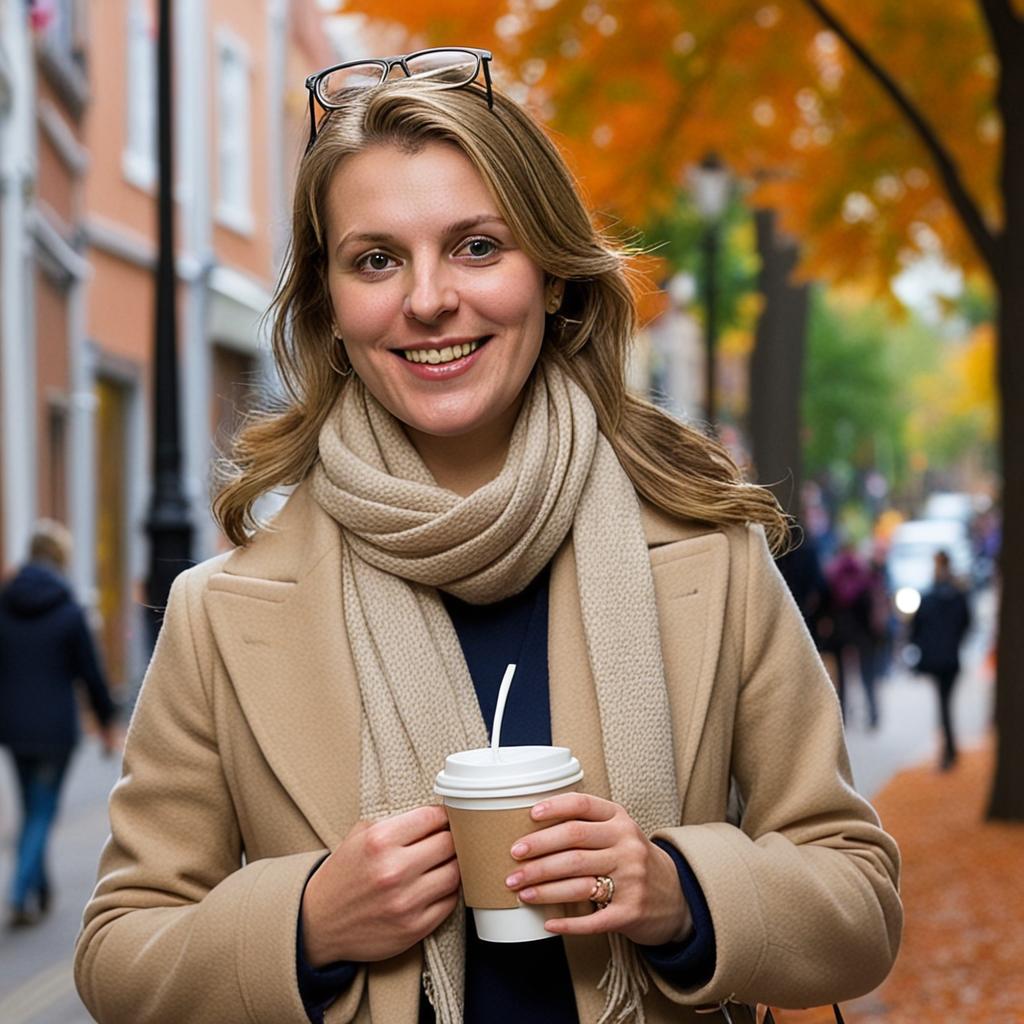} &
        \includegraphics[width=0.15\textwidth]{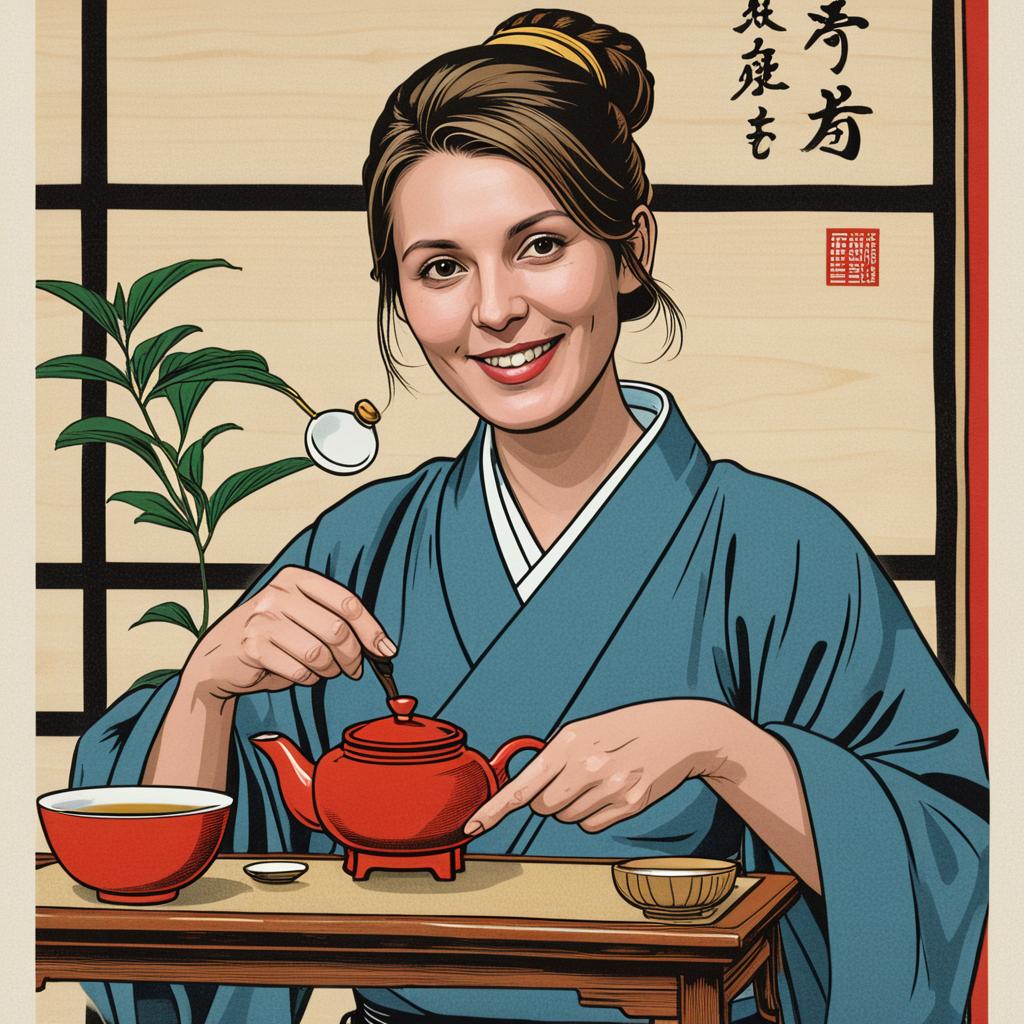} &
        \includegraphics[width=0.15\textwidth]{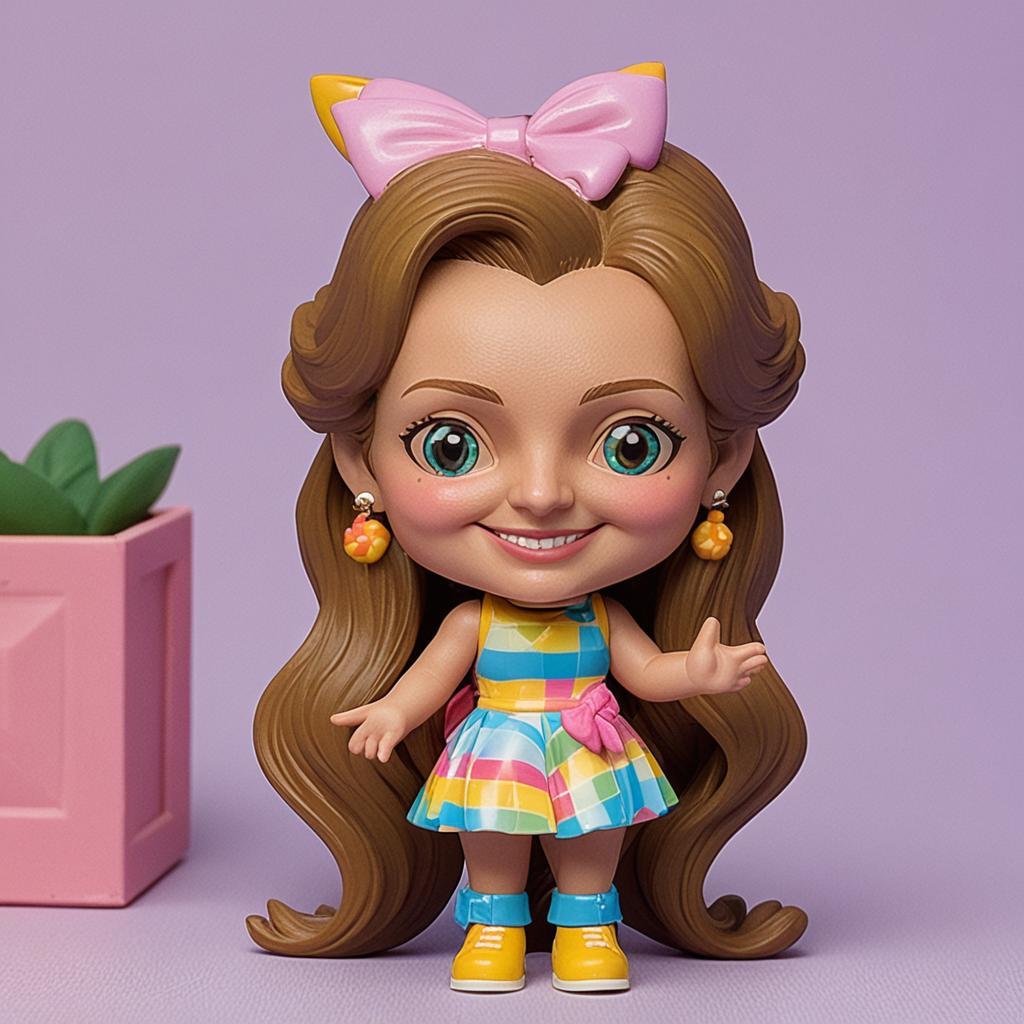} \\

        Input image &
        \begin{tabular}{c}A young artist in\\overalls with paint\\stains, holding a\\sketchbook, studio\\background with\\canvases\end{tabular} &
        \begin{tabular}{c}A young woman\\with windswept hair,\\wearing a long coat\\and boots, standing\\on a foggy\\mountain road,\\cinematic mood\end{tabular} &
        \begin{tabular}{c}A woman in a\\soft wool coat\\with a scarf,\\holding a paper\\cup of coffee,\\walking along an\\autumn street\end{tabular} &
        \begin{tabular}{c}A traditional\\woodblock style\\image of a tea master,\\with clean-lines and\\deliberate simplicity\\honoring Japanese\\ukiyo-e\end{tabular} &
        \begin{tabular}{c}Popmart blind\\box\end{tabular}\\     

        \includegraphics[width=0.15\textwidth]{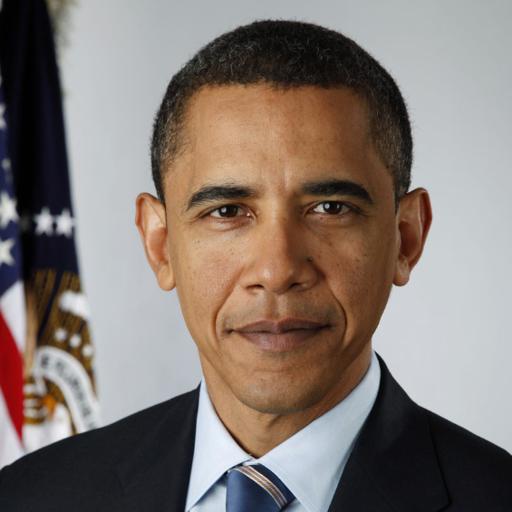} &
        \includegraphics[width=0.15\textwidth]{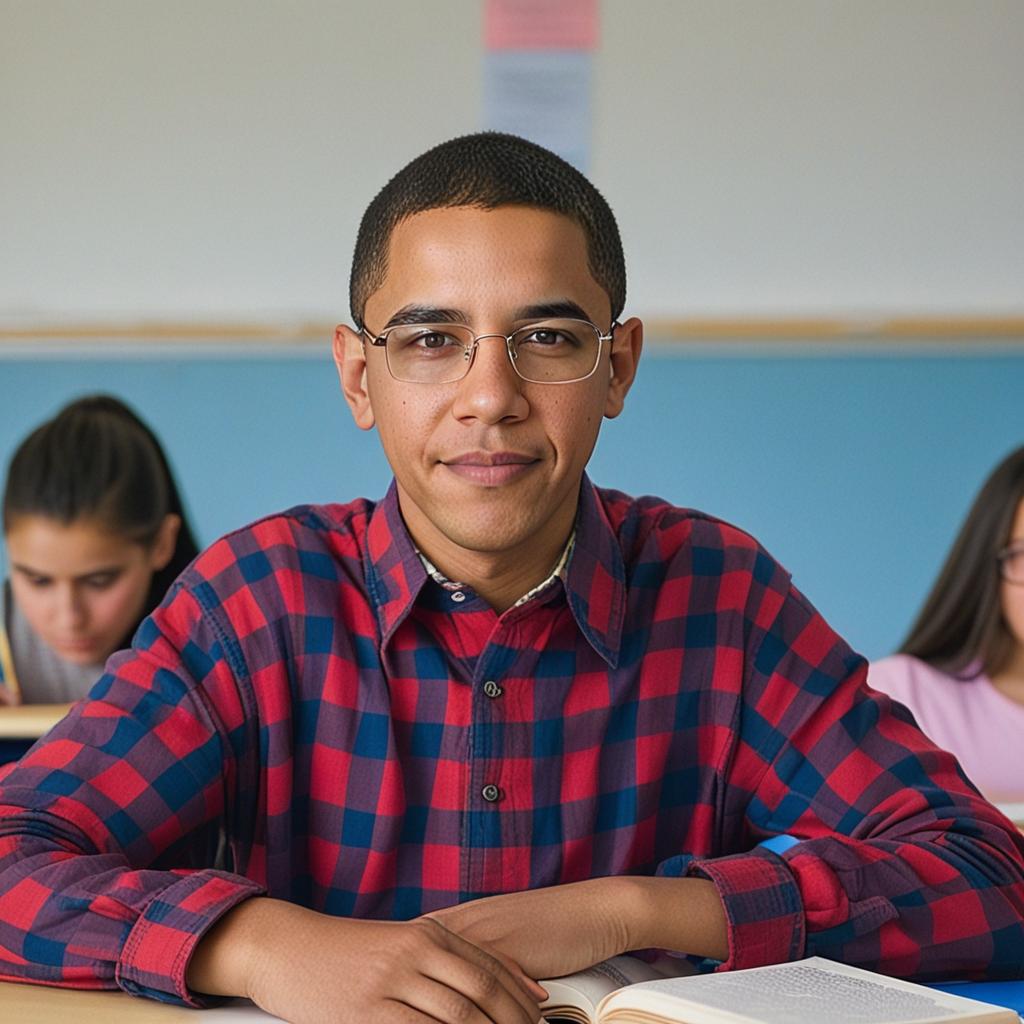} &
        \includegraphics[width=0.15\textwidth]{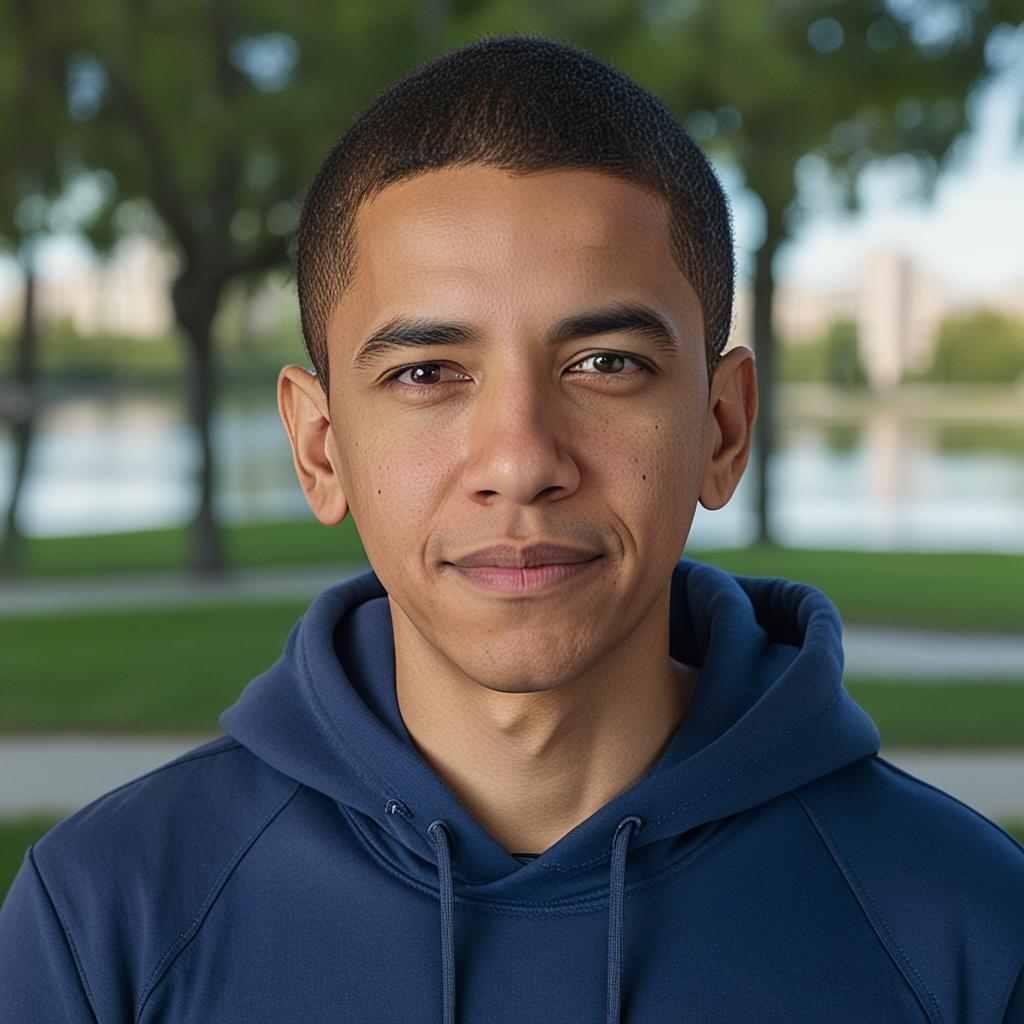} &
        \includegraphics[width=0.15\textwidth]{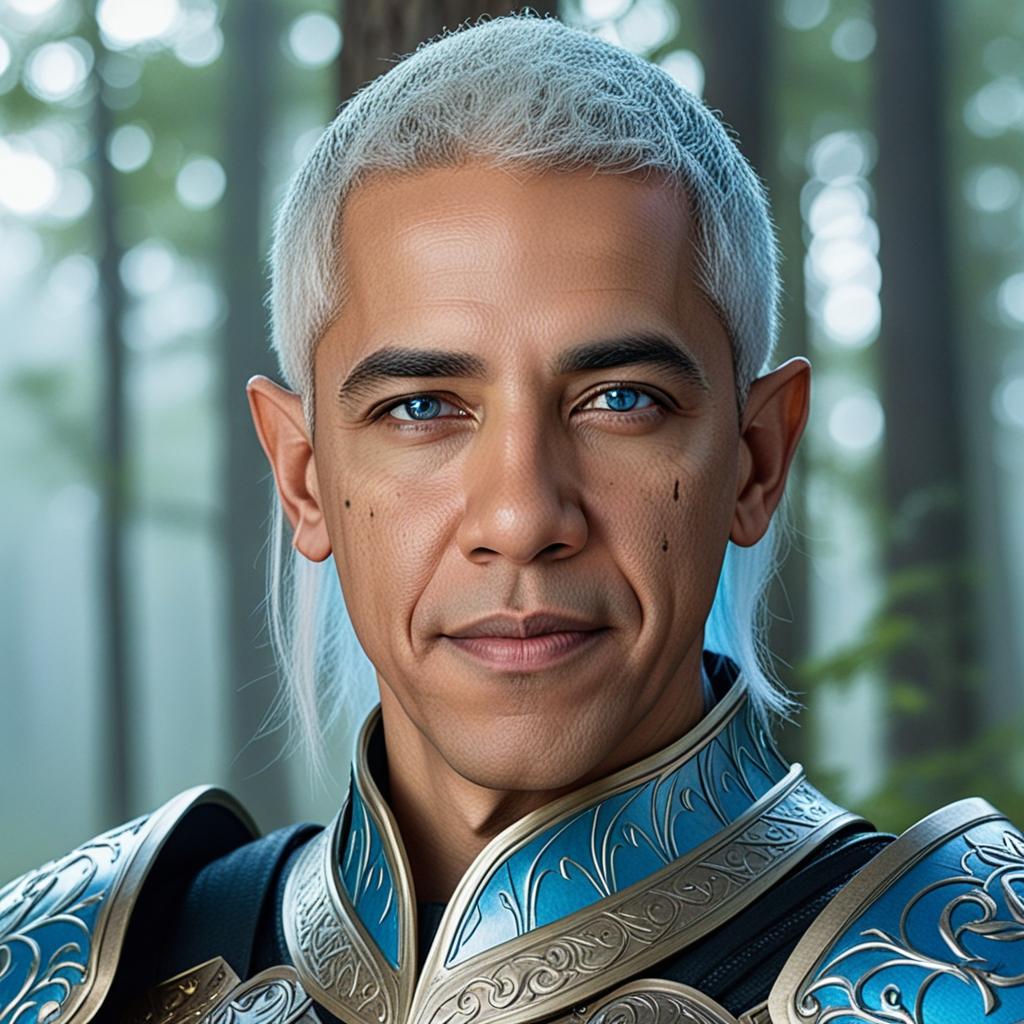} &
        \includegraphics[width=0.15\textwidth]{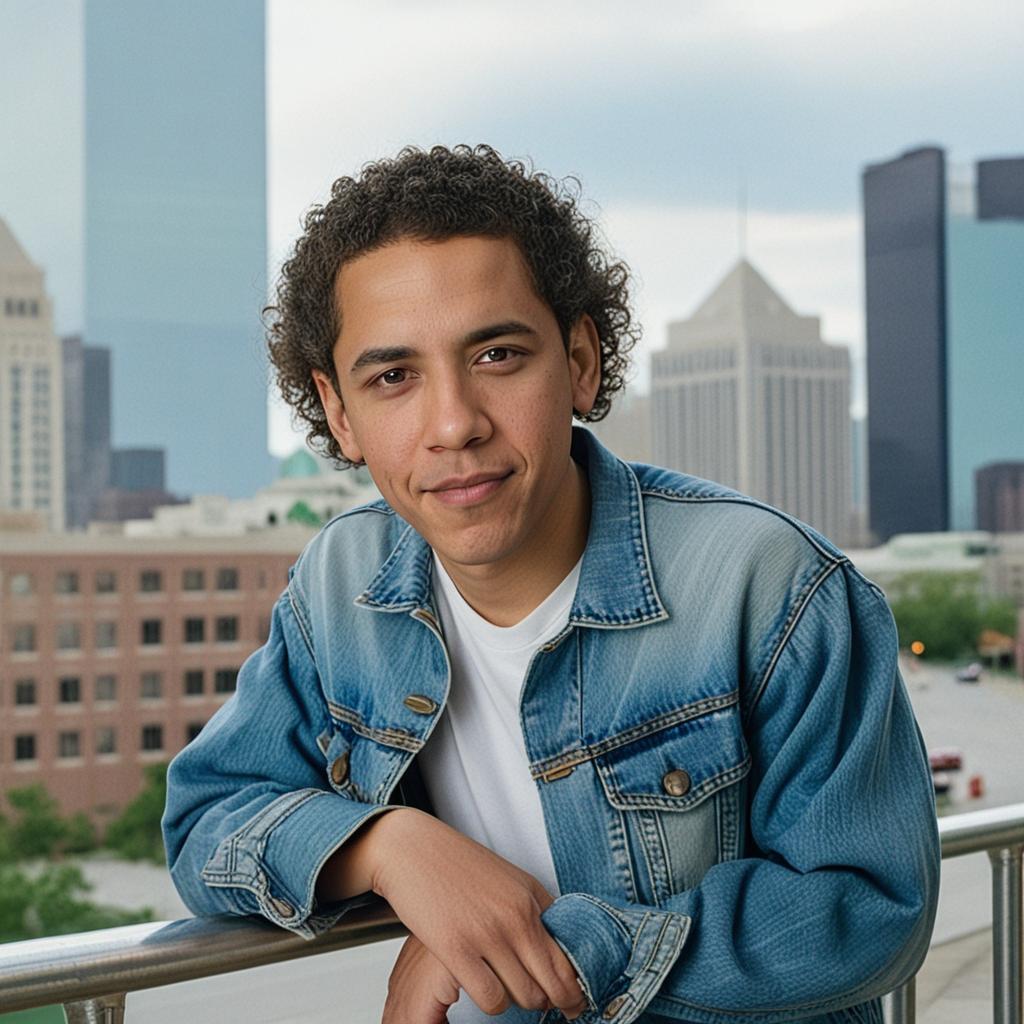} &
        \includegraphics[width=0.15\textwidth]{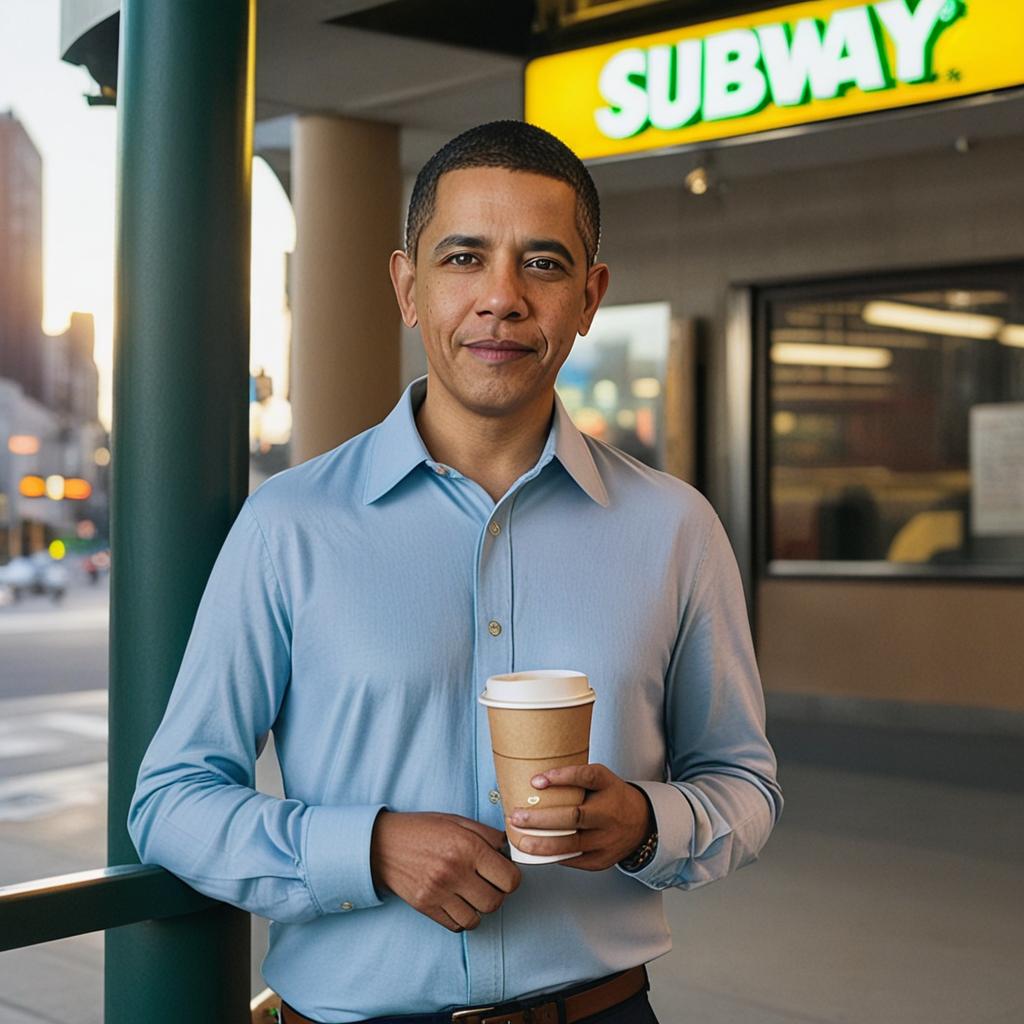} \\

        Input image &
        \begin{tabular}{c}A college student\\wearing glasses and\\a checkered shirt,\\sitting at a desk\\with books, classroom\\environment\end{tabular} &
        \begin{tabular}{c}A young man in\\a navy blue hoodie,\\looking calmly at\\the camera, natural\\lighting, city park\\background\end{tabular} &
        \begin{tabular}{c}A fantasy elf warrior\\with white hair and\\glowing blue eyes,\\ornate armor, misty\\forest background,\\epic concept art style\end{tabular} &
        \begin{tabular}{c}A man in his\\twenties with curly\\hair, wearing a\\denim jacket, leaning\\against a railing with\\city buildings behind\end{tabular} &
        \begin{tabular}{c}A man in casual wear\\holding a takeaway\\coffee, standing near\\a subway entrance,\\early morning light.\end{tabular}\\     

        \includegraphics[width=0.15\textwidth]{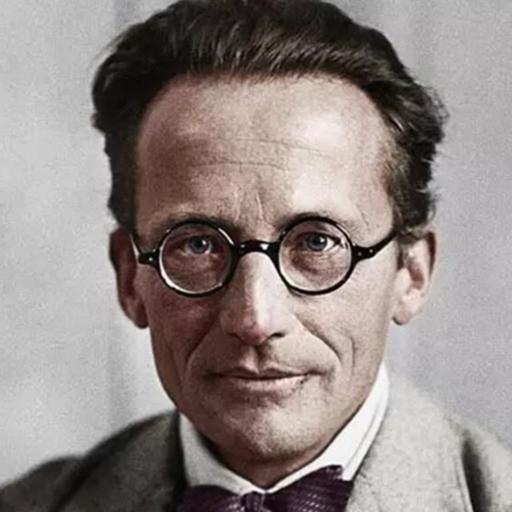} &
        \includegraphics[width=0.15\textwidth]{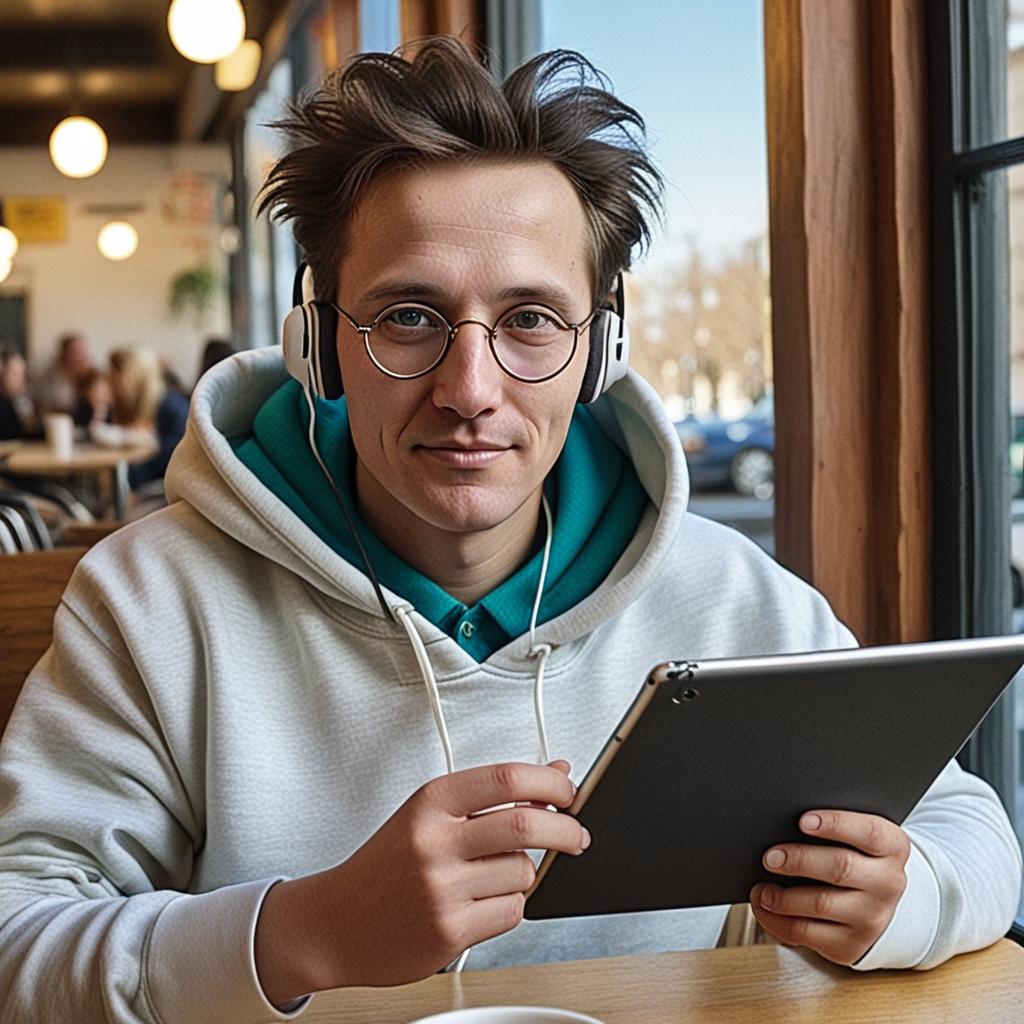} &
        \includegraphics[width=0.15\textwidth]{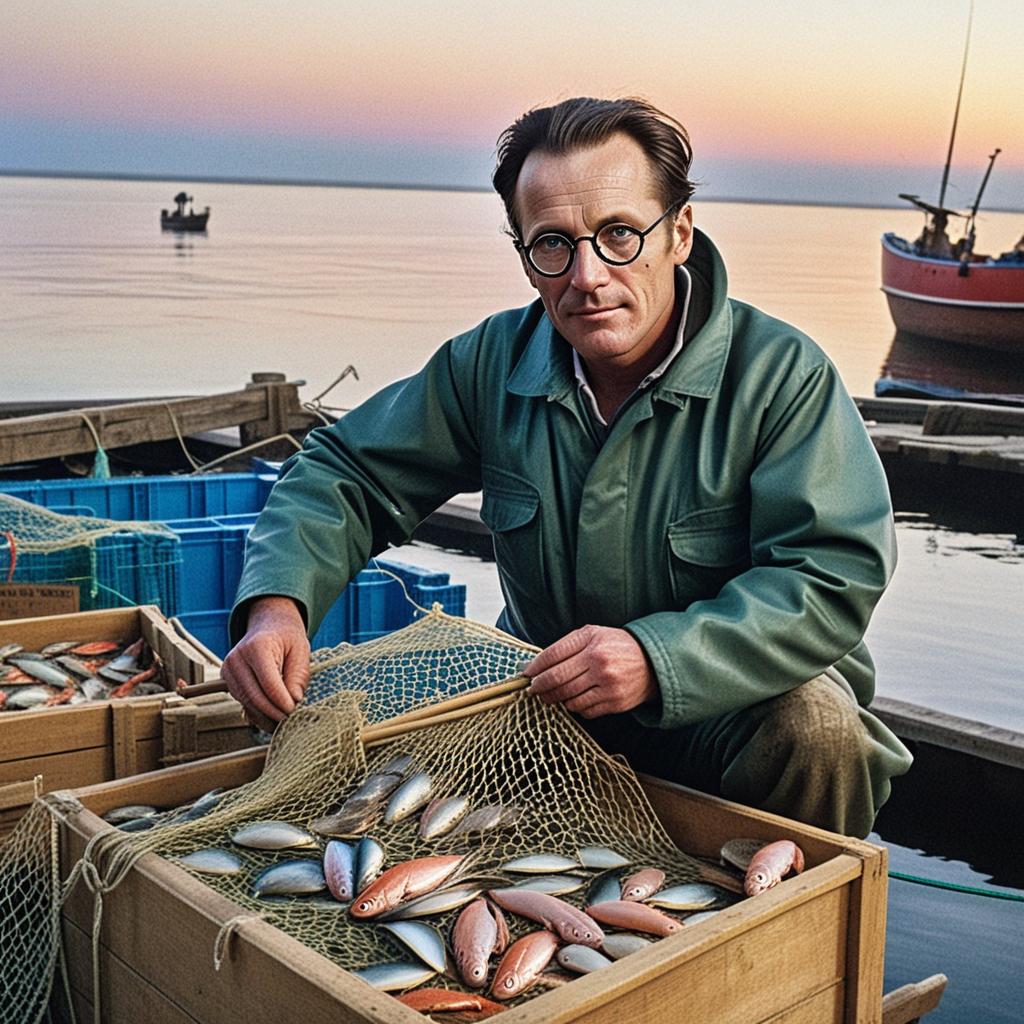} &
        \includegraphics[width=0.15\textwidth]{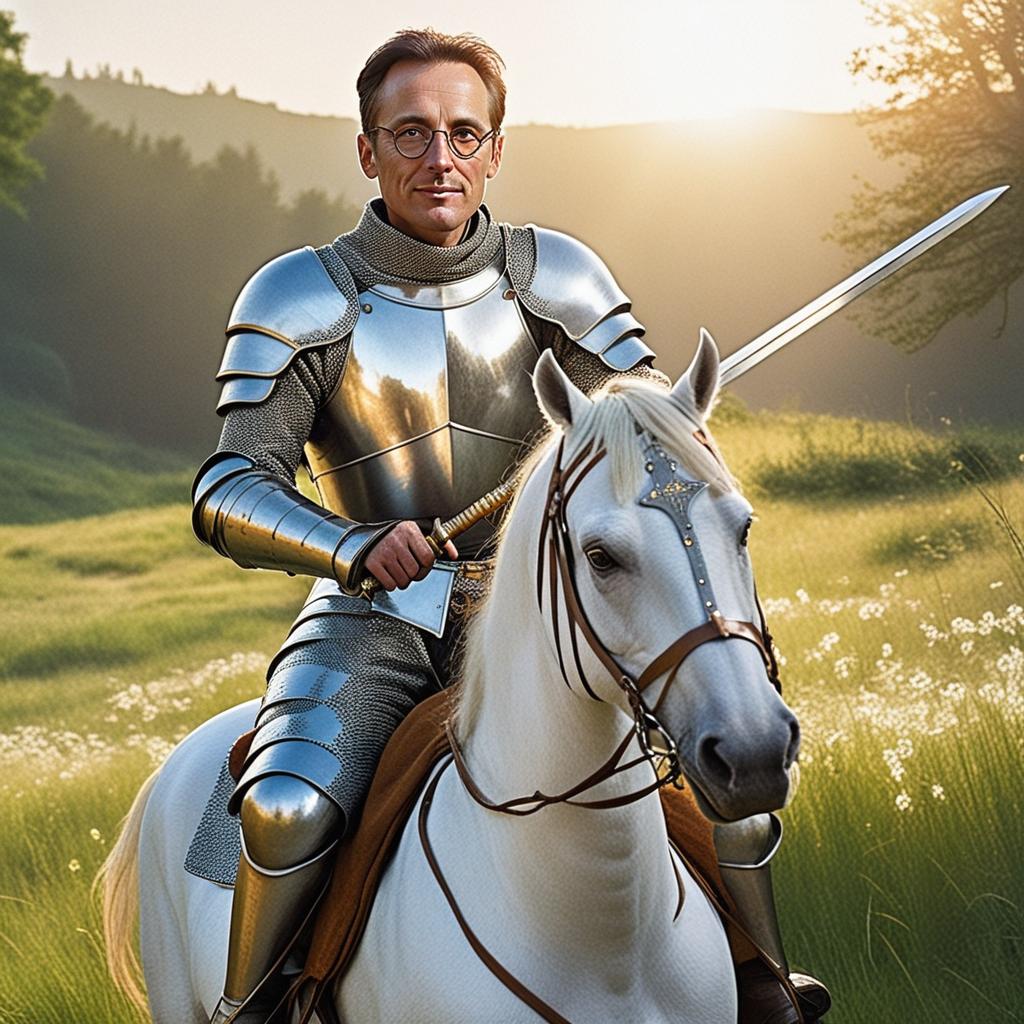} &
        \includegraphics[width=0.15\textwidth]{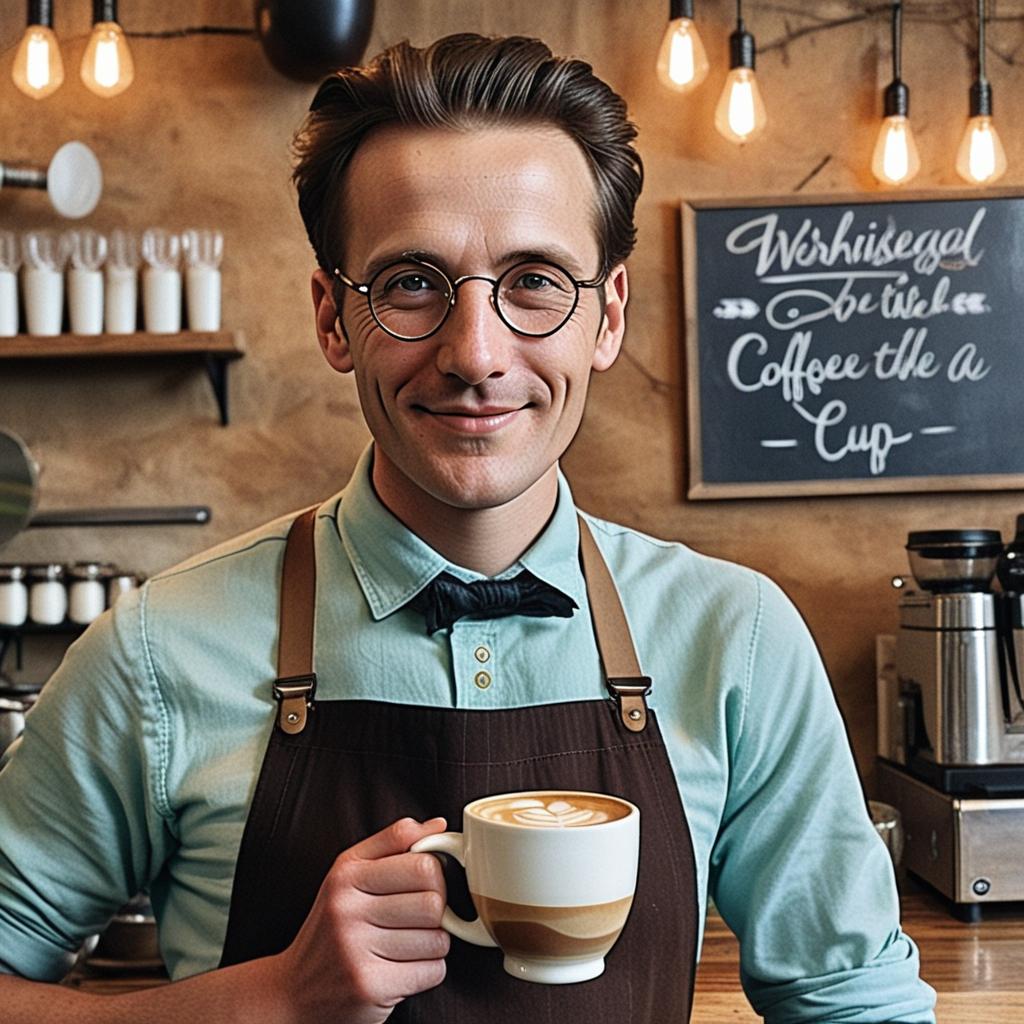} &
        \includegraphics[width=0.15\textwidth]{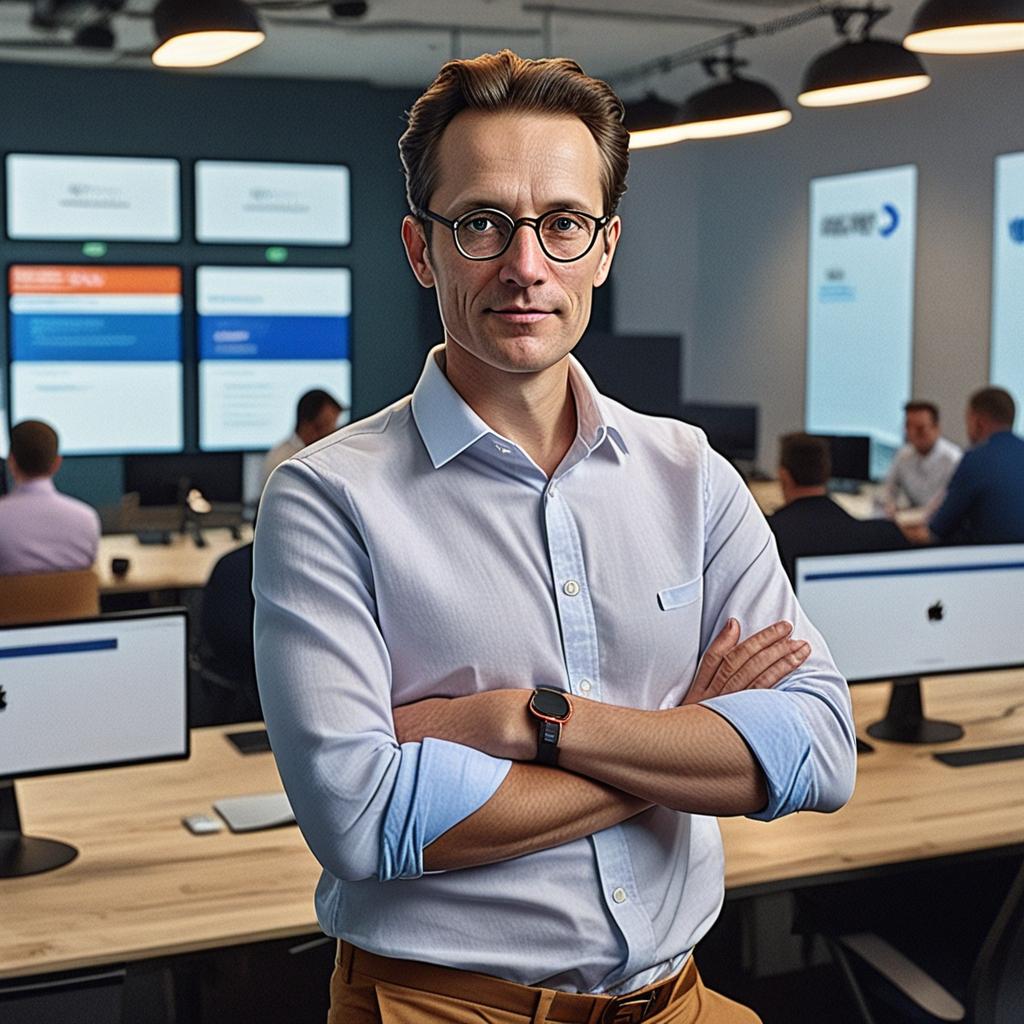} \\

        Input image &
        \begin{tabular}{c}A college student\\with tousled hair,\\wearing headphones\\and a hoodie,\\holding a tablet in a\\cafe setting\end{tabular} &
        \begin{tabular}{c}A fisherman mending\\a net on a wooden\\dock at dawn, wearing\\a waterproof jacket\\and surrounded by\\crates of fresh seafood\end{tabular} &
        \begin{tabular}{c}A knight in shining\\armor, riding a white\\steed across a sunlit\\meadow, holding a\\sword in a medieval\\fantasy realm\end{tabular} &
        \begin{tabular}{c}A young barista\\with a friendly\\grin, holding a\\latte art cup,\\standing inside a\\rustic coffee bar\end{tabular} &
        \begin{tabular}{c}A tech entrepreneur\\in a crisp shirt,\\arms folded, modern\\co‑working space with\\screens behind, realistic.\end{tabular}\\     
    \end{tabular}
    }
    \caption{Additional qualitative results by UniID.}
    \label{fig:appendix_realvision}
    \vspace{-7pt}
\end{figure*}

To further demonstrate the effectiveness of our approach, we provide additional qualitative results in Figure~\ref{fig:appendix_realvision}. As shown in Figure~\ref{fig:appendix_realvision}, our method consistently produces high-fidelity personalized images that maintain strong identity preservation while accurately following the text prompts. The results span a wide range of challenging scenarios.

\section{Qualitative Comparison based on SDXL}
\label{sec:appendix_all_sdxl}
\begin{figure*}
    \centering
    \includegraphics[width=0.8\textwidth]{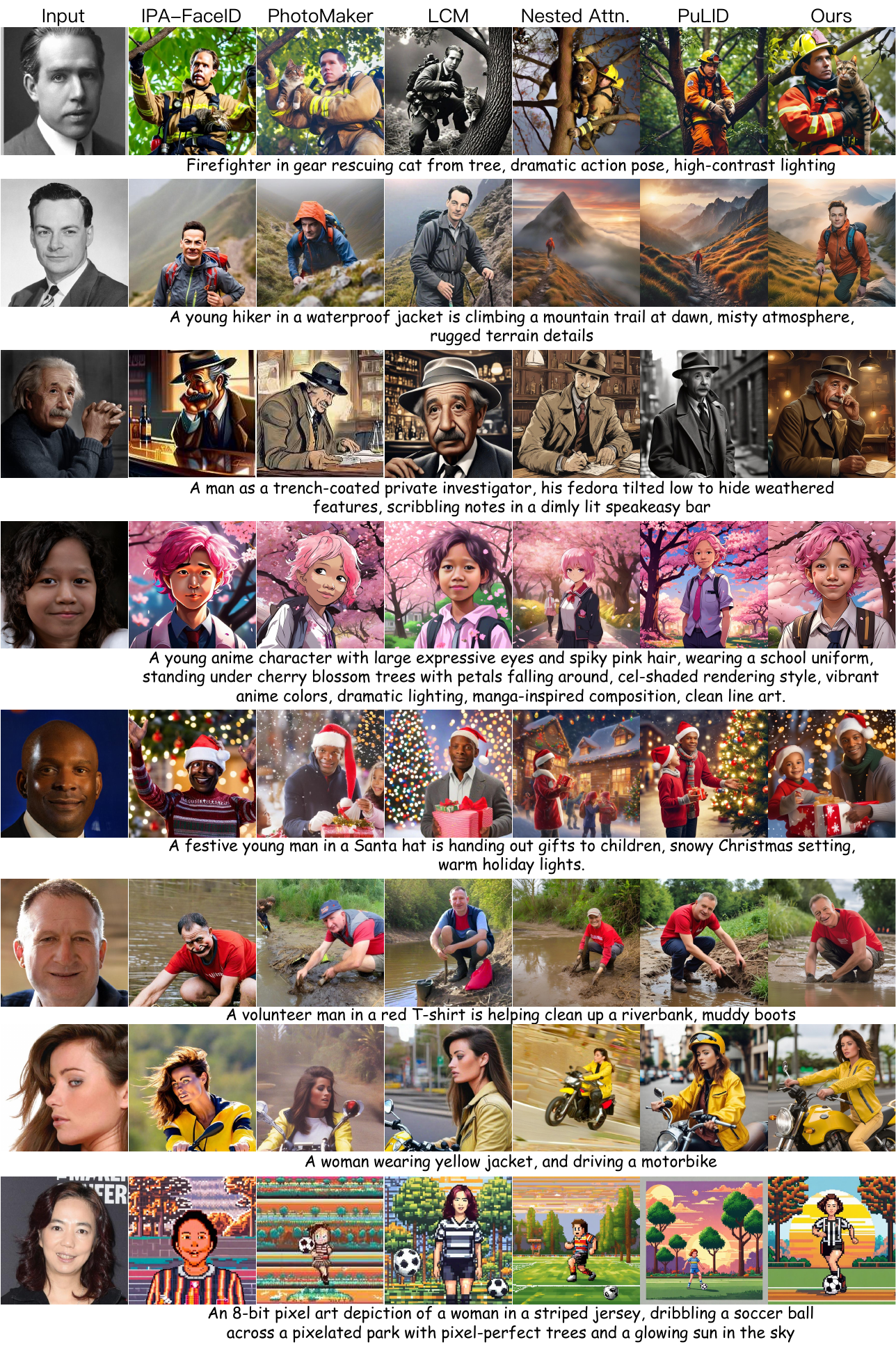} 
    \caption{\textbf{Additional qualitative comparison}. We compare our method with five baseline methods using SDXL as the base model, including IPA-FaceID~\cite{ipa}, PhotoMaker~\cite{li2023photomaker}, LCM~\cite{lcm}, Nested Attention~\cite{nested}, and PuLID~\cite{pulid}.}
    \label{fig:appendix_qualitative_comparison} 
\end{figure*}

To ensure a fair and controlled evaluation, we conduct a comprehensive comparison using SDXL~\cite{sdxl} as the unified backbone model across all methods. We compare our approach against five state-of-the-art methods, including IPA-FaceID~\cite{ipa}, PhotoMaker~\cite{li2023photomaker}, LCM~\cite{lcm}, Nested Attention~\cite{nested}, and PuLID~\cite{pulid}.
Figure~\ref{fig:appendix_qualitative_comparison} presents a visual comparison of the results. Our findings are consistent with the observations reported in Section~\ref{sec:results}. Specifically, IPA-FaceID and PhotoMaker demonstrate substantial deficiencies in identity preservation, often failing to preserve critical facial characteristics from the reference image. LCM exhibits limited text controllability and produces outputs that appear blurry. Nested Attention similarly struggles with prompt alignment, with particularly pronounced degradation in stylization scenarios. PuLID shows inconsistent identity preservation, with notable failures in full-body generation scenarios. In contrast, our method consistently produces high-fidelity results that simultaneously achieve robust identity preservation and accurate alignment with the text prompts.

\section{Comparison with FLUX.1 Kontext}
\label{sec:appendix_kontext}
\begin{figure*}
    \centering
    \includegraphics[width=0.88\textwidth]{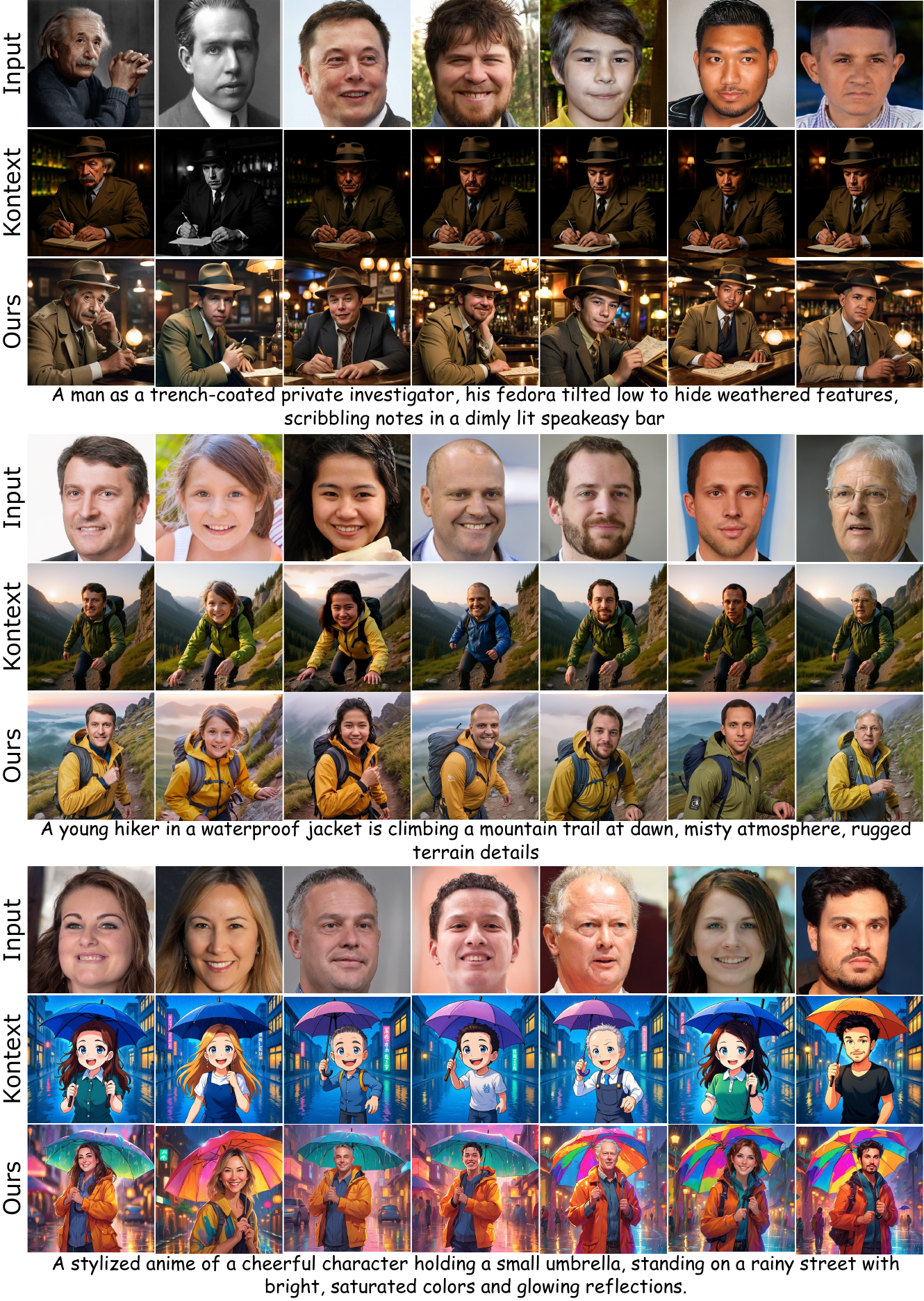} 
    \caption{\textbf{Comparison with FLUX.1 Kontext}. As an image editing model, Kontext often suffers from severe identity inconsistency (top, bottom) and disproportionate facial scaling (middle) when the generated image requires substantial structural changes from the input image.}
    \label{fig:appendix_kontext} 
\end{figure*}

In this section, we compare our method against FLUX.1 Kontext~\cite{kontext}, a state-of-the-art image editing model. Figure~\ref{fig:appendix_kontext} presents a visual comparison between the two approaches.
As an image editing model, FLUX.1 Kontext excels at tasks where the overall structure remains largely consistent with the input image. It demonstrates particularly strong performance in local modifications, such as adding, removing, or modifying objects. However, personalization tasks often require more substantial structural changes to the composition, which poses challenges for editing-based approaches.
As illustrated in Figure~\ref{fig:appendix_kontext}, when the generated image diverges significantly from the input in terms of overall structure, such as transforming a close-up portrait into a full-body shot or generating heavily stylized images, FLUX.1 Kontext exhibits notable limitations. Specifically, we observe severe identity consistency issues and disproportionate facial scaling, where faces occupy an inappropriately large portion relative to the body.
We hypothesize that these limitations stem from the training data distribution of image editing models. Most training pairs in such datasets consist of images with similar overall structures, focusing primarily on local variations. Consequently, when provided with a portrait as input, FLUX.1 Kontext displays a strong inductive bias toward generating outputs where the face dominates the composition or fails to preserve the identity of the reference face. In contrast, our method, designed specifically for personalization rather than editing, demonstrates superior flexibility in adapting to diverse compositional requirements while maintaining robust identity consistency.

\section{Effect of Varying Rescaling Weights}
\label{sec:appendix_rescaling}
\begin{figure*}
    \centering
    \renewcommand{\arraystretch}{0.3}
    \setlength{\tabcolsep}{0.5pt}

    {\normalsize
    \begin{tabular}{c c c c c c c}
        Input&&&&&&\\
        \includegraphics[width=0.139\textwidth]{images/rescaling_weight_2/text/input.jpg} &
        \raisebox{0.03\textwidth}{$\alpha=0$} & 
        \raisebox{0.03\textwidth}{$\alpha=0.5$} & 
        \raisebox{0.03\textwidth}{$\alpha=1$} & 
        \raisebox{0.03\textwidth}{$\alpha=1.5$} & 
        \raisebox{0.03\textwidth}{$\alpha=2$} & 
        \raisebox{0.03\textwidth}{$\alpha=2.5$}  \\

        \raisebox{0.06\textwidth}{$\beta=0$} &
        \includegraphics[width=0.139\textwidth]{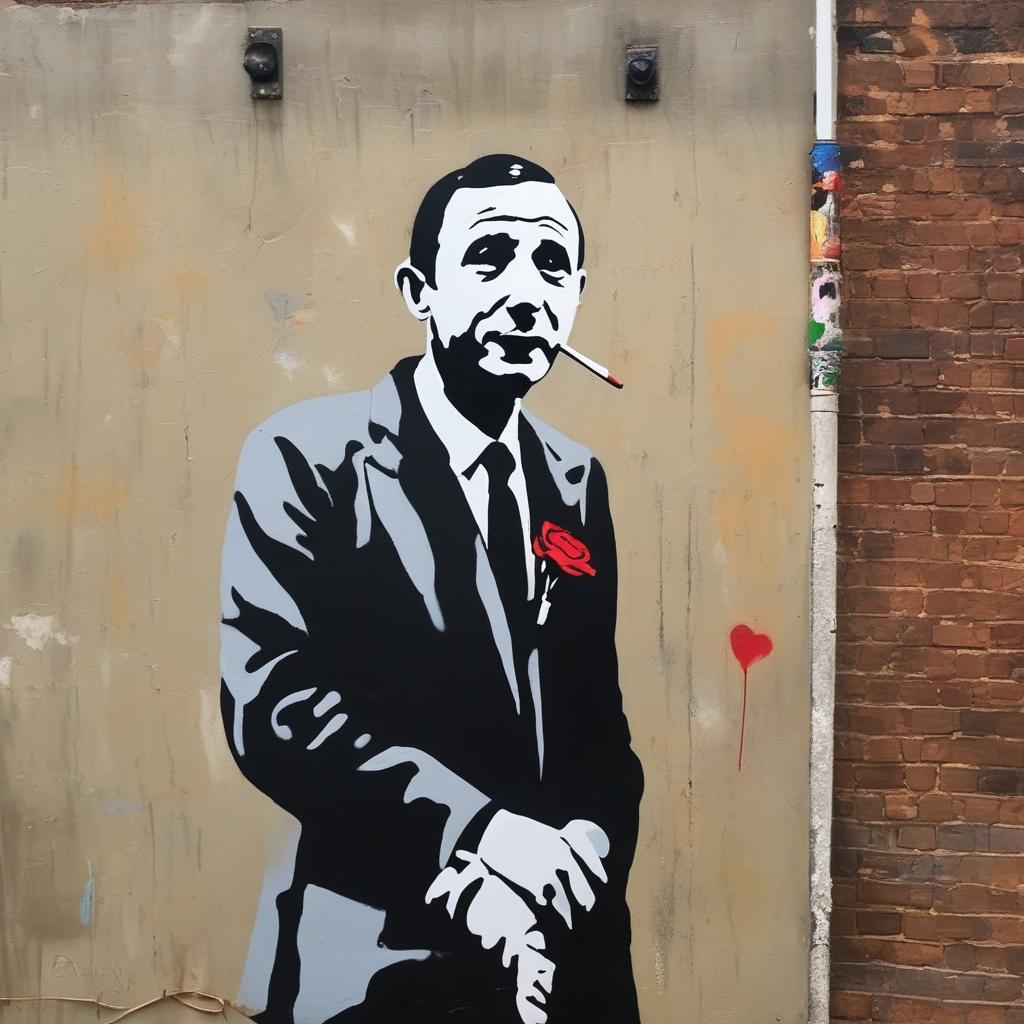} & 
        \includegraphics[width=0.139\textwidth]{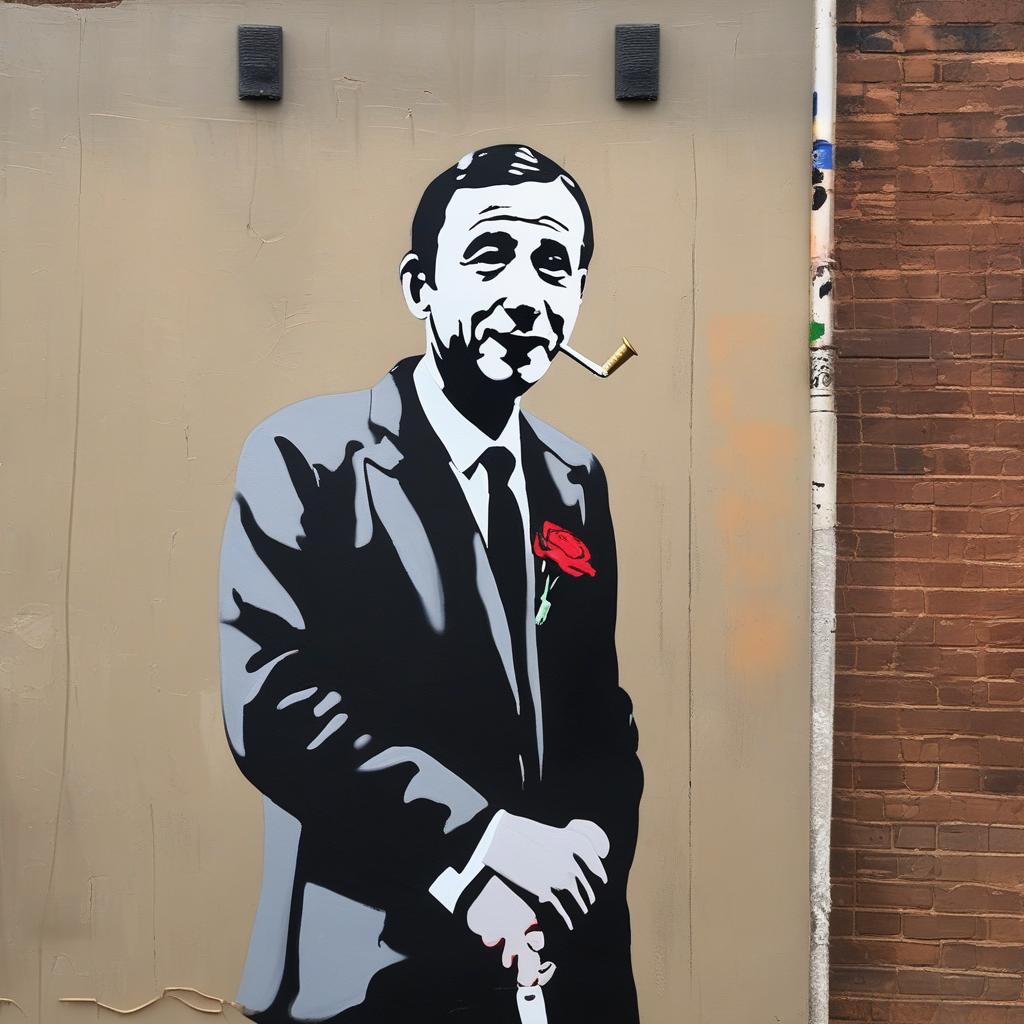} & 
        \includegraphics[width=0.139\textwidth]{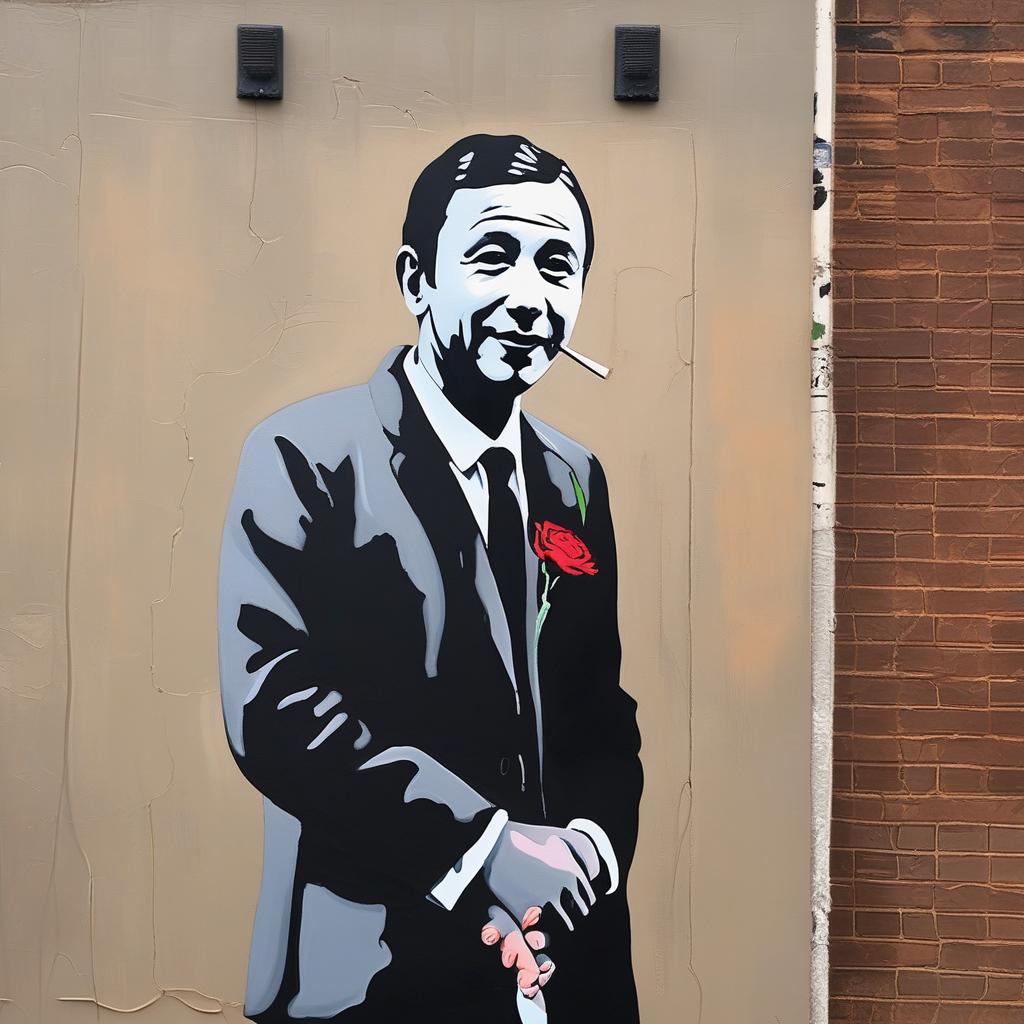} & 
        \includegraphics[width=0.139\textwidth]{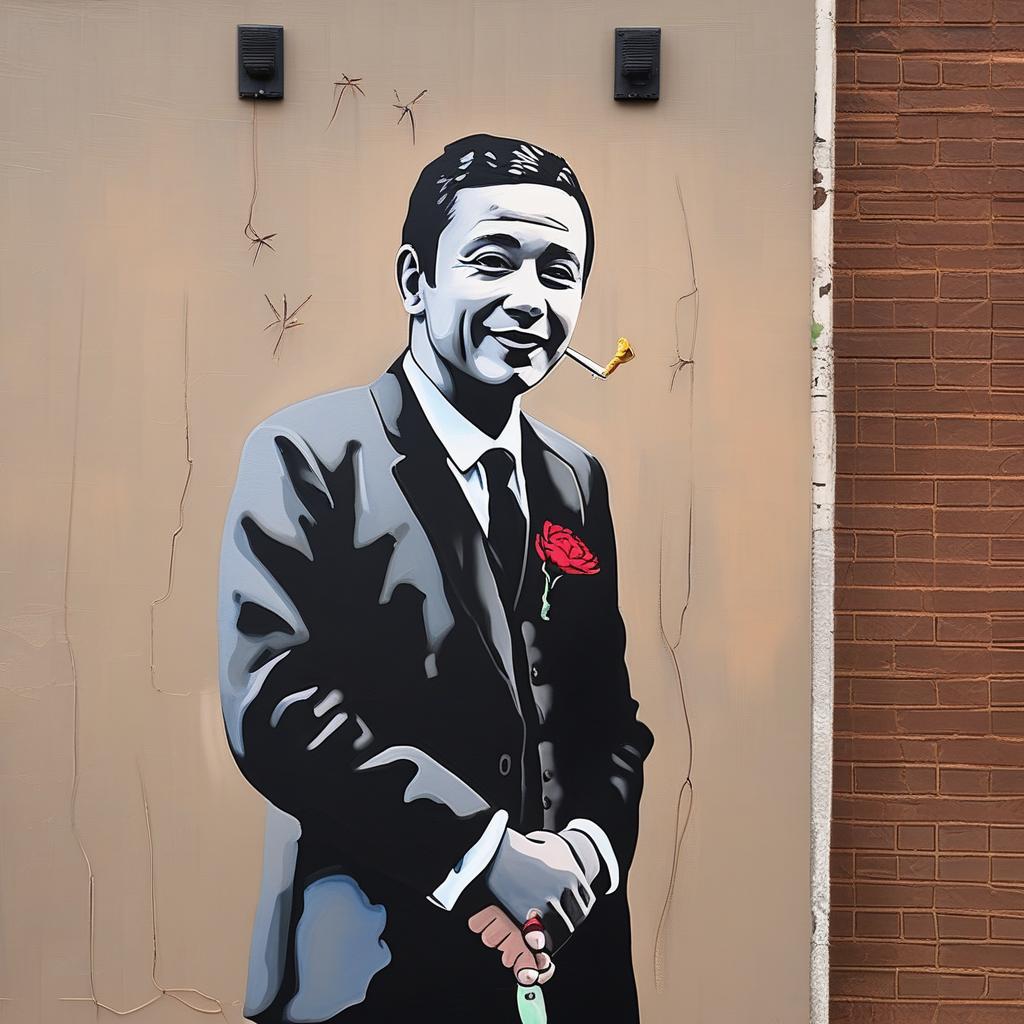} & 
        \includegraphics[width=0.139\textwidth]{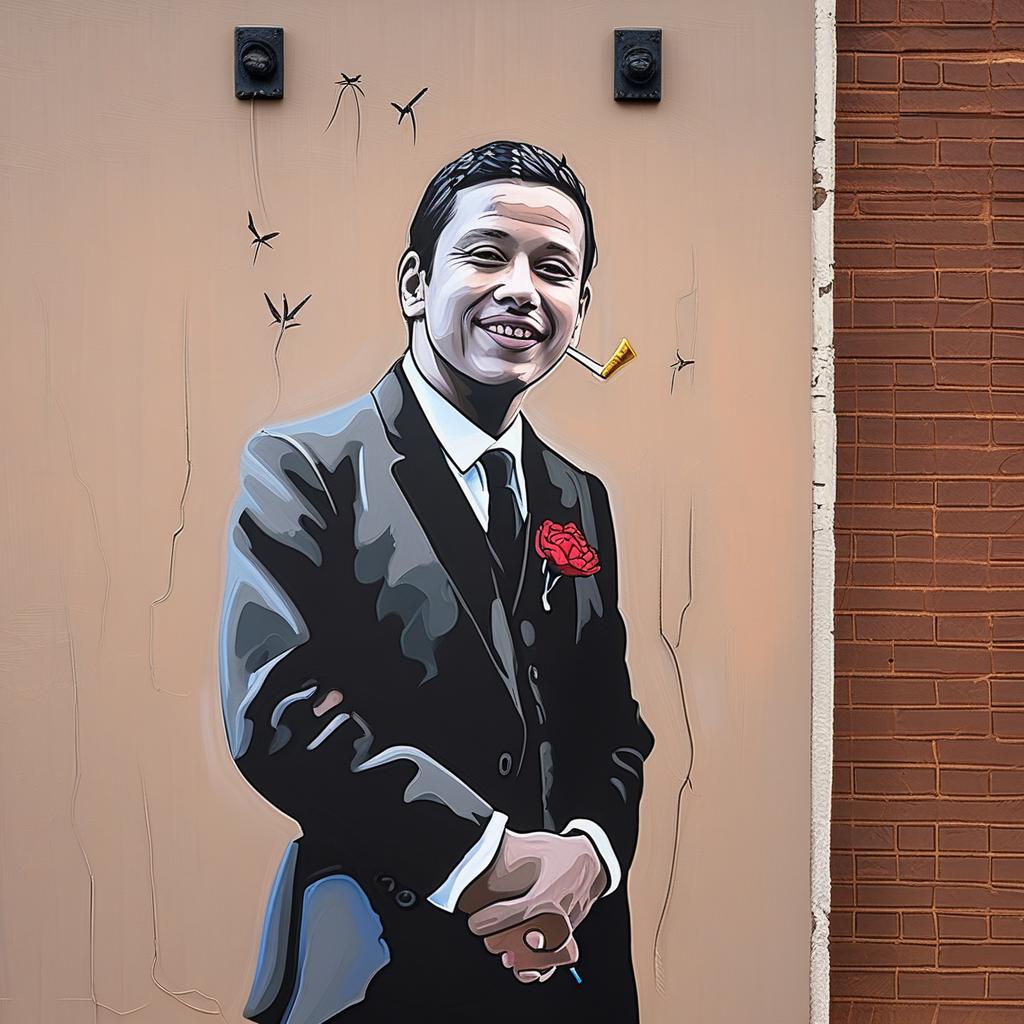} & 
        \includegraphics[width=0.139\textwidth]{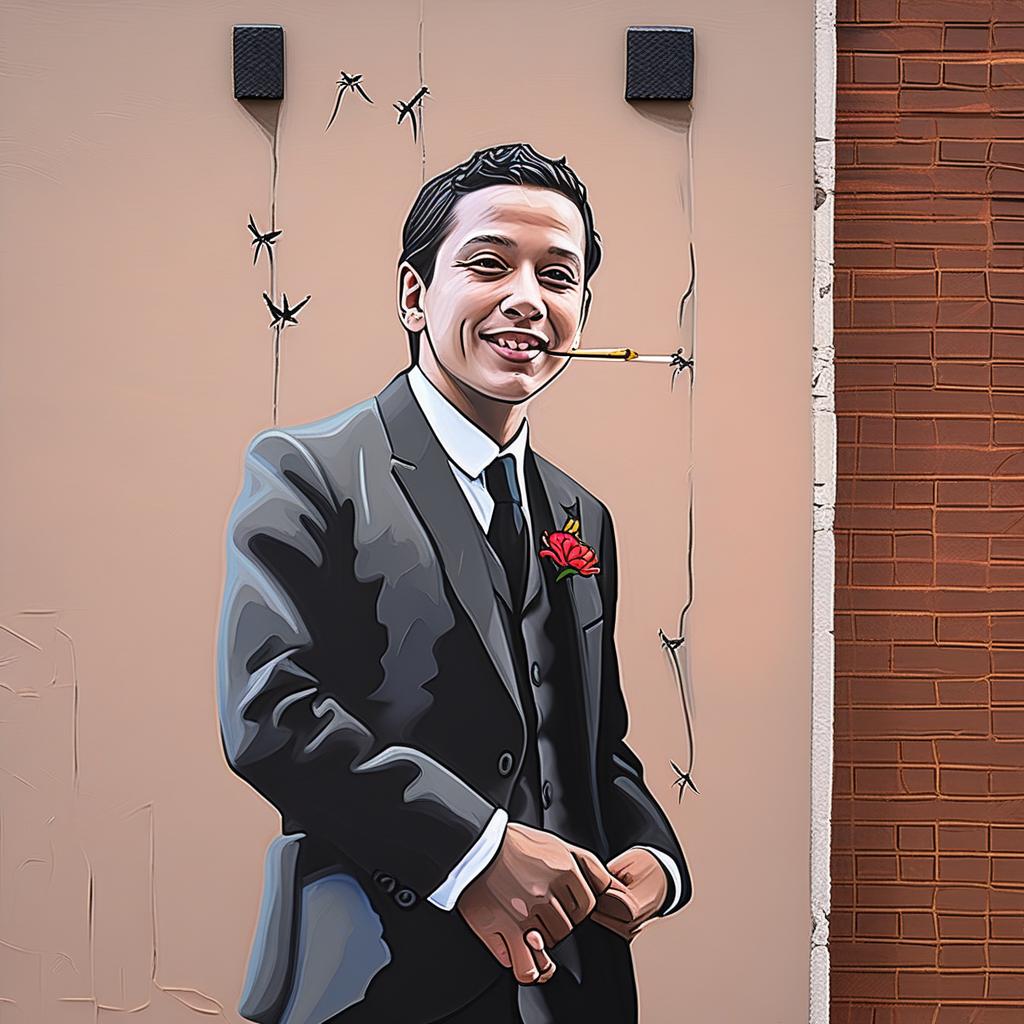} \\

        \raisebox{0.06\textwidth}{$\beta=0.5$} &
        \includegraphics[width=0.139\textwidth]{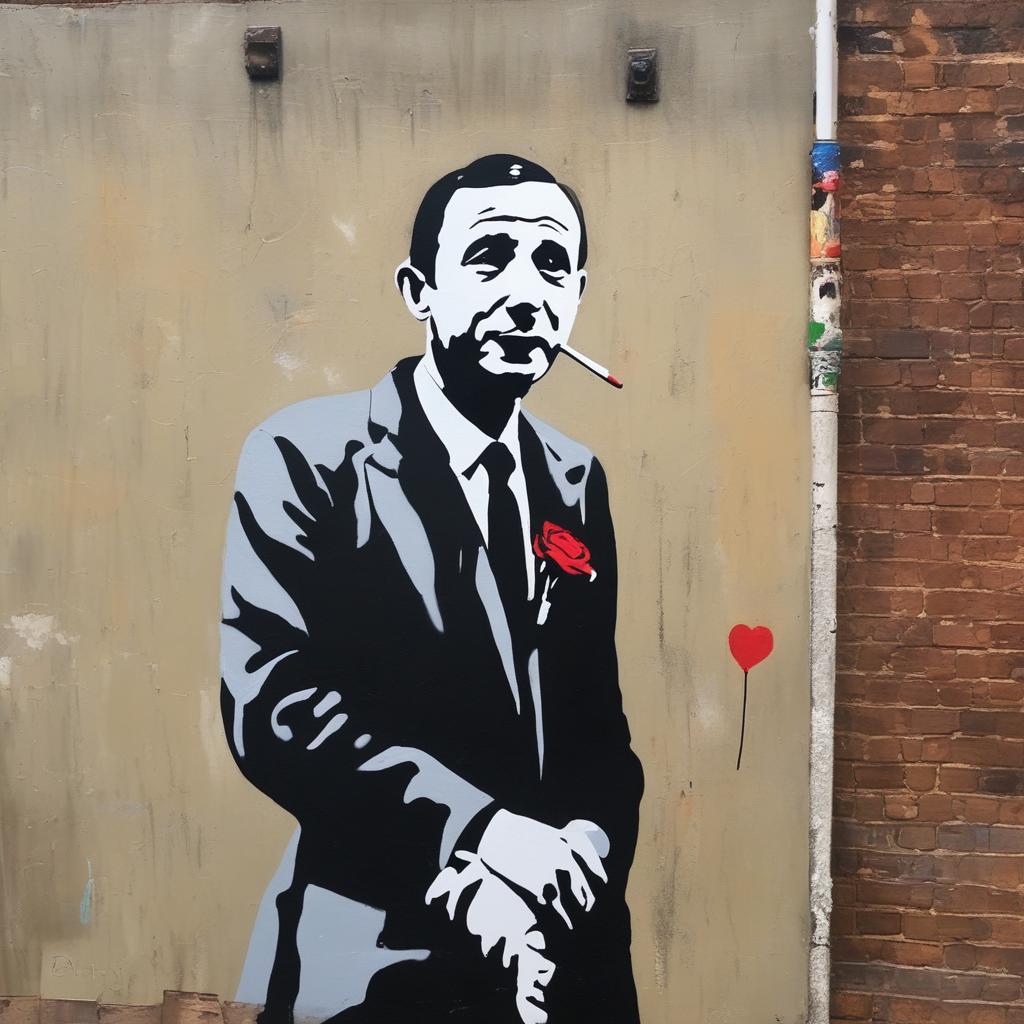} & 
        \includegraphics[width=0.139\textwidth]{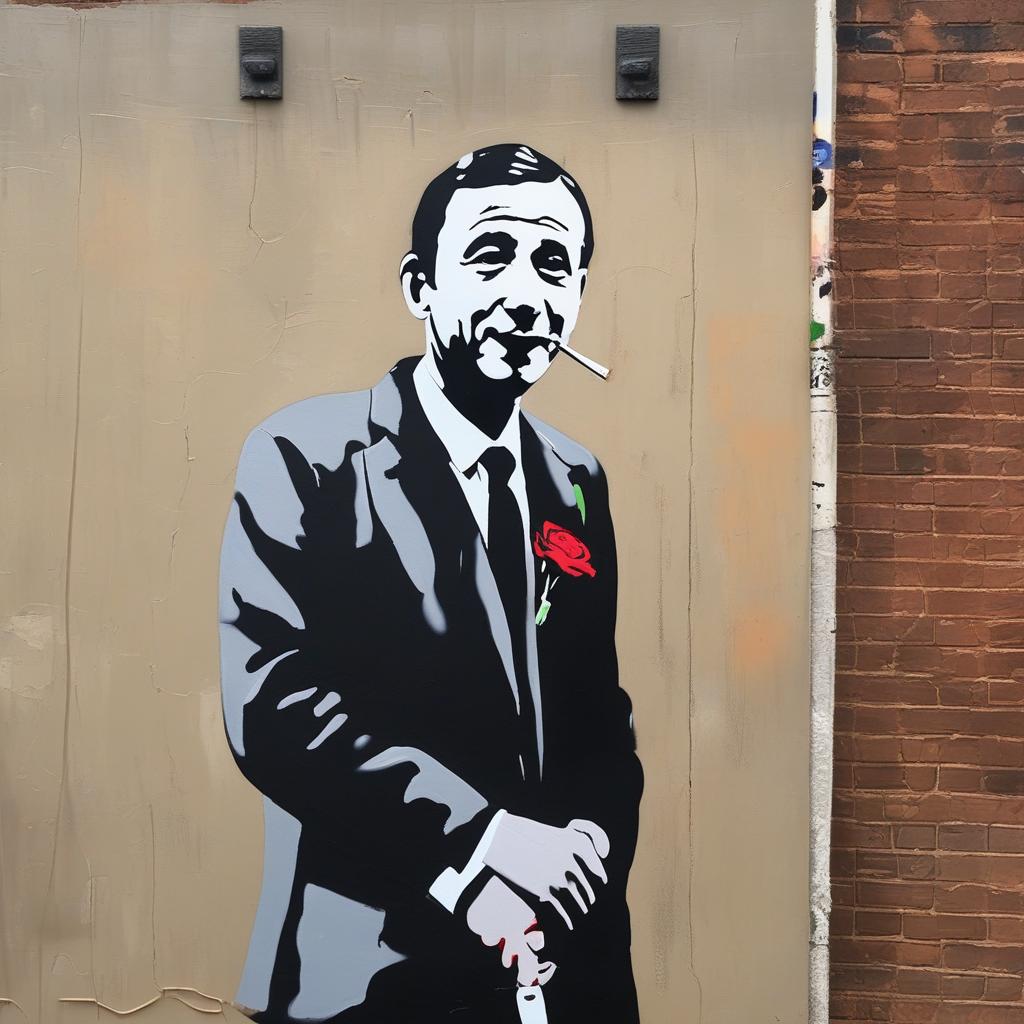} & 
        \includegraphics[width=0.139\textwidth]{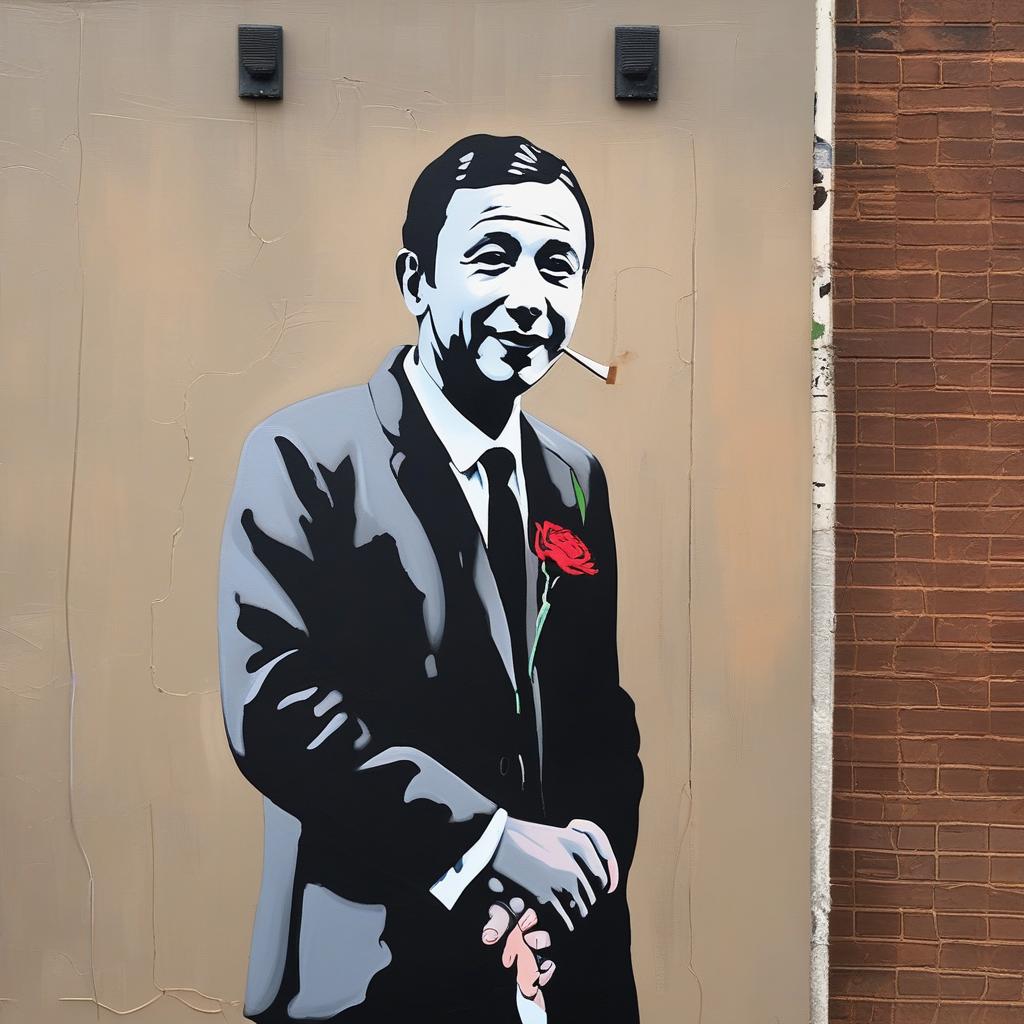} & 
        \includegraphics[width=0.139\textwidth]{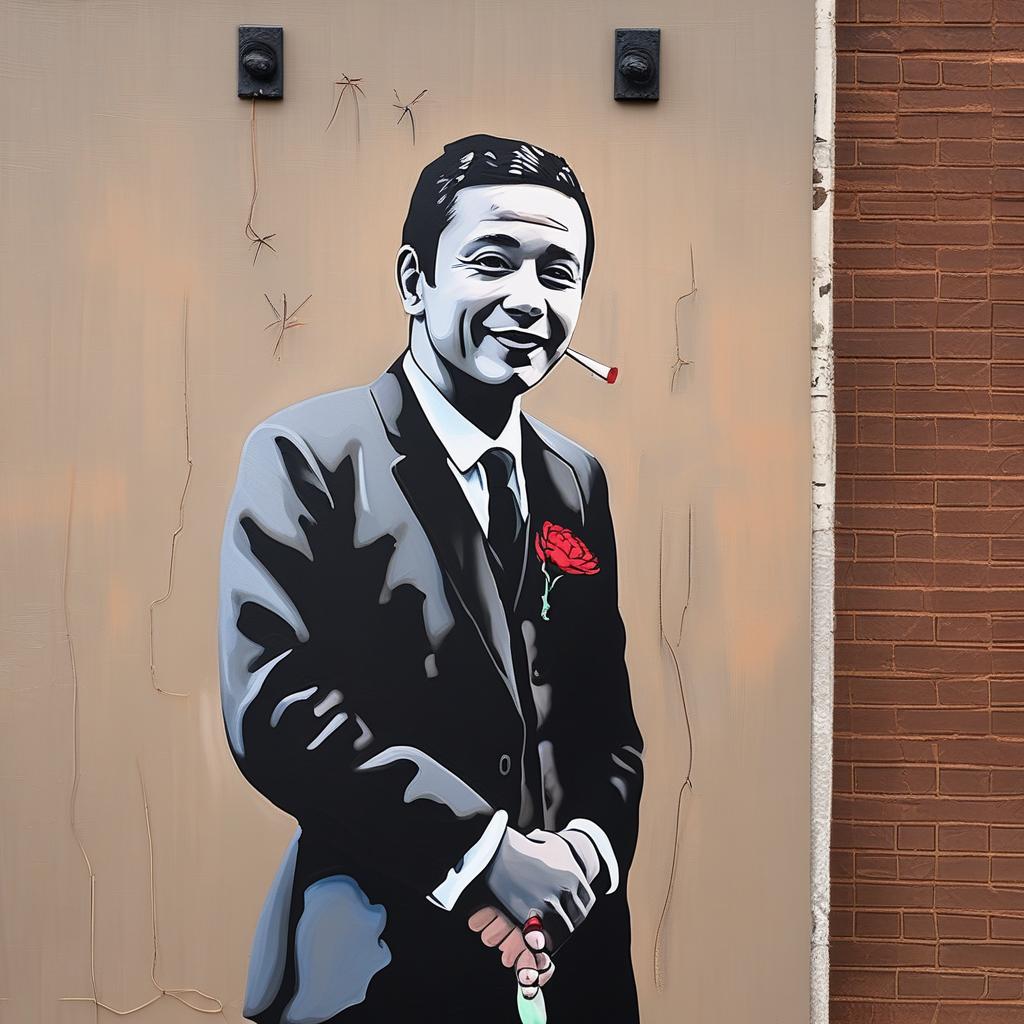} & 
        \includegraphics[width=0.139\textwidth]{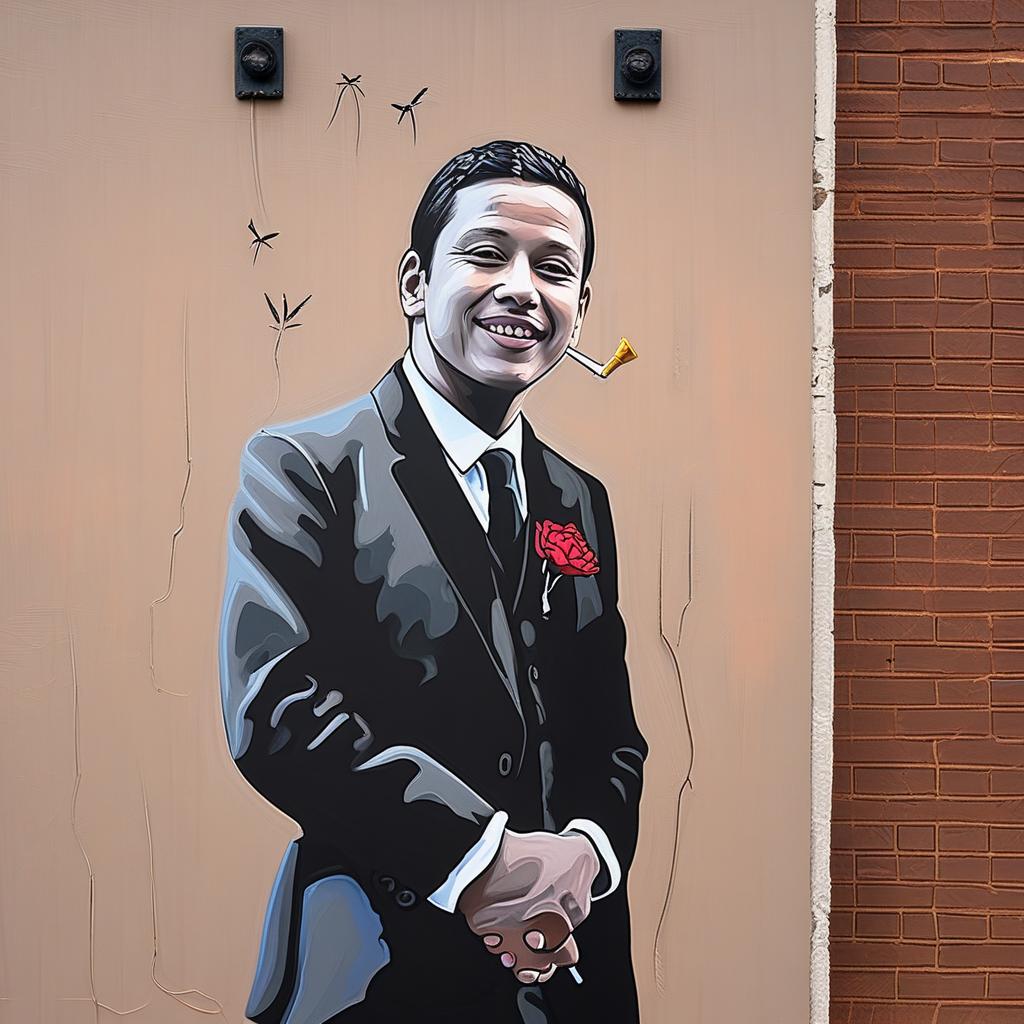} & 
        \includegraphics[width=0.139\textwidth]{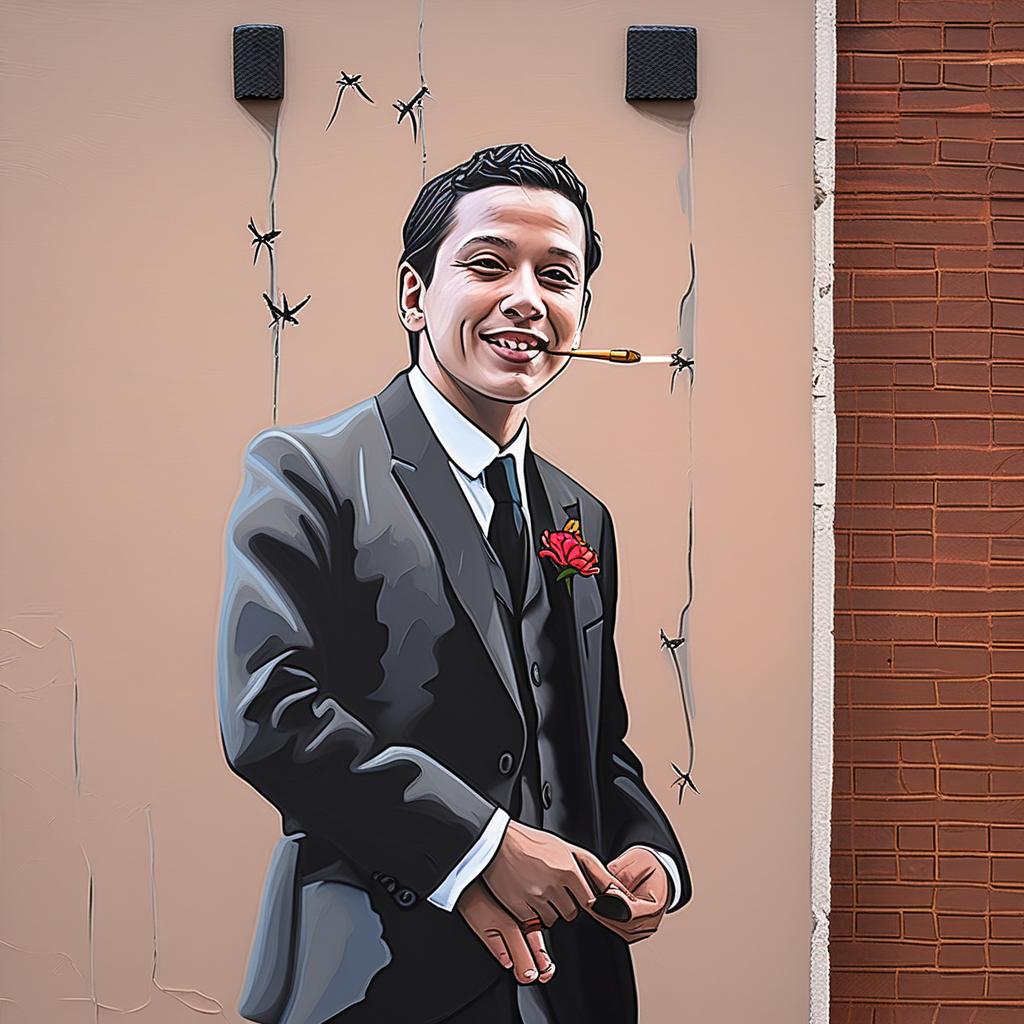} \\

        \raisebox{0.06\textwidth}{$\beta=1$} &
        \includegraphics[width=0.139\textwidth]{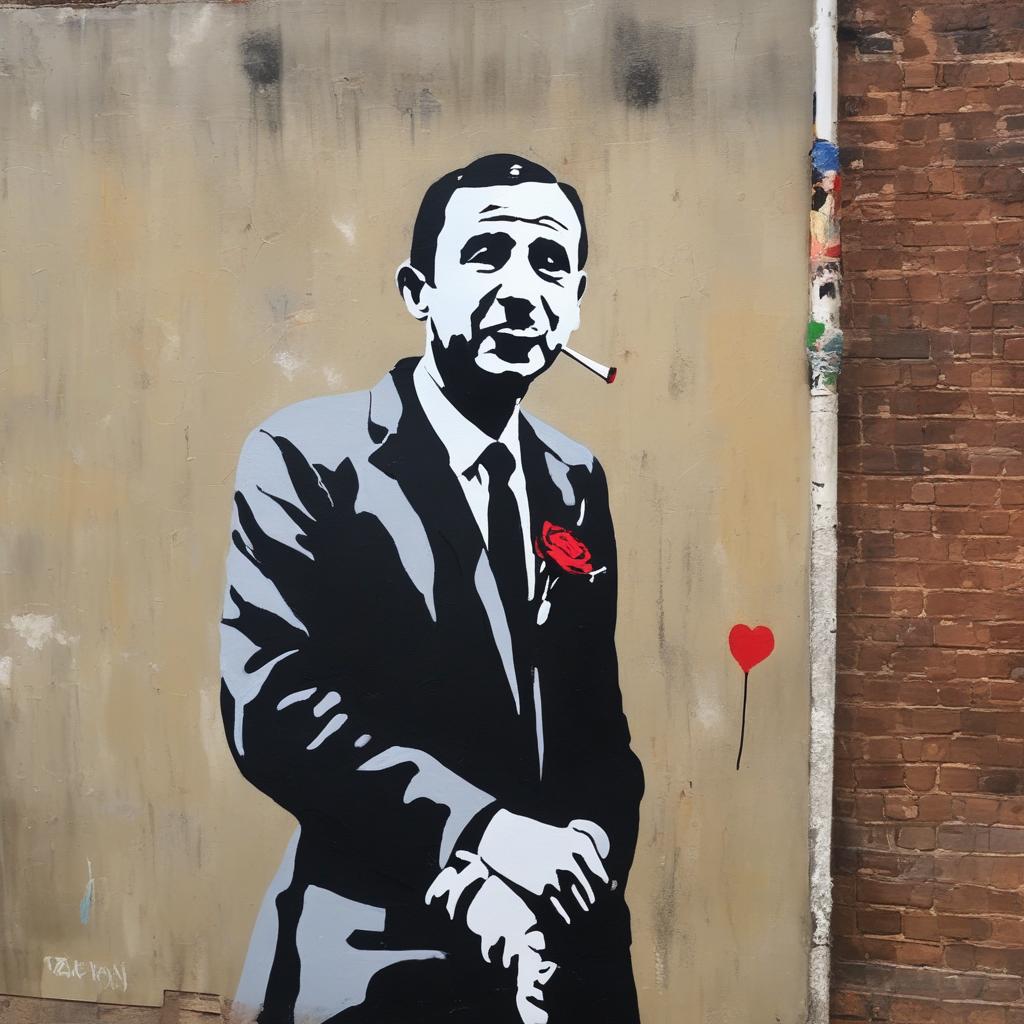} & 
        \includegraphics[width=0.139\textwidth]{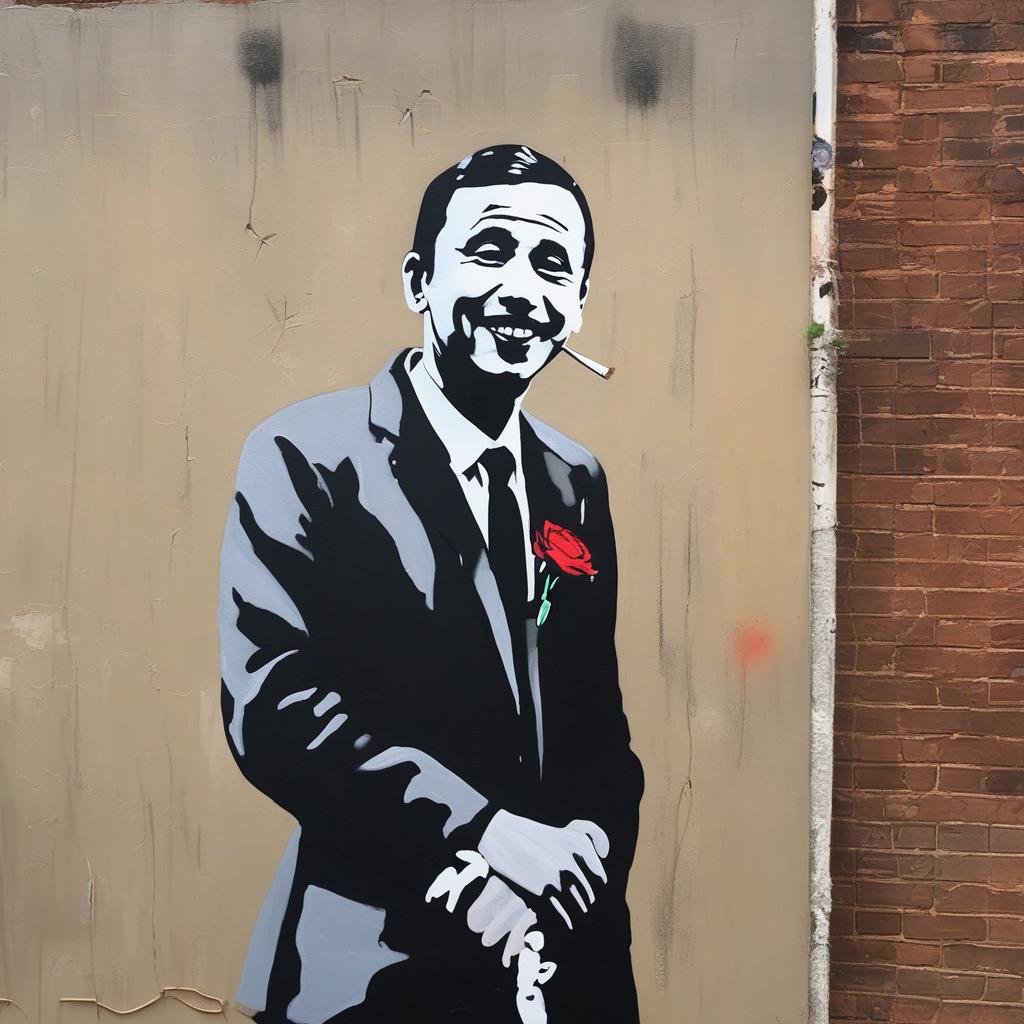} & 
        \includegraphics[width=0.139\textwidth]{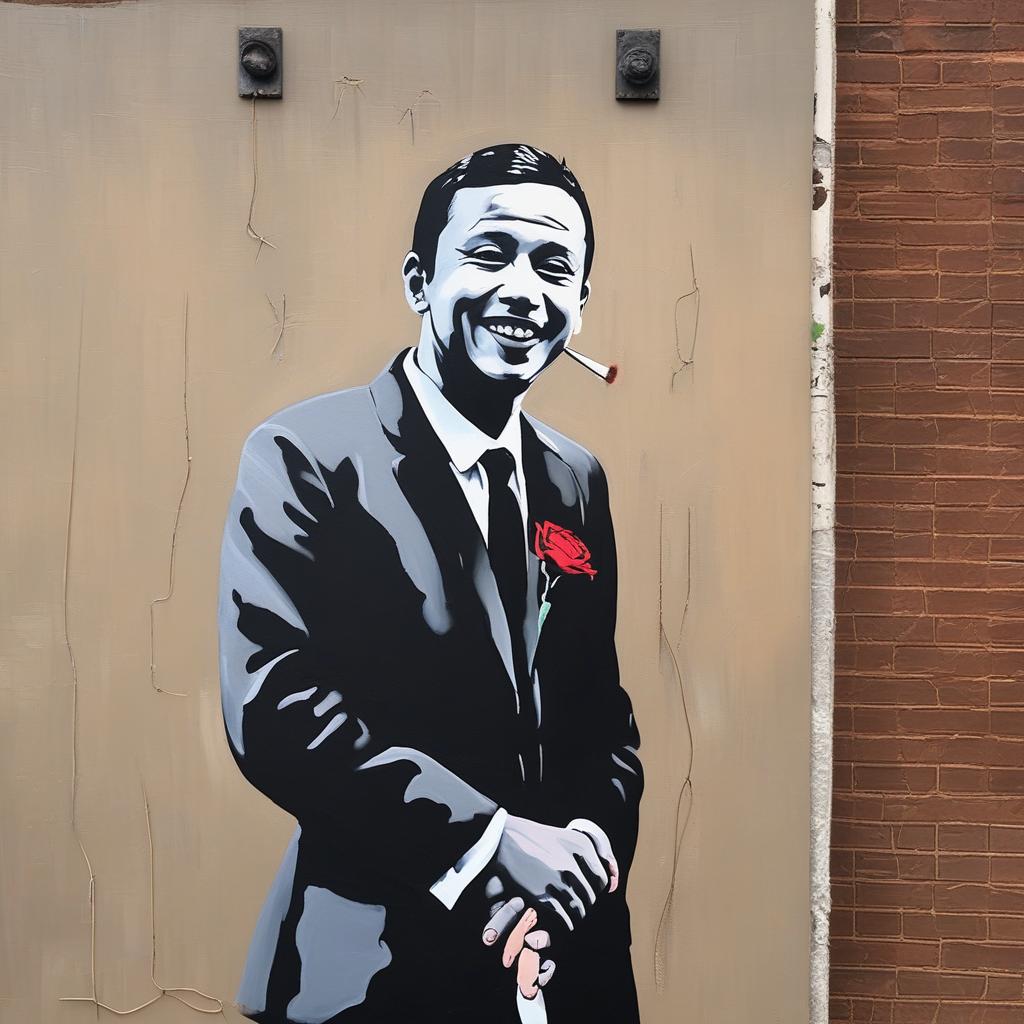} & 
        \includegraphics[width=0.139\textwidth]{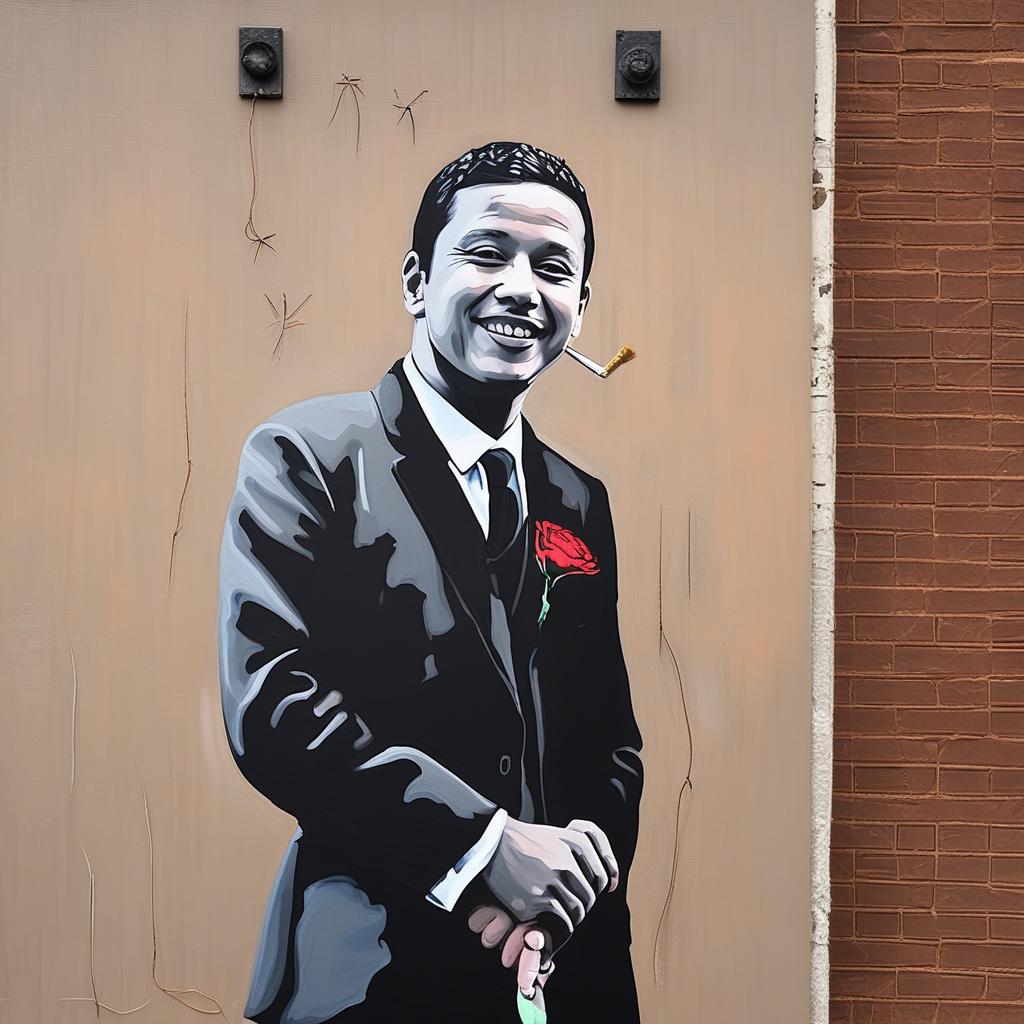} & 
        \includegraphics[width=0.139\textwidth]{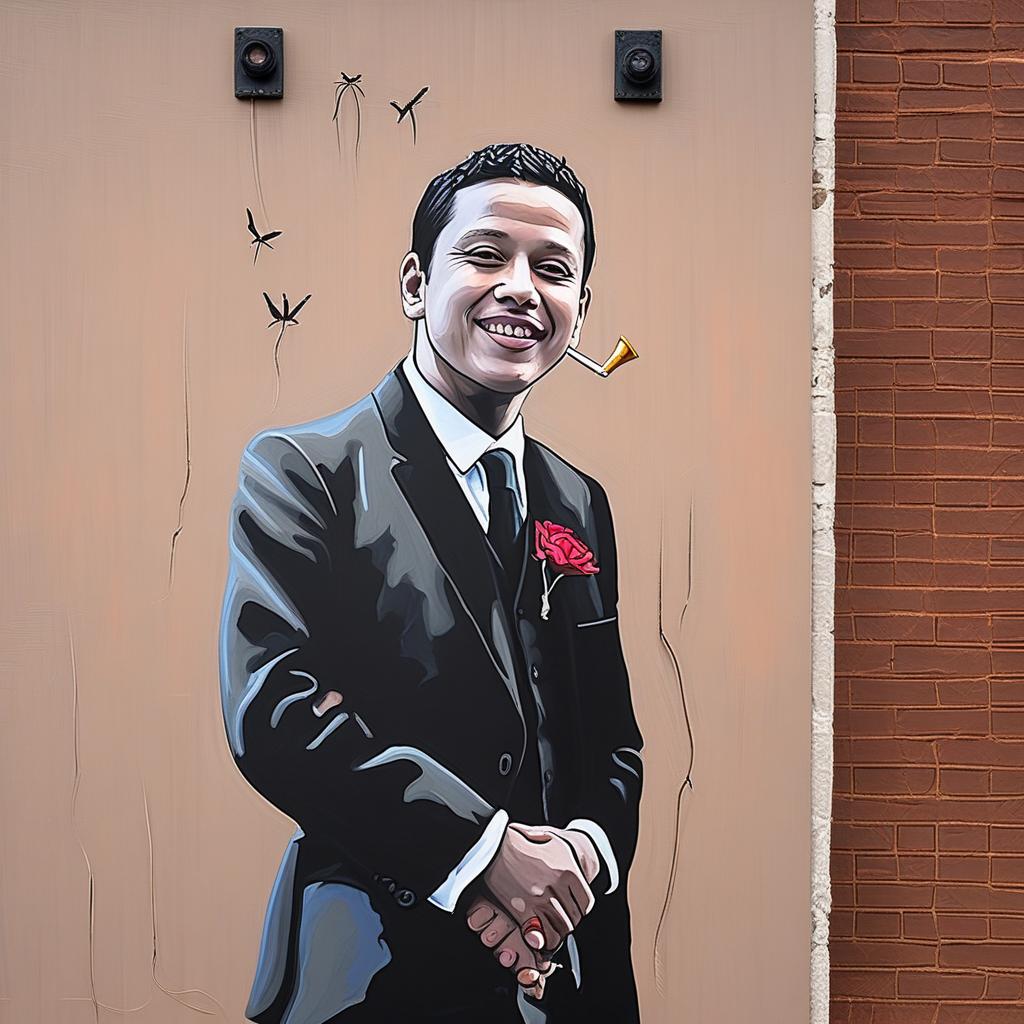} & 
        \includegraphics[width=0.139\textwidth]{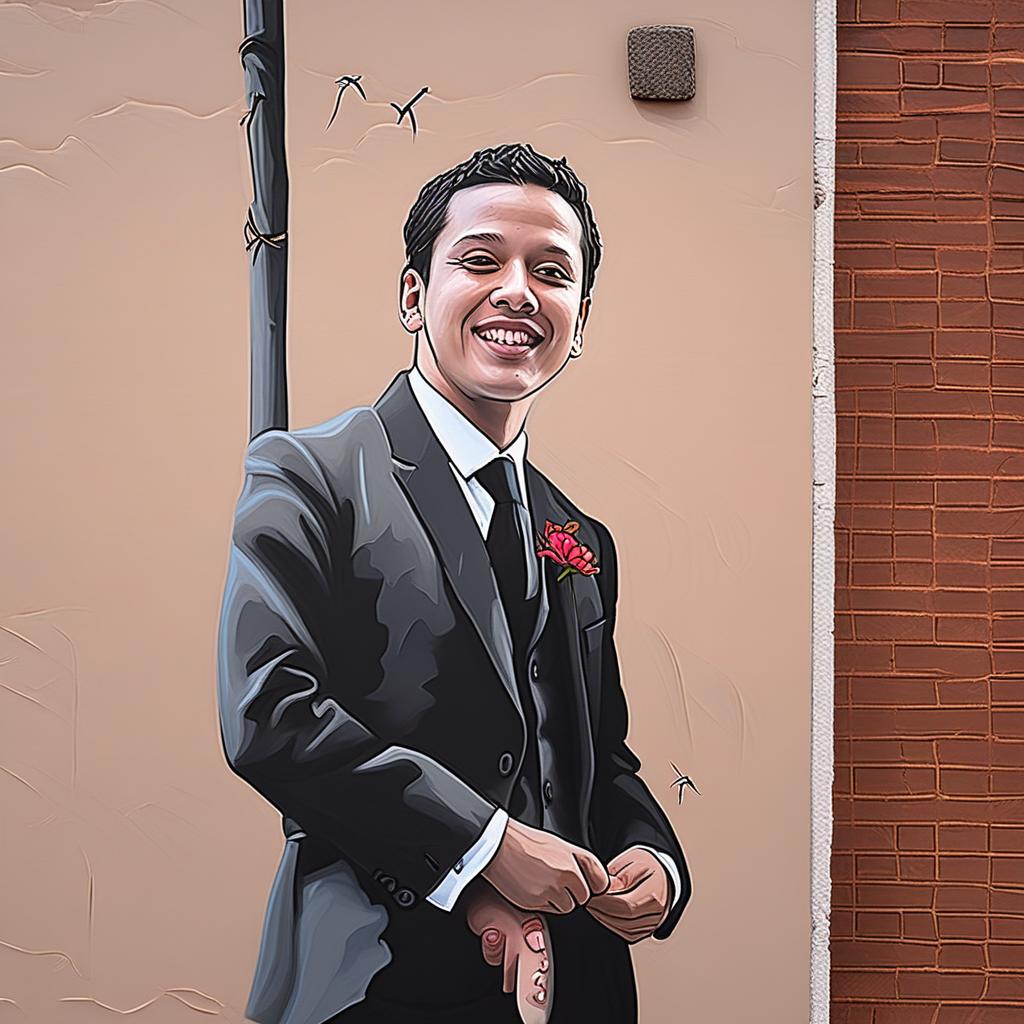} \\

        \raisebox{0.06\textwidth}{$\beta=1.5$} &
        \includegraphics[width=0.139\textwidth]{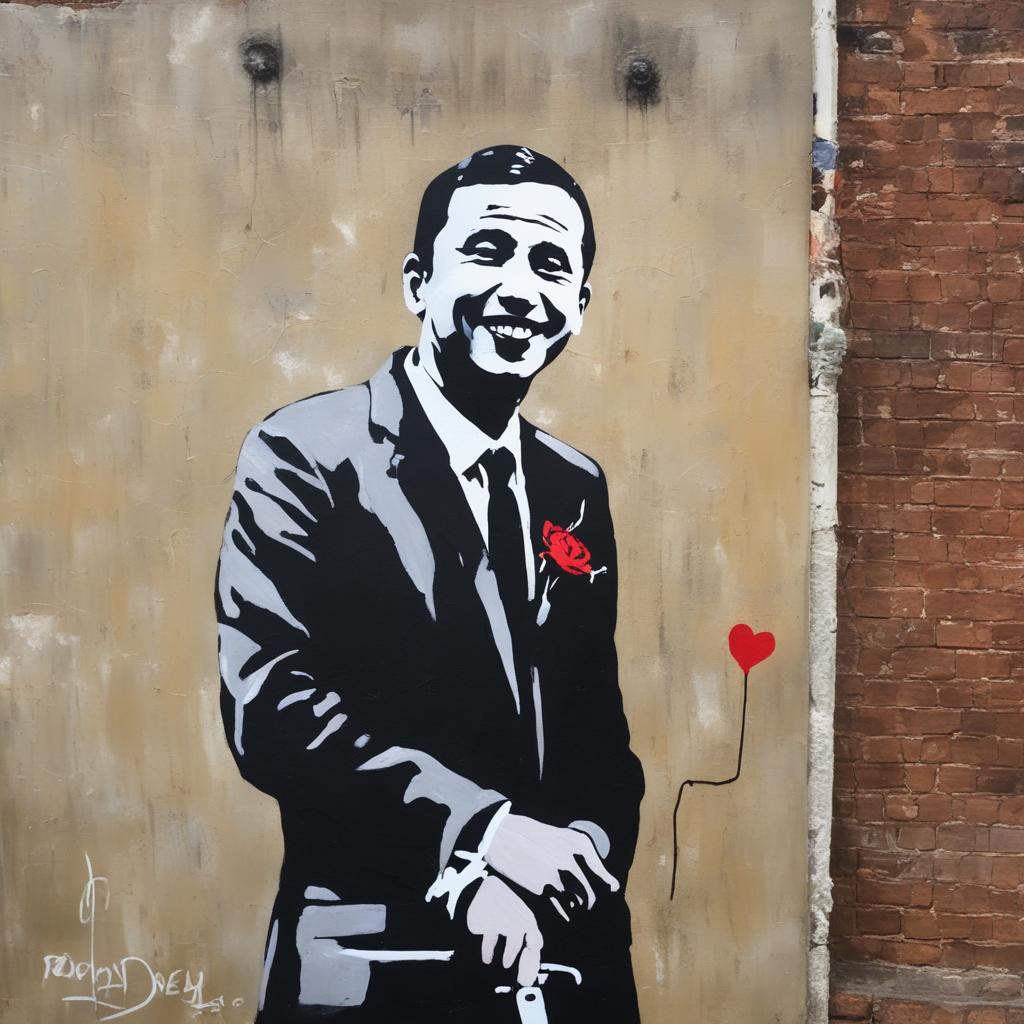} & 
        \includegraphics[width=0.139\textwidth]{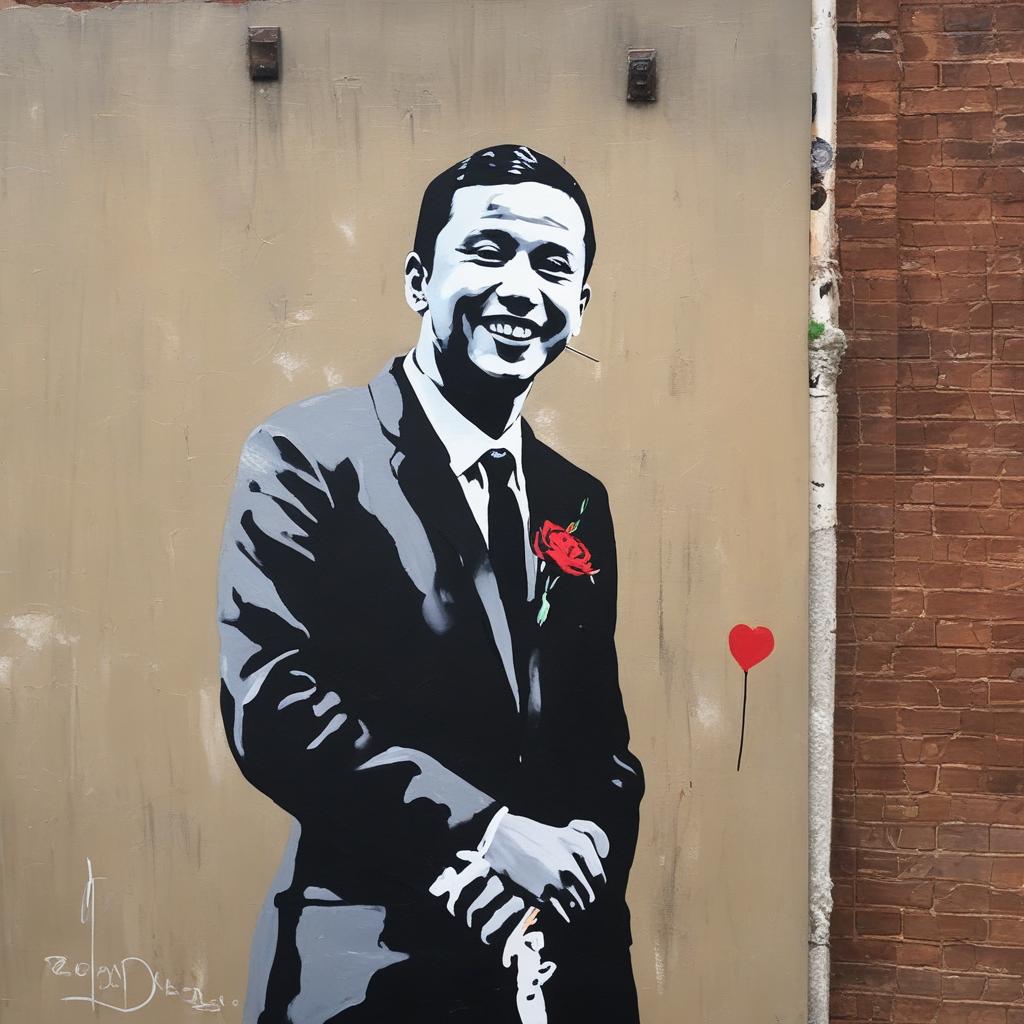} & 
        \includegraphics[width=0.139\textwidth]{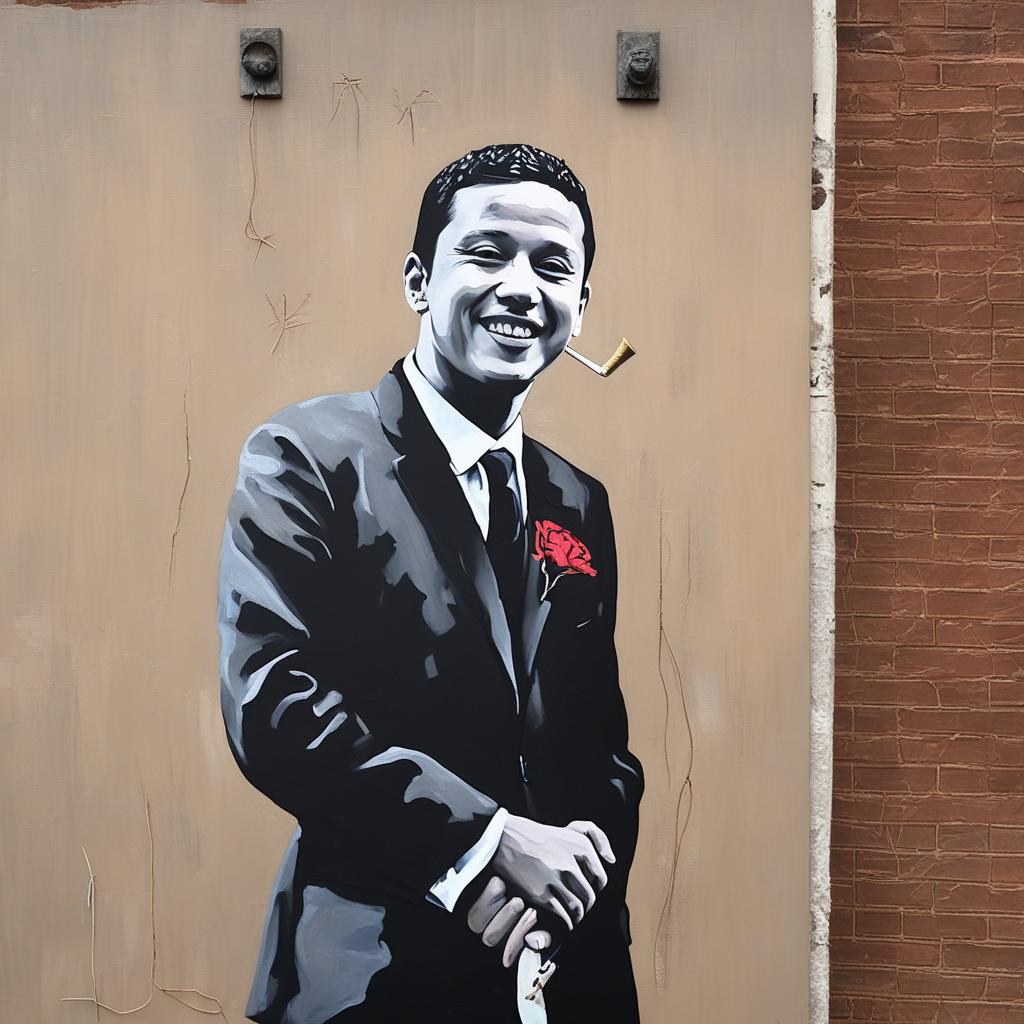} & 
        \includegraphics[width=0.139\textwidth]{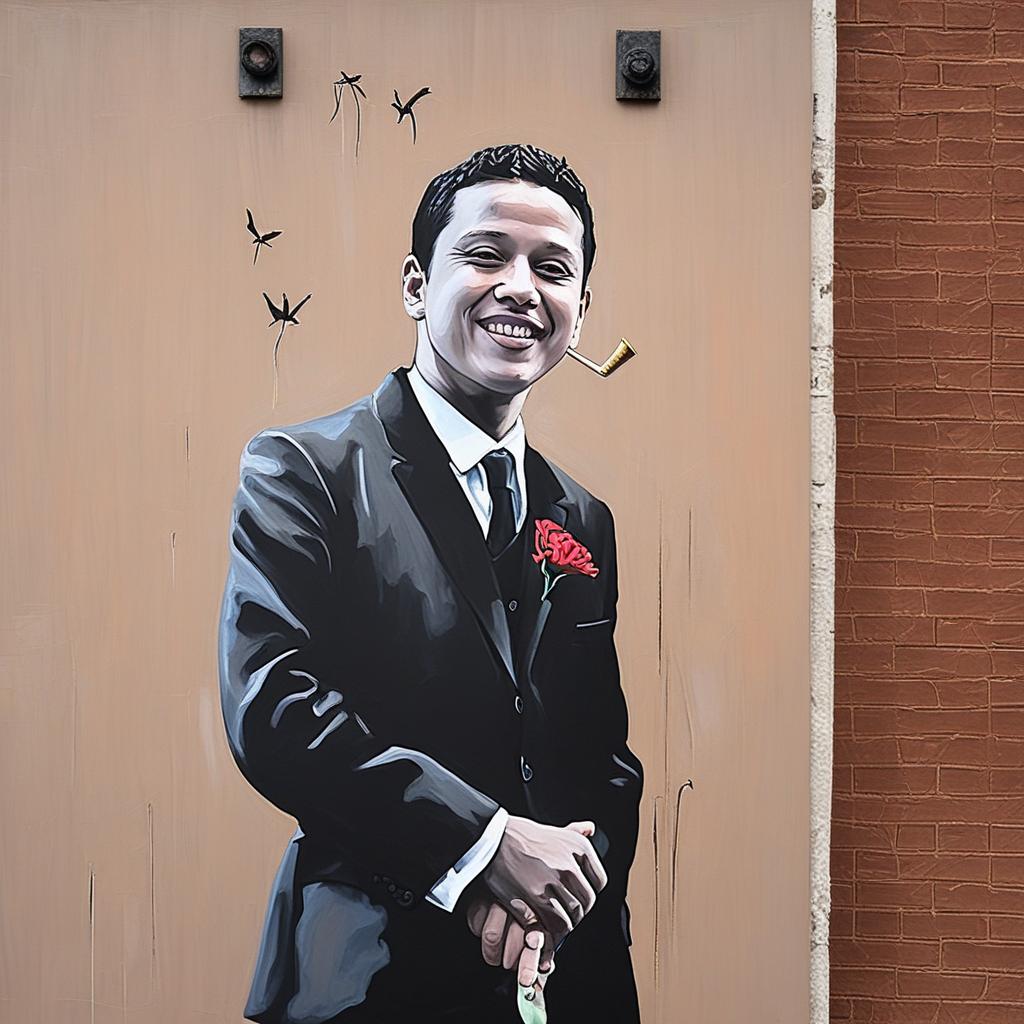} & 
        \includegraphics[width=0.139\textwidth]{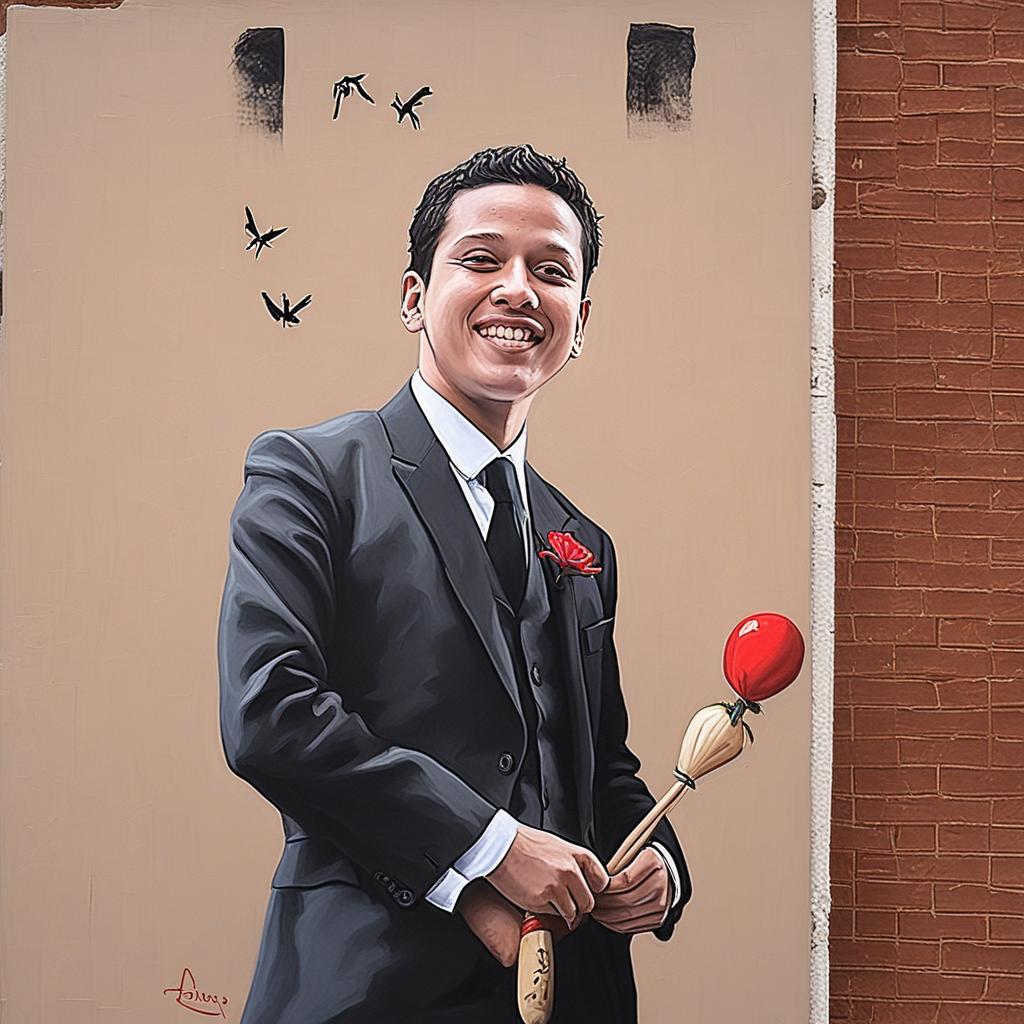} & 
        \includegraphics[width=0.139\textwidth]{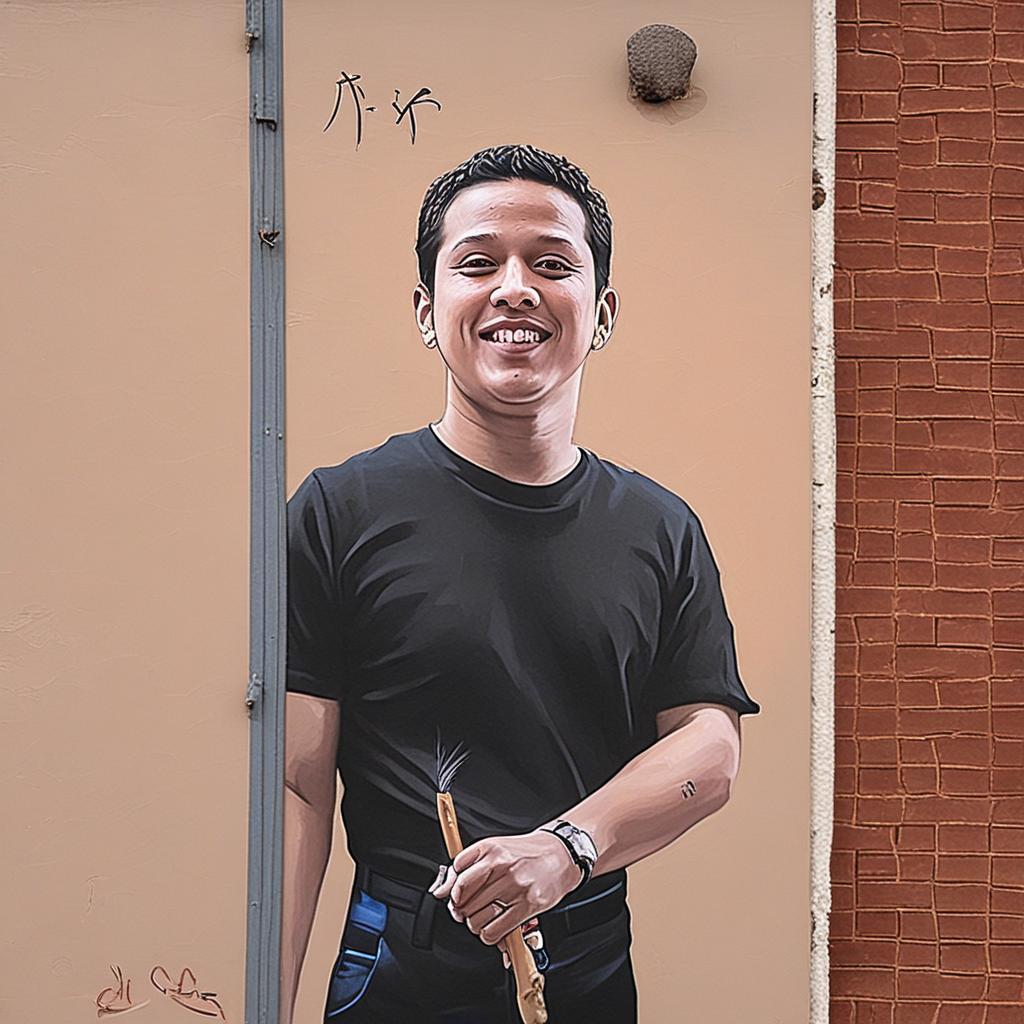} \\

        \raisebox{0.06\textwidth}{$\beta=2$} &
        \includegraphics[width=0.139\textwidth]{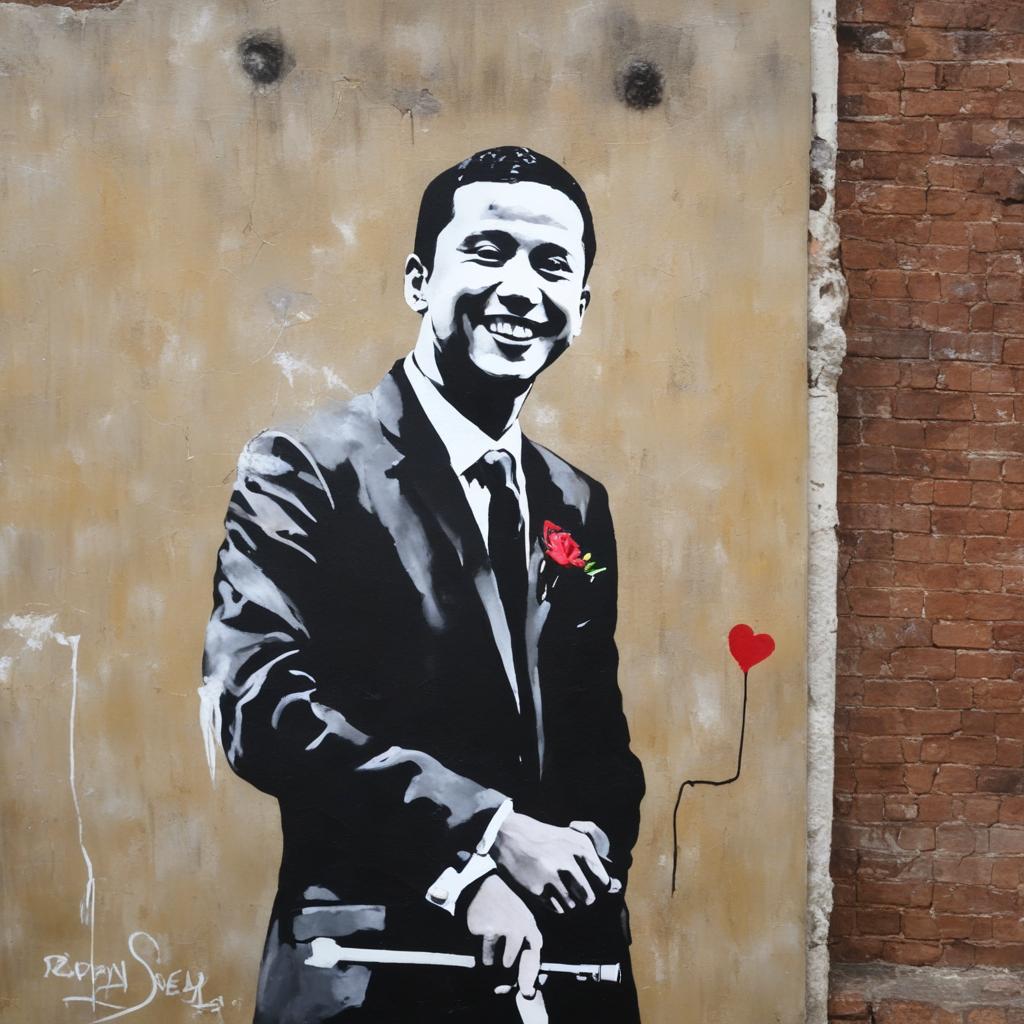} & 
        \includegraphics[width=0.139\textwidth]{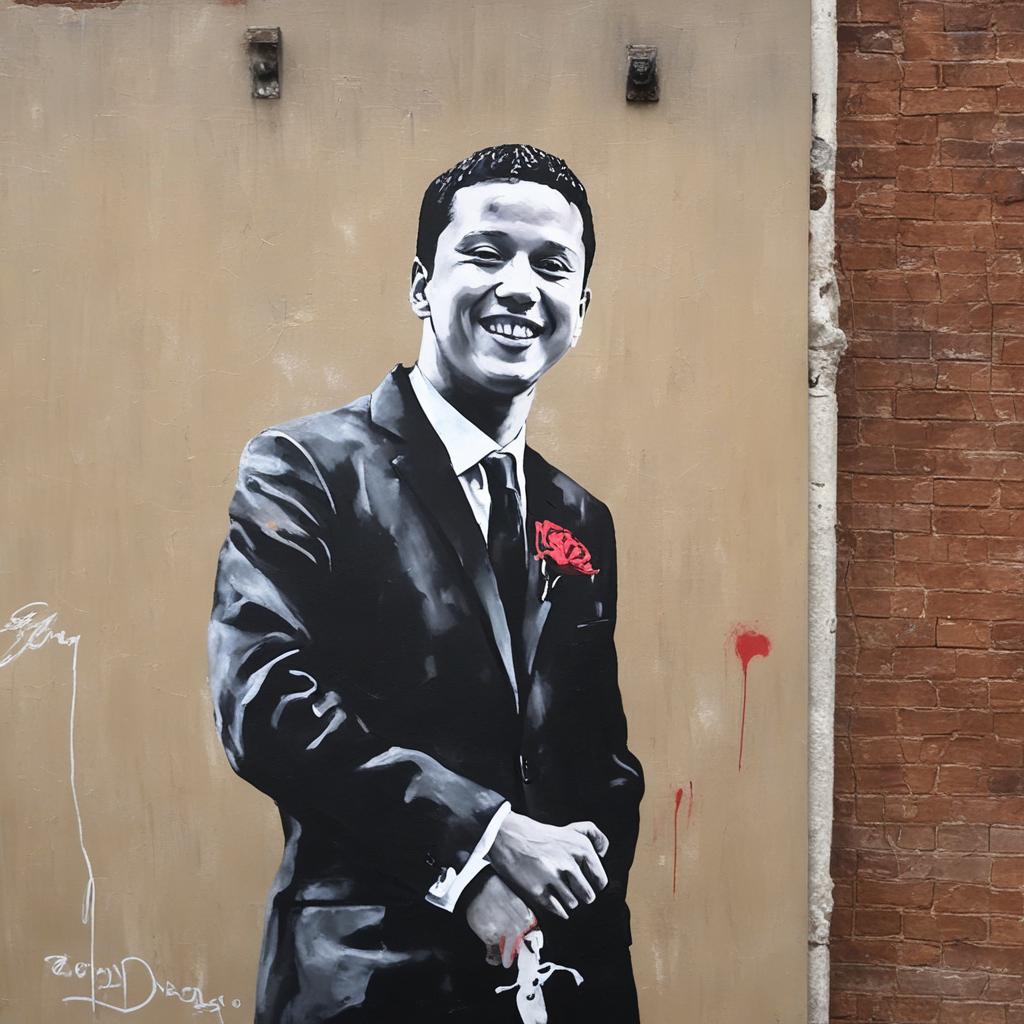} & 
        \includegraphics[width=0.139\textwidth]{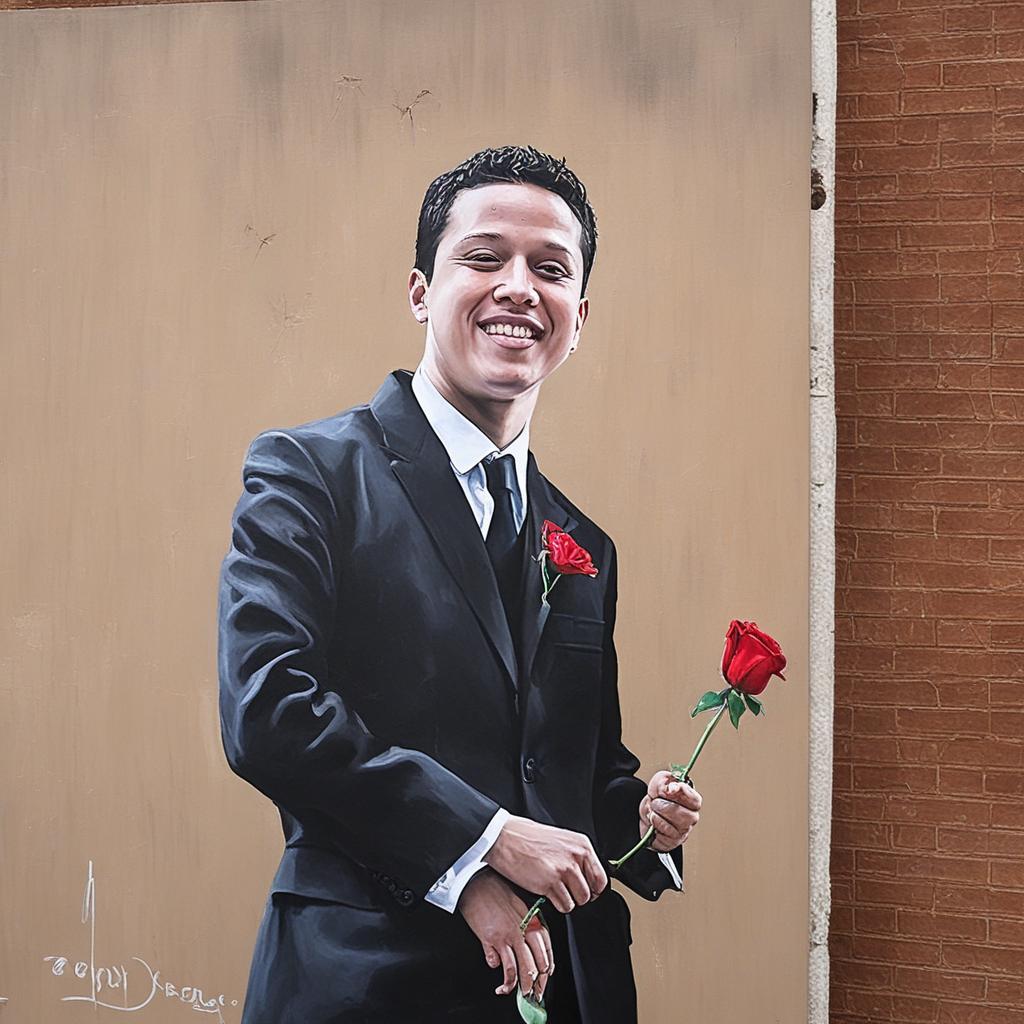} & 
        \includegraphics[width=0.139\textwidth]{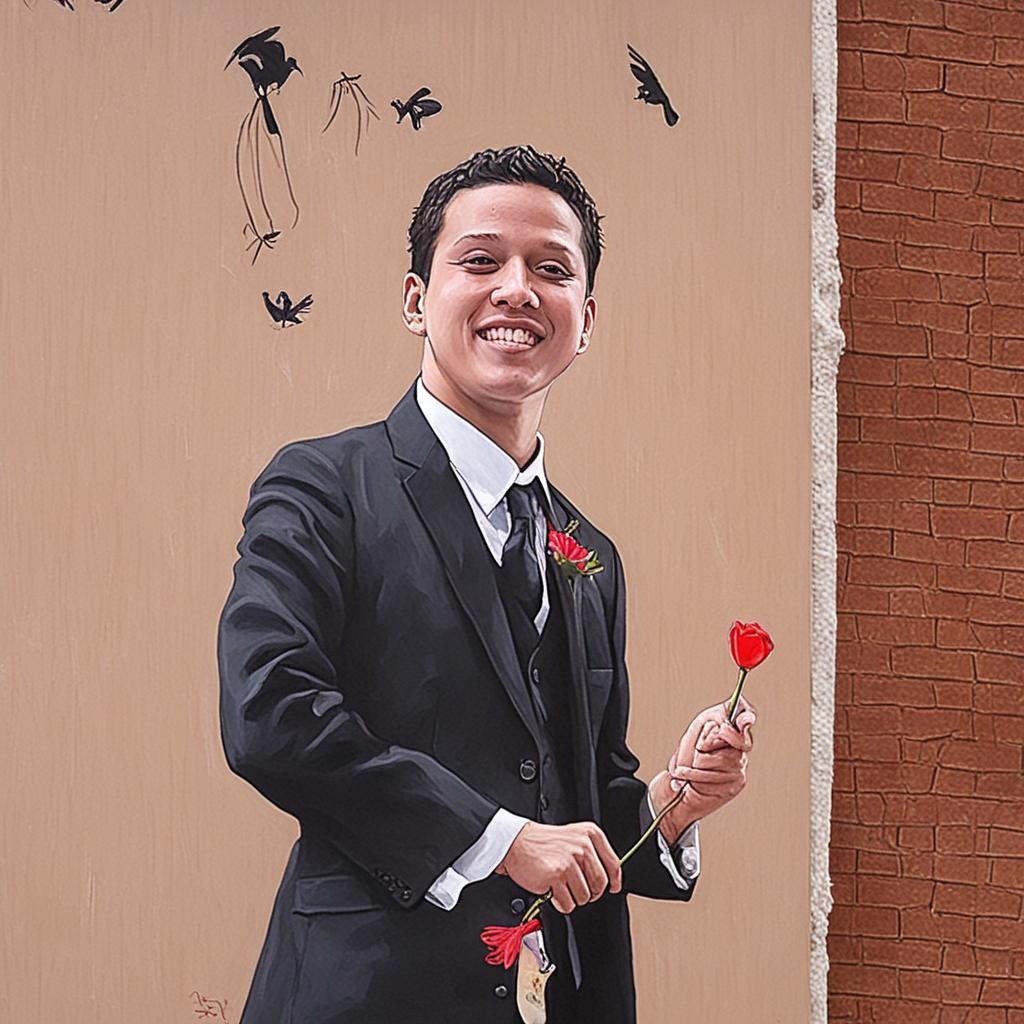} & 
        \includegraphics[width=0.139\textwidth]{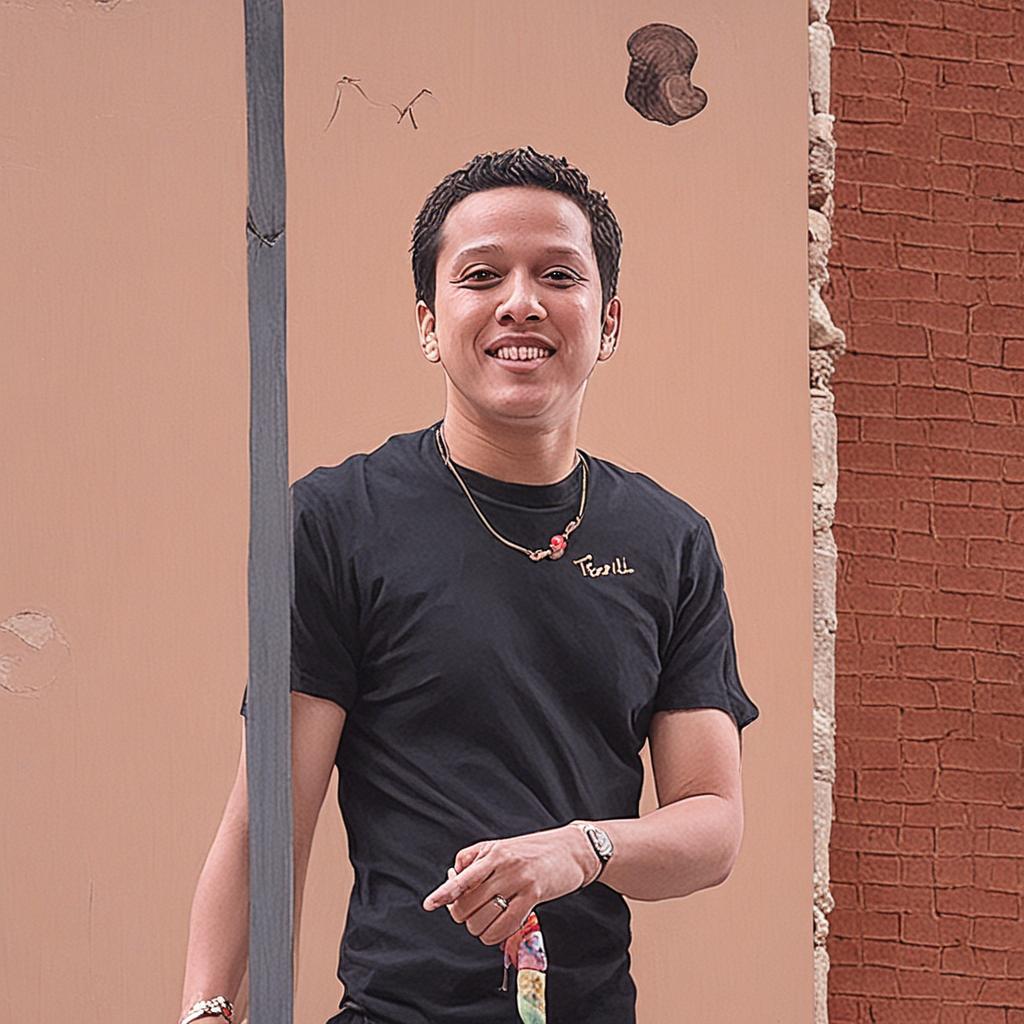} & 
        \includegraphics[width=0.139\textwidth]{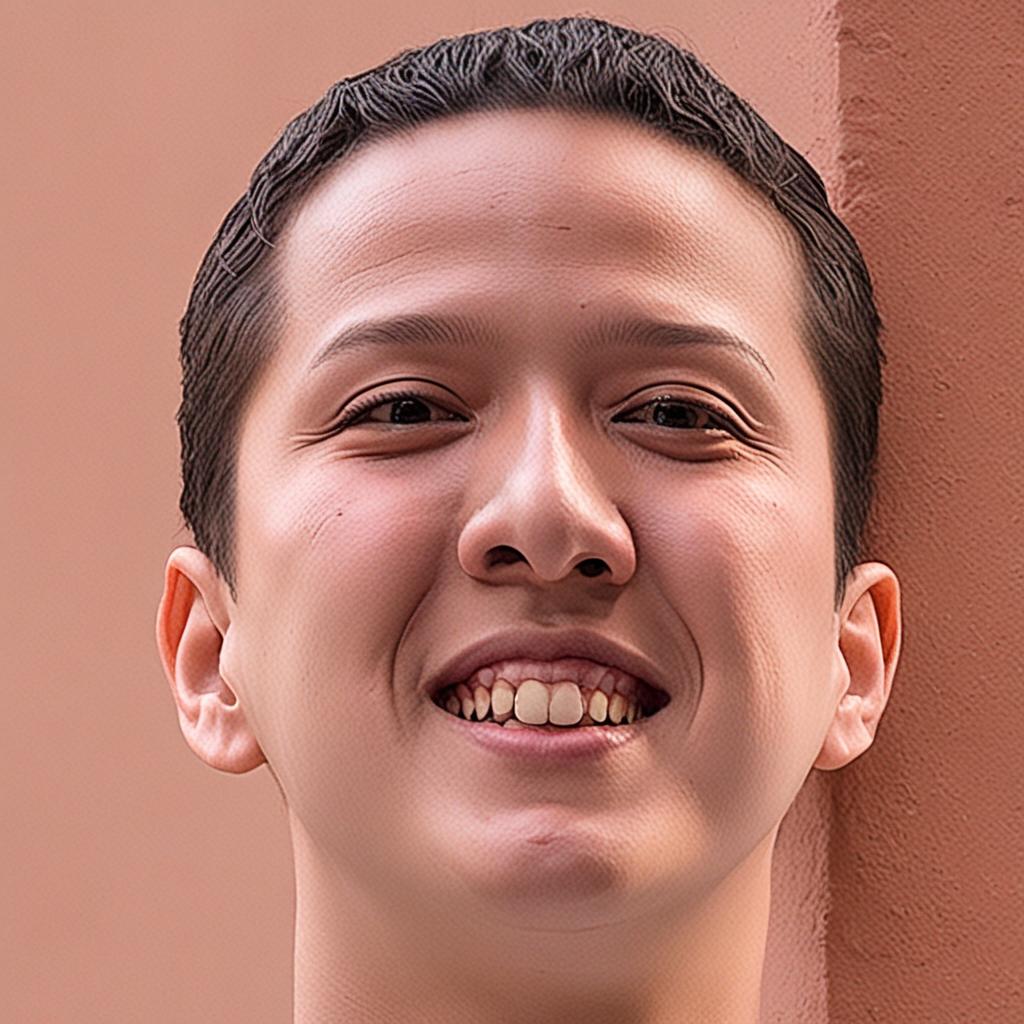} \\

        \raisebox{0.06\textwidth}{$\beta=2.5$} &
        \includegraphics[width=0.139\textwidth]{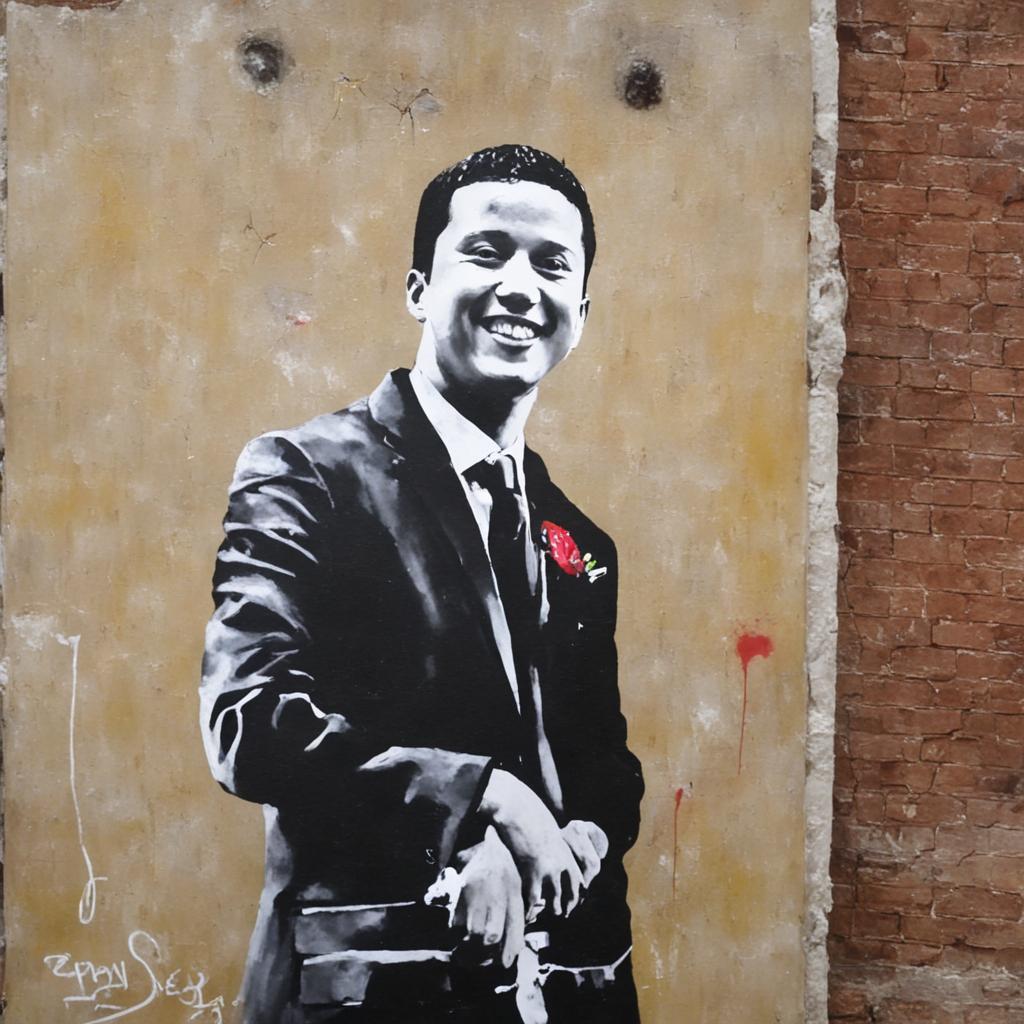} & 
        \includegraphics[width=0.139\textwidth]{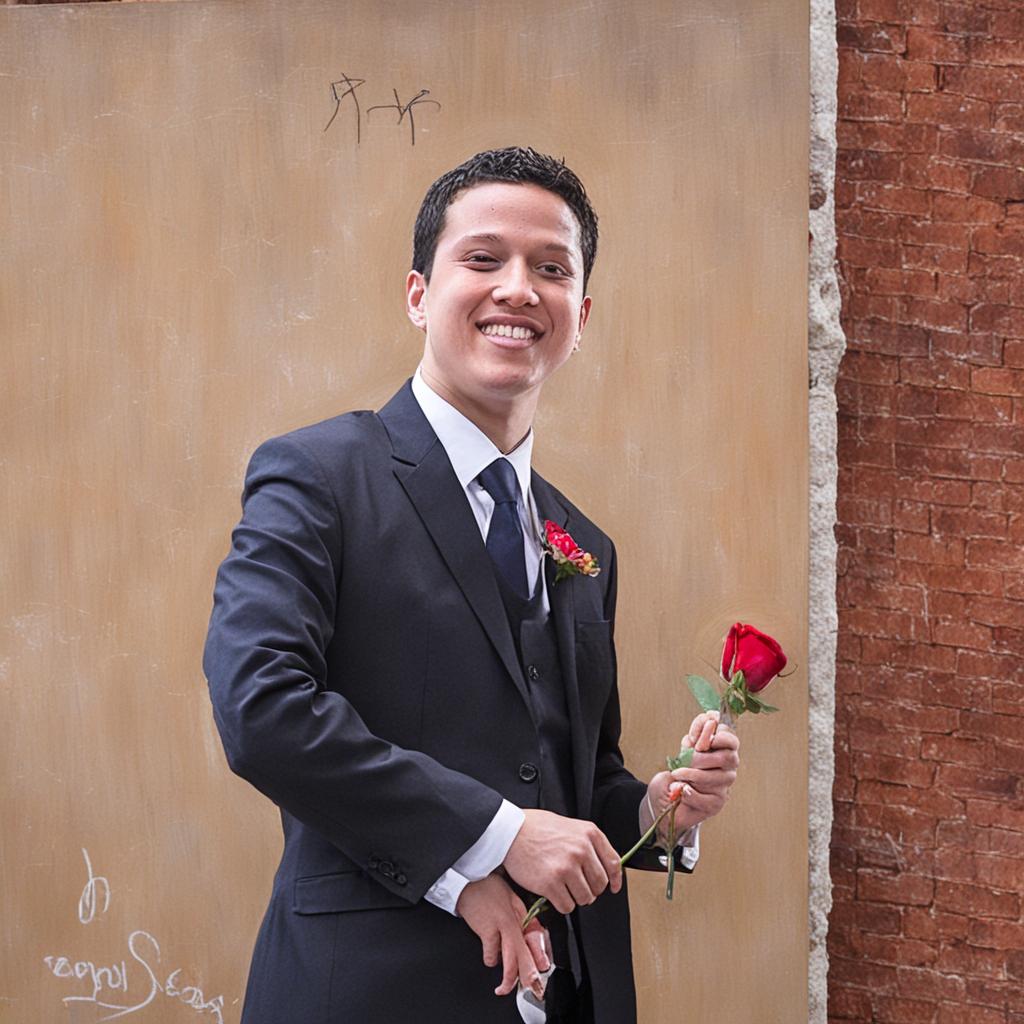} & 
        \includegraphics[width=0.139\textwidth]{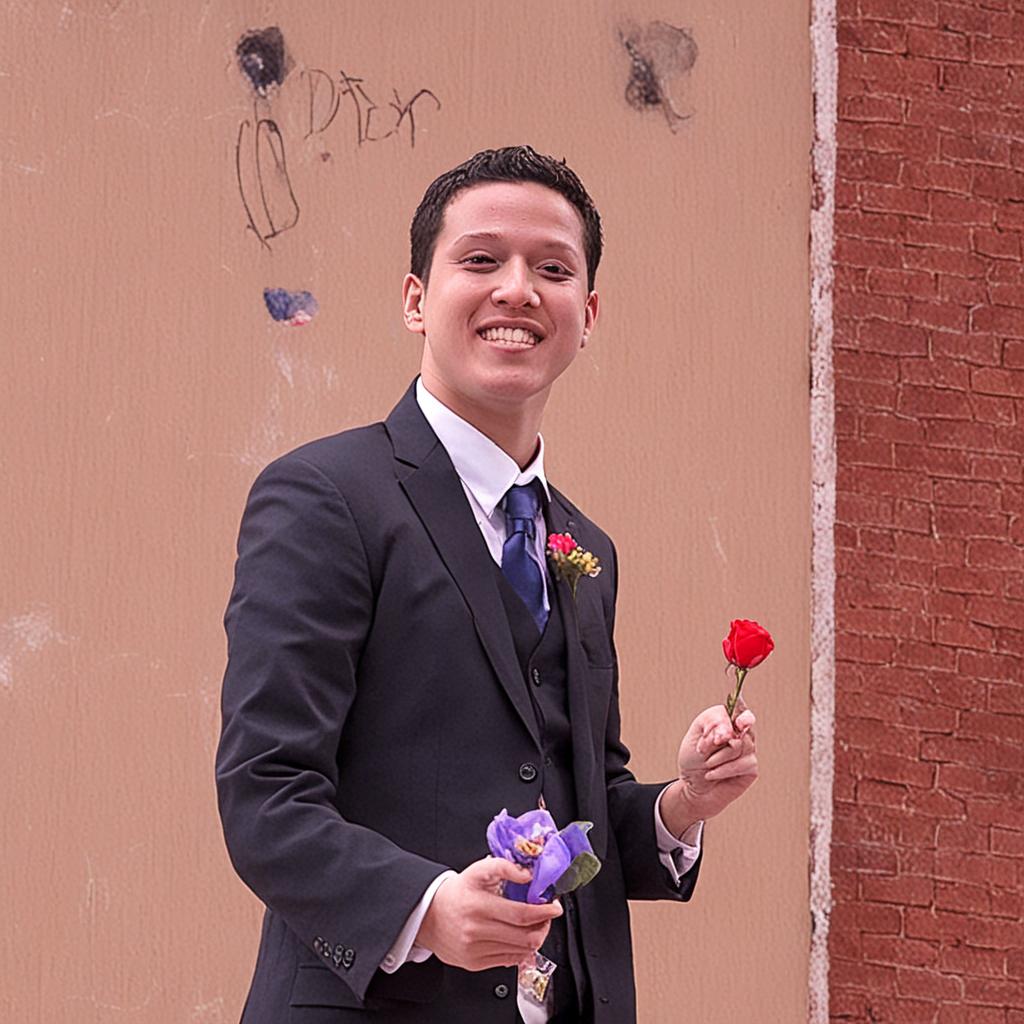} & 
        \includegraphics[width=0.139\textwidth]{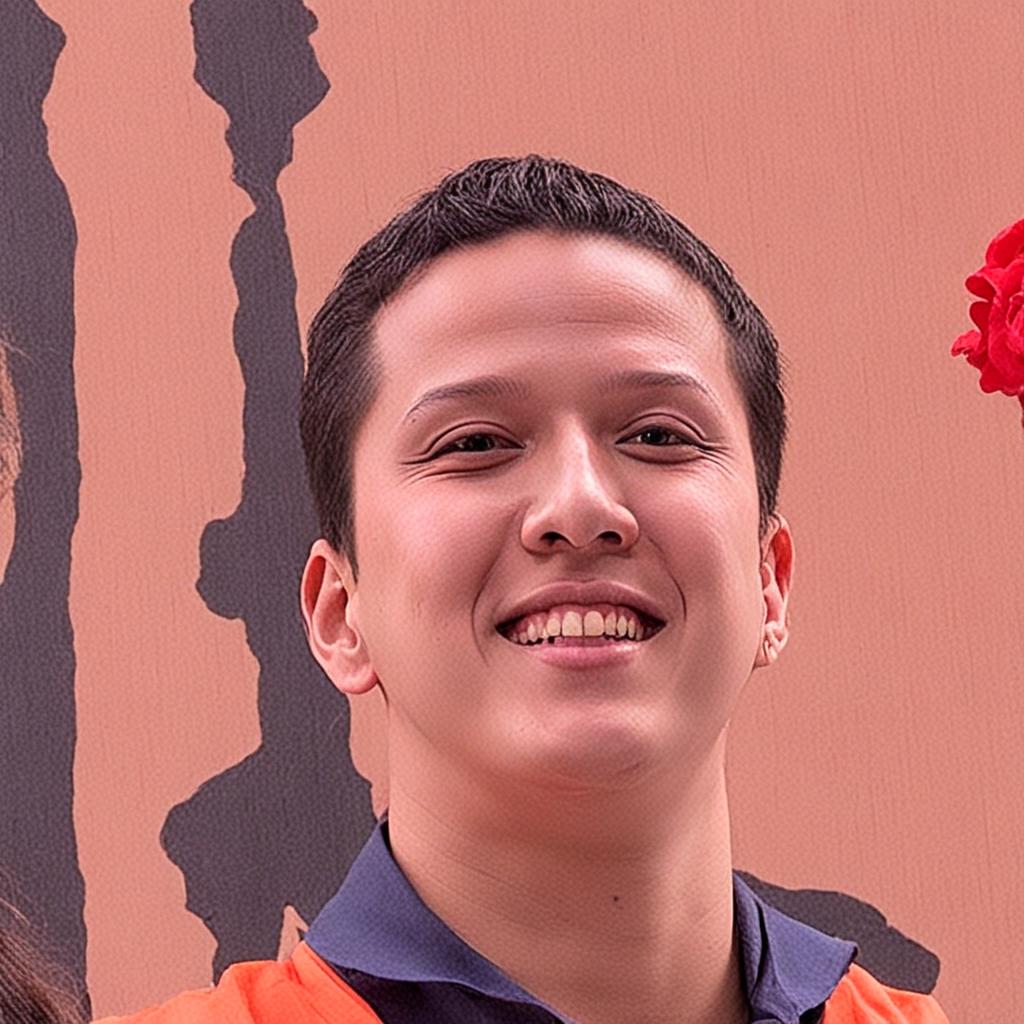} & 
        \includegraphics[width=0.139\textwidth]{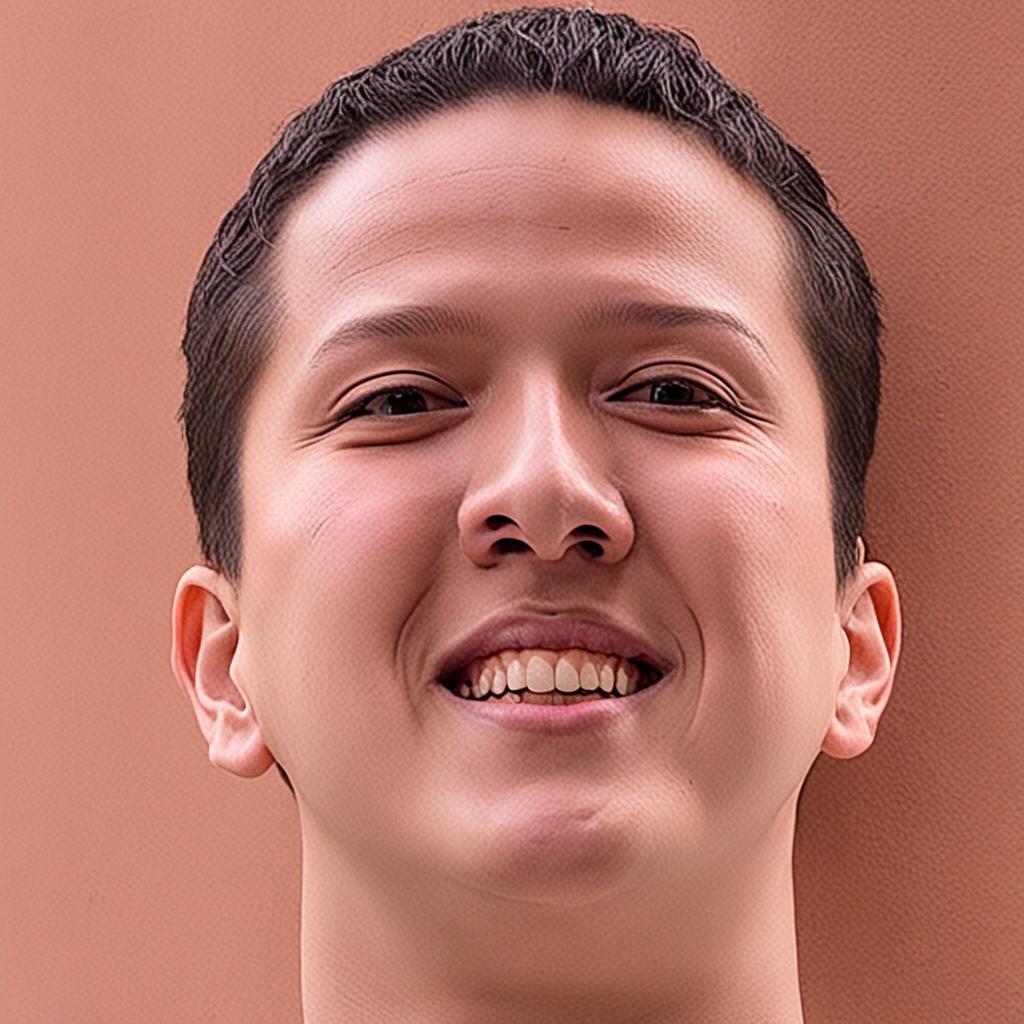} & 
        \includegraphics[width=0.139\textwidth]{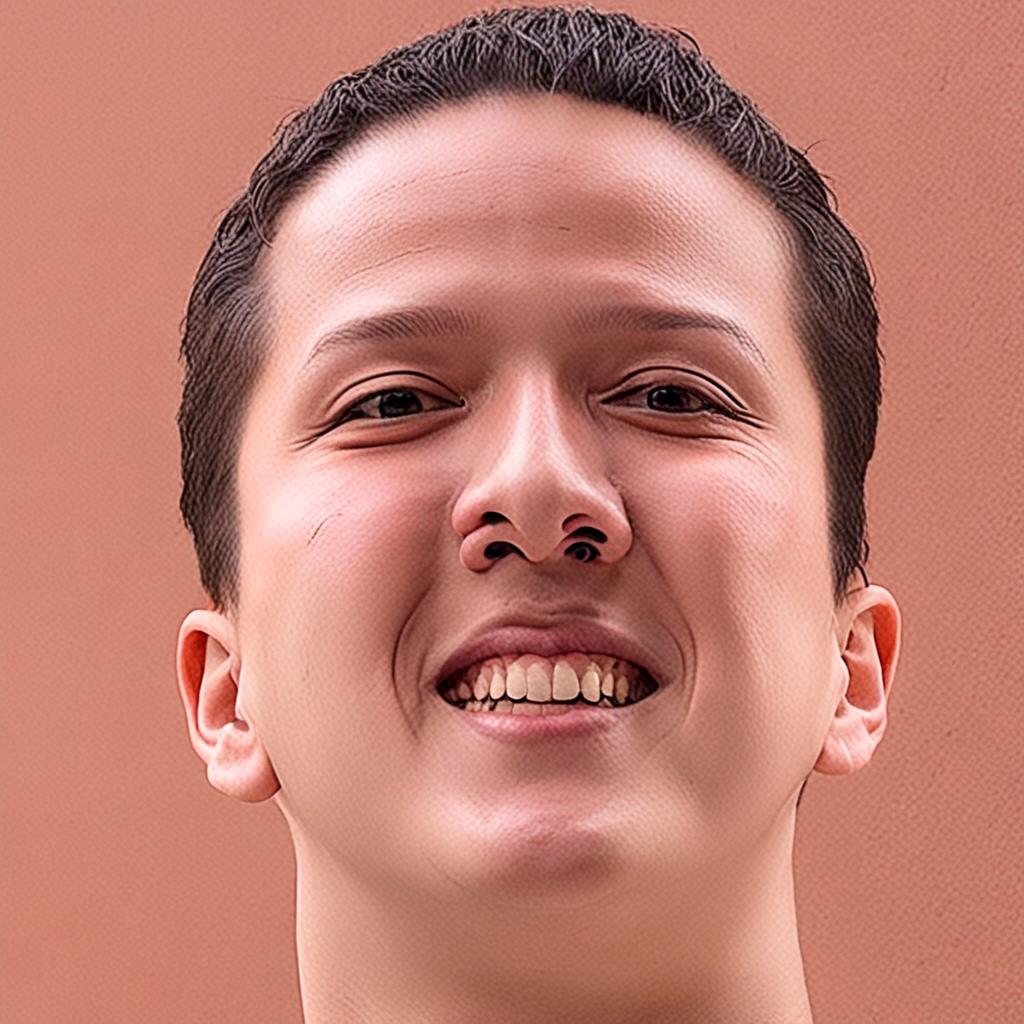} \\
    \end{tabular}
    }
    \caption{\textbf{Effect of varying rescaling weights.} Merging the two branches substantially improves identity preservation while maintaining text controllability comparable to the individual branches when rescaling weights are set to relatively small values. Text prompt: ``a painting of a man in the style of Banksy''. $\alpha$ denotes the rescaling weight in the adapter branch, while $\beta$ represents the rescaling weight in the text embedding branch.}

\label{fig:appendix_rescaling_weight}
\vspace{-10pt}
\end{figure*}

We analyze the impact of rescaling weights for the text embedding and adapter branches in Figure~\ref{fig:appendix_rescaling_weight}. Our analysis reveals that these weight parameters enable fine-grained control over the balance between identity preservation and text controllability.
Importantly, our results demonstrate that the combination of these two branches yields substantial improvements in identity preservation compared to using either branch independently.

\section{Text Prompts}
\label{sec:appendix_prompt}
Table~\ref{tab:prompts} presents the complete set of 20 text prompts employed in our quantitative evaluation. These prompts encompass diverse semantic modifications, including background alterations, environmental context changes, action variations, and artistic style transfers, enabling comprehensive assessment of identity preservation and text alignment across varied generation scenarios.

\section{Computational Efficiency Analysis}
\label{sec:appendix_influence}
Table~\ref{tab:inference_time} reports the inference time and memory consumption for both single-branch and dual-branch variants with a batch size of 8. The results demonstrate that the computational overhead introduced by the dual-branch architecture is negligible. This efficiency stems from the fact that each branch requires only a single forward pass, while the primary computational bottleneck remains the iterative denoising process of the diffusion model. 


\begin{table*}[b]
    \centering
    \setlength{\tabcolsep}{3pt}   
    \renewcommand{\arraystretch}{1.2}
    \caption{The prompts used in the quantitative evaluation. The placeholder ``<class word>'' is replaced with either ``man'' or ``woman''.}
    \begin{tabular}{@{\hspace{0.8cm}}c@{\hspace{0.8cm}}}
        \toprule
        a <class word> latte art \\
        \hline
        colorful graffiti of a <class word> \\
        \hline
        a <class word> on the beach \\
        \hline
        a <class word> wearing yellow jacket, and driving a motorbike \\
        \hline
        a <class word> wearing the sweater, a backpack and camping stove, outdoors, RAW, ultra high res \\
        \hline
        a <class word> wearing a scifi spacesuit in space \\
        \hline
        a <class word> eating bread in front of the Eiffel Tower \\
        \hline
        a <class word> in the style of wash painting \\
        \hline
        a <class word> wearing a Superman outfit \\
        \hline
        a <class word> with red hair \\
        \hline
        a <class word> wearing a doctoral cap \\
        \hline
        a <class word> holding a bottle of red wine \\
        \hline
        a <class word> driving a bus in the desert \\
        \hline
        a <class word> playing basketball \\
        \hline
        a <class word> swimming in the pool \\
        \hline
        a <class word> climbing a mountain \\
        \hline
        a <class word> on top of pink fabric \\
        \hline
        a street art mural of a <class word> \\
        \hline
        a black and white photograph of a <class word> \\
        \hline
        a pointillism painting of a <class word> \\
        \bottomrule
    \end{tabular}
    \label{tab:prompts}
    \vspace{-0.3cm}
    \end{table*}
    
\begin{table*}[t]
\centering
\caption{Inference time and memory consumption comparison between single-branch and dual-branch architectures with batch size of 8.}

\vspace{-5pt}
        \begin{tabular}{@{\hspace{0.5cm}}l c c c c@{\hspace{0.5cm}}}
          \toprule
          \quad  & Original SDXL & Adapter branch & Textual branch & Two branches
          \\
          \midrule
          Inference time in seconds & 25.91 & 28.12 & 25.96 & 28.15   \\
          Inference memory in MB & 37610.77 & 39735.84 & 39187.82 & 39967.71 \\
          \bottomrule
        \label{tab:inference_time}
        \vspace{-0.55cm}
    \end{tabular}
\end{table*}

\end{document}